\providecommand{\iComAILogoFile}{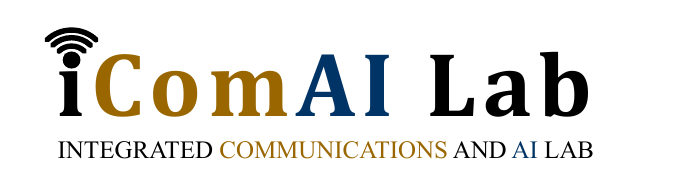}
\providecommand{\iComAILogoOpacity}{1}
\PassOptionsToPackage{table}{xcolor}
\PassOptionsToPackage{hyphens}{url}
\PassOptionsToPackage{capitalize}{cleveref}
\PassOptionsToPackage{sort&compress}{natbib}
\documentclass{fairmeta}
\usepackage{amsmath,amssymb,amsfonts,array,textcomp,makecell,tabularx,pifont}
\usepackage{float,longtable,wrapfig,needspace}
\definecolor{cvprblue}{rgb}{0.21,0.49,0.74}
\usepackage{array,tabularx,multirow}
\usepackage{microtype}
\usepackage{colortbl}
\definecolor{resultgroup}{RGB}{235,235,248}
\definecolor{resultours}{RGB}{247,238,233}
\definecolor{resultgain}{RGB}{38,118,107}
\definecolor{resultloss}{RGB}{179,75,72}
\newcommand{\gain}[1]{\textcolor{resultgain}{#1}}
\newcommand{\loss}[1]{\textcolor{resultloss}{#1}}
\definecolor{tableband}{RGB}{235,235,248}
\newcommand{\featyes}{\ensuremath{\bullet}}
\newcommand{\featno}{\ensuremath{\circ}}

\newcolumntype{C}{>{\centering\arraybackslash}X}
\newcommand{\bench}{EgoWitness}
\newcommand{\agentname}{EgoClue}
\newcommand{\blank}{\hphantom{00.00}}

\newcolumntype{Y}{>{\raggedright\arraybackslash}X}
\newcommand{\tool}[1]{\texttt{#1}}
\newcommand{\mlogo}[1]{\makebox[1.15em][c]{$\vcenter{\hbox{%
  \includegraphics[width=1.1em,height=0.85em,keepaspectratio]{figures/logos/#1}}}$}%
  \hspace{0.3em}}
\renewcommand{\dbltopfraction}{0.85}

\definecolor{slotink}{RGB}{51,71,88}
\definecolor{slotborder}{RGB}{177,190,201}
\definecolor{slotbg}{RGB}{246,248,250}

\usepackage{titletoc}
\usepackage{multicol}
\titlecontents{lsection}[1.6em]
  {\addvspace{3.5pt}\small\bfseries}
  {\contentslabel[\textcolor{cvprblue}{\thecontentslabel}]{1.6em}}
  {\hspace*{-1.6em}}
  {\hfill\contentspage}
  [\addvspace{0.5pt}]
\titlecontents{lsubsection}[4.2em]
  {\footnotesize}
  {\contentslabel{2.6em}}
  {\hspace*{-2.6em}}
  {\titlerule*[0.45pc]{.}\contentspage}
\makeatletter
\newcommand{\suppkeepsubsections}{%
  \patchcmd{\ttl@outnoblock}{\addpenalty{\z@}}{\addpenalty{\@M}}{}%
    {\PackageWarning{preamble}{titletoc penalty patch failed}}}
\makeatother

\newcommand{\maketitlesupplementarycontents}{%
  {\centering\Large\sffamily\bfseries Supplementary Material\par}\medskip
  \startcontents[supp]
\noindent\textbf{Overview.} This supplement documents how \bench{} is built
and how models are evaluated on it.
Appendices~\ref{sec:supp_sources}--\ref{sec:supp_questions} cover the data
sources, the room, object, and landmark annotations, and question
construction and verification.
Appendices~\ref{sec:supp_training}--\ref{sec:supp_synthesis} cover
\agentname{} training, the evaluation setup, additional results, and answer
synthesis, and \cref{sec:supp_case_studies} presents case studies and error
analysis.
\par\vspace{7pt}
\noindent\begin{minipage}{\linewidth}
\hypersetup{linkcolor=black}%
{\color{cvprblue}\rule{\linewidth}{0.8pt}}\par\vspace{3pt}
{\large\bfseries Contents}\par\vspace{-2pt}
\setlength{\columnsep}{2.2em}\suppkeepsubsections\raggedcolumns
\begin{multicols}{2}
\printcontents[supp]{l}{1}{\setcounter{tocdepth}{2}}
\end{multicols}
\vspace{-5pt}
{\color{cvprblue}\rule{\linewidth}{0.4pt}}
\end{minipage}
\par\vspace{6pt}
}

\setcitestyle{numbers,square,comma}
\title{EgoWitness: Benchmarking Instance-Level Spatial\\Memory and Reasoning in Egocentric Video Streams}
\renewcommand{\titlelist}{{\fontsize{19.8}{23.5}\selectfont\sffamily\bfseries \raisebox{-1.5pt}{\includegraphics[width=20pt]{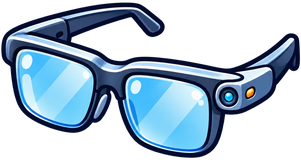}}\hspace{3pt}EgoWitness: Benchmarking Instance-Level Spatial\\Memory and Reasoning in Egocentric Video Streams}}
\renewcommand{\authorlist}{{\fontsize{10}{13}\selectfont
Zhengru~Fang$^{1,*}$, Zehao~Liu$^{2,*}$, Mingyu~Tsoi$^{1}$, Yaoyu~Zhu$^{3}$, Junlong~Li$^{4}$, Fengmao~Lv$^{5}$, Sijie~Cheng$^{2,6}$, Yuanliang~Sun$^{6}$, Fang~Yang$^{2}$, Jun~Zhang$^{1}$}}
\renewcommand{\affiliationlist}{{\fontsize{9}{11.5}\selectfont
$^{1}$The Hong Kong University of Science and Technology \quad $^{2}$Tsinghua University\\
$^{3}$Southeast University \quad $^{4}$The Hong Kong Polytechnic University\\
$^{5}$Southwest Jiaotong University \quad $^{6}$RayNeo\\
$^{*}$Equal contribution.}}
\hypersetup{pdftitle={EgoWitness: Benchmarking Instance-Level Spatial Memory and Reasoning in Egocentric Video Streams},pdfauthor={Zhengru Fang, Zehao Liu, Mingyu Tsoi, Yaoyu Zhu, Junlong Li, Fengmao Lv, Sijie Cheng, Yuanliang Sun, Fang Yang, Jun Zhang}}

\titlespacing*{\section}{0pt}{13pt plus 3pt minus 2pt}{6pt plus 1pt}
\titlespacing*{\subsection}{0pt}{9pt plus 2pt minus 2pt}{4pt plus 1pt}
\titlespacing*{\subsubsection}{0pt}{7pt plus 2pt minus 1pt}{3pt plus 1pt}
\titlespacing*{\paragraph}{0pt}{6pt plus 1pt minus 1pt}{0.7em}
\begin{document}
\abstract{Recalling where and when an object was last seen and who saw it requires
recognizing it across days and people. Existing long-video benchmarks focus
on events, object categories, or relative locations, leaving instance-level
spatial memory largely untested in multimodal models.
To address this gap, we present \bench, a benchmark
for instance-level spatial memory and reasoning in egocentric video. Its
8,453 human-verified questions span approximately 324 hours of week-long
and short-horizon recordings and test seven abilities, from recognizing an object or room to recalling
who saw it and reasoning about room visits. Over one million reviewed masks of
34 registered landmarks and 21 movable-object labels are linked through
room timelines. Among twelve direct vision-language models (VLMs), open-weight models reach at most 46.97\% accuracy, and none exceeds
35\% on identifying who saw an object last. We therefore propose
\agentname, a spatial-memory agent that retrieves candidate intervals,
verifies instance identity, and computes spatial and temporal relations
with five tools. Two-stage training with supervised fine-tuning (SFT) and
Group Relative Policy Optimization (GRPO) raises its 4B controller from
23.64\% to 72.65\% accuracy on 1,320 held-out questions. Evidence-trace synthesis
matches the best direct VLM at 74.85\% on week-long questions and achieves
80.56\% in environments excluded from
training.\footnote{Project page: \url{https://egowitness.github.io/}. Code, dataset, and annotations are coming soon.}
}
\maketitle
\section{Introduction}
\label{sec:intro}
Smart glasses equipped with cameras and microphones hold the
promise of becoming personal AI assistants that understand and
remember our everyday experiences. Such assistants must be capable of more
than simply recognizing visible objects. They must also remember
where and when those objects appeared.
Imagine a user rushing to an important meeting and asking,
``Did I last see my work laptop in the upstairs living room or
the downstairs one?'' Answering it requires more than recognizing
a laptop or a living room. The assistant must distinguish the
user's laptop from lookalikes, recognize the specific room,
and connect the sighting to its time and
location~\cite{egotracks,airroom,remind}.
In a shared environment, answering ``Who saw it last?''
also requires linking observations across people.
We refer to this capability as
\emph{instance-level spatial memory and reasoning}.
It becomes particularly challenging in multi-day, multi-wearer
egocentric streams, where the same objects and rooms reappear
across changing viewpoints, repeated visits, and different people.

\begin{figure*}[!t]
\centering
\includegraphics[width=0.90\textwidth]{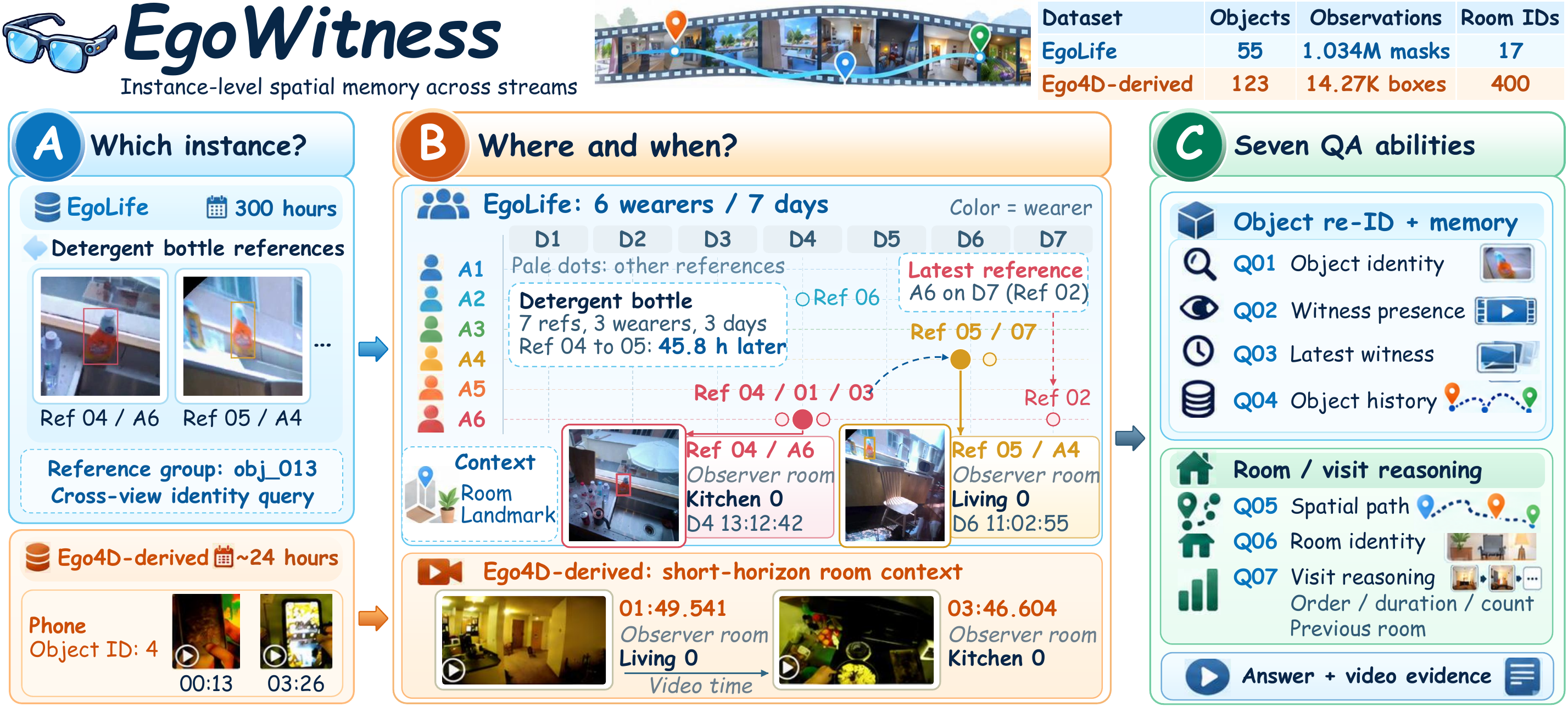}
\caption{\textbf{Instance-level spatial memory across egocentric streams.}
(A)~Instance-reference comparison. (B)~Historical observations with wearer,
time, and room context. (C)~Seven QA abilities in two groups: object identity
and memory, and room identity and visit reasoning.}
\label{fig:overview}
\end{figure*}

Despite progress in long-video understanding~\cite{egoschema,videomme,hourvideo},
this capability is rarely evaluated directly. Object-centric
benchmarks~\cite{egotracks,refego,egomask} track or localize targets within
one recording, while memory-centric QA benchmarks~\cite{egolife,ucsbench,maegoqa}
ask about past events and relative locations across days or wearers.
Both rarely test whether today's object is the physical instance seen
days earlier or who saw it last. Asking where scissors were last used
differs from identifying the pair a housemate saw two days ago.
Few annotations link each sighting to an instance, a room, and its wearer.

To address this gap, we introduce \bench\ for instance-level spatial
memory across week-long, multi-wearer recordings and short-horizon
environments. Its human-reviewed annotations link questions to evidence
and record each wearer's room over time to distinguish lookalike instances
across visits. In \cref{fig:overview}A, detergent-bottle views are matched
to references, while \cref{fig:overview}B links three wearers' sightings
across three days to times and rooms. The seven abilities in
\cref{fig:overview}C form two groups. Object identity and memory questions
test which instance appeared, when, and who saw it, while room identity and visit
reasoning questions test room recognition, paths, and visits.
Ego4D-derived recordings with aligned
annotations~\cite{ego4d,egotracks,egomask,refego,egoenv,egoplan_media}
extend evaluation beyond one home. As shown in \cref{tab:related}, \bench\ alone
combines object and room re-identification, cross-wearer questions,
scored evidence, masks, and room instances with intervals.

We evaluate twelve vision-language models (VLMs) that answer week-long
questions directly from uniformly sampled frames. The strongest reaches 73.56\% accuracy, the best
open-weight model reaches 46.97\%, and no model exceeds 35\% on
identifying who saw an object last. They cannot
request another view or search for later sightings.

These results motivate an agent that revisits clips, verifies identity,
and compares sightings across time. Tool-using video
agents~\cite{videoagent_wang,videoagent_fan,egor1} mainly retrieve
relevant clips. Building on them, we propose \agentname, which also
verifies sightings against references and computes room paths and
durations with five tools. Its 4B controller learns tool use
through supervised fine-tuning (SFT) followed
by Group Relative Policy Optimization (GRPO)~\cite{grpo}.

Our contributions are summarized as follows:
\begin{itemize}\setlength{\itemsep}{1pt}\setlength{\parskip}{0pt}
\item \textbf{An instance-level spatial-memory benchmark.}
\bench\ provides 8,453 questions across seven abilities over approximately
324 hours. Twelve direct VLMs struggle most when identifying who saw an
object last.
\item \textbf{Room-guided, human-reviewed annotations.}
Over one million masks cover 34 registered landmark
instances at 1\,Hz and 21 movable-object labels at 5\,Hz, linked through
room timelines and reviewed identities.
\item \textbf{A spatial-memory agent framework.}
\agentname\ retrieves candidate intervals, verifies instance identity, and
computes room paths and visit durations with five tools. Two-stage training
raises its 4B controller from 23.6\% to 72.7\% on 1,320 held-out questions.
Trace-based synthesis reaches 74.8\%, matching the best direct VLM.
\end{itemize}

\begin{table*}[!t]
\caption{\textbf{Related benchmarks and annotation resources.}
(a)~Identity-related QA. (b)~Object and environment annotations.
\featyes: included in the reported protocol; \featno: not specified.
QA evidence: reference evidence per question; \bench{} also scores it.
}
\label{tab:related}
\centering\footnotesize
\setlength{\tabcolsep}{2pt}
\renewcommand{\arraystretch}{1.08}
\begin{tabular}{@{}>{\raggedright\arraybackslash}m{0.19\textwidth}>{\centering\arraybackslash}m{0.102\textwidth}>{\centering\arraybackslash}m{0.145\textwidth}>{\centering\arraybackslash}m{0.145\textwidth}>{\centering\arraybackslash}m{0.132\textwidth}>{\centering\arraybackslash}m{0.075\textwidth}>{\centering\arraybackslash}m{0.16\textwidth}@{}}
\toprule
\rowcolor{tableband}\multicolumn{7}{l}{\textbf{(a) Memory and spatial question answering}} \\
Benchmark & Scale & Span & Object identity & Place identity & \makecell{Cross-\\wearer} & QA evidence \\
\midrule
EgoLifeQA~\cite{egolife} & 300 h & 7 days & Object recall & Location recall & \featno & Evidence time \\
MA-EgoQA~\cite{maegoqa} & 266 h & 7 days & Object usage & Event context & \featyes & Context windows \\
HD-EPIC~\cite{hdepic} & 41 h & 3 days & Object itineraries & Fixture identity & \featno & \featno \\
\mbox{AMB / AMEGO~\cite{amego}} & 22.7 h & 14 min avg. & Interaction links & Activity zones & \featno & \featno \\
UCS-Bench~\cite{ucsbench} & $170+$ h & Seconds to hours & Trajectory recall & Relative places & \featno & Evidence segments \\
\rowcolor{resultours}\mlogo{egowitness-glasses}\textbf{\bench} & $\sim$324 h & Minutes to week & \textbf{Object Re-ID} & \textbf{Room Re-ID} & \featyes & \textbf{Scored evidence} \\
\bottomrule
\end{tabular}
\par\vspace{3pt}
\begin{tabular}{@{}>{\raggedright\arraybackslash}m{0.19\textwidth}>{\centering\arraybackslash}m{0.102\textwidth}>{\centering\arraybackslash}m{0.145\textwidth}>{\centering\arraybackslash}m{0.145\textwidth}>{\centering\arraybackslash}m{0.132\textwidth}>{\centering\arraybackslash}m{\dimexpr0.235\textwidth+4pt\relax}@{}}
\toprule
\rowcolor{tableband}\multicolumn{6}{l}{\textbf{(b) Object tracking, grounding, and environment resources}} \\
Resource & Scale & Span & Boxes & Masks & Room annotations \\
\midrule
EgoTracks~\cite{egotracks} & 602.9 h & 6.13 min avg. & \featyes & \featno & \featno \\
RefEgo~\cite{refego} & 41 h & 5 to 20 s & \featyes & \featno & \featno \\
EgoMask~\cite{egomask} & 315 videos & 5 s to 16 min & \featyes & \featyes & \featno \\
LongEgoRefer~\cite{longegorefer} & 433 videos & 45 min avg. & \featyes & \featno & \featno \\
EgoEnv~\cite{egoenv} & 1,259 scenes & Source-specific & \featno & \featno & Types + local IDs \\
\rowcolor{resultours}\mlogo{egowitness-glasses}\textbf{\bench} & $\sim$324 h & Minutes to week & \featyes & \featyes & Instances + intervals \\
\bottomrule
\end{tabular}
\par\vspace{2pt}\raggedright\footnotesize
Days denote collection span; avg. denotes mean video length. Cross-wearer requires multi-wearer evidence.
EgoMask reports its test set; EgoEnv reports the Ego4D subset.
EgoEnv provides room-type intervals and local instance metadata.
\end{table*}

\section{Related work}
\label{sec:related}
\paragraph{Egocentric data and representations.}
Large egocentric datasets supply the recordings on which spatial memory
builds. EPIC-KITCHENS~\cite{epickitchens,epic100} annotates kitchen
actions and objects, Ego4D~\cite{ego4d} spans diverse daily environments,
and Goal-Step~\cite{goalstep} adds procedural labels.
Ego-Exo4D~\cite{egoexo4d} synchronizes first- and third-person views, while
EgoExoLearn~\cite{egoexolearn} connects asynchronous demonstrations.
EgoVLP, EgoVLPv2, and LaViLa~\cite{egovlp,egovlpv2,lavila} learn
video-language representations from narrations. EgoLife~\cite{egolife}
records shared week-long experience, and HD-EPIC~\cite{hdepic} links object
movements to fixtures and connects them into itineraries. However, these
resources mainly organize supervision around activities, narrations, or
object movements. In comparison, \bench\ records which room
each wearer occupies and which physical instance each mask shows across
days and wearers.

\paragraph{Long-video and embodied question answering.}
EgoSchema and
LongVideoBench~\cite{egoschema,longvideobench} pose questions over long
clips, while Video-MME, MLVU, LVBench, and HourVideo~\cite{videomme,mlvu,lvbench,hourvideo}
expand duration and task coverage. SuperMemory-VQA~\cite{supermemory} adds
unanswerable memory queries, and EgoMonth~\cite{egomonth} extends personal
memory across 20 to 120 days. Embodied QA and OpenEQA~\cite{eqa,openeqa}
test navigation and episodic scene memory,
and R2R, REVERIE, and SOON~\cite{r2r,reverie,soon} ground navigation in
instructions and target descriptions. The memory QA benchmarks in
\cref{tab:related}a~\cite{egolife,maegoqa,ucsbench} ask about object recall,
usage, and trajectories. Beyond3D~\cite{beyond3d} tests spatiotemporal reasoning
about relocated objects after they leave view. In contrast, \bench\ links room
and object identities across days and wearers to recover where, when, and by
whom instances were observed, scoring both answers and evidence.

\paragraph{Instance-level objects, places, and memory.}
\cref{tab:related}b summarizes instance-level resources.
EgoObjects~\cite{egoobjects} separates categories from physical instances,
and MyEgo~\cite{myego} personalizes object queries.
EgoTracks, RefEgo, and EgoMask~\cite{egotracks,refego,egomask}
localize targets, and IT3DEgo~\cite{it3dego} tracks registered instances in 3D.
Whareformer~\cite{chalk2026whareformer} combines appearance and 3D location
to associate observations with object tracks and maintain locations out of view.
AMEGO~\cite{amego} links interactions and locations in memory;
EgoEnv~\cite{egoenv} learns environment-aware representations.
ESOM~\cite{esom} builds online object memory from detections and tracks.
Its oracle experiments identify detection and tracking bottlenecks,
motivating identity and evidence verification. For cross-day, cross-wearer
questions, \bench\ links physical instances to local tracks, room timelines,
and the wearers who saw them.

\paragraph{Memory and tool-using video agents.}
Video agents extend memory through retrieval and tools. MovieChat and
LLoVi~\cite{moviechat,llovi} reason through compressed memory and clip
captions. Wang \textit{et~al.}~\cite{videoagent_wang} select evidence
iteratively, while Fan \textit{et~al.}~\cite{videoagent_fan} maintain event
and object memory. VideoTree~\cite{videotree} adapts a video hierarchy to
the query, Video-RAG~\cite{videorag} retrieves auxiliary visual information,
and EgoButler~\cite{egolife} retrieves relevant life experience.
Self-RAG, Reflexion, and RAFT~\cite{selfrag,reflexion,raft} refine
generation through critique, feedback, or retrieval-conditioned
training. Ego-R1~\cite{egor1} learns dynamic tool
selection through SFT and reinforcement learning. However, retrieval mainly
identifies candidate content, whereas finding who saw an object last also requires
verifying instance identity and checking later sightings. Building on
these agents, \agentname\ pairs visual verification with deterministic path and
duration tools. Its controllers share perception, retrieval, and execution
budgets across training stages, which isolates how training changes
evidence acquisition.

\section{The EgoWitness benchmark}
\label{sec:data}
\begin{figure*}[!t]
\centering
\includegraphics[width=\textwidth]{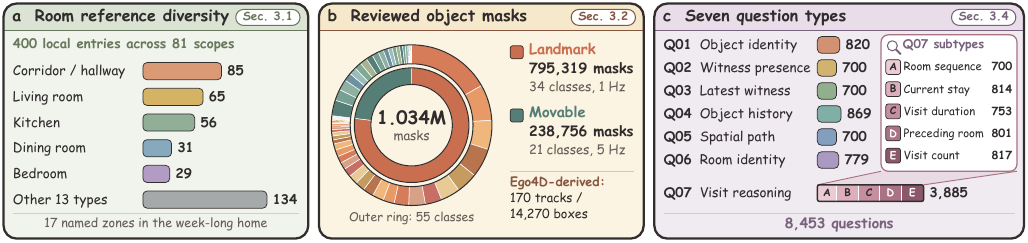}
\caption{\textbf{Room references, object annotations, and question coverage.}
\textbf{a.} The external room catalog contains 400 source-local reference entries across
81 scopes and 18 room types, separate from the 17 named zones in the shared home.
\textbf{b.} The week-long component contains 34 registered landmark instances and 21 movable-object labels.
\textbf{c.} Questions cover both components; Q07 is split into five subtypes.}
\label{fig:statistics}
\end{figure*}

\begin{figure*}[!t]
\centering
\includegraphics[width=\textwidth]{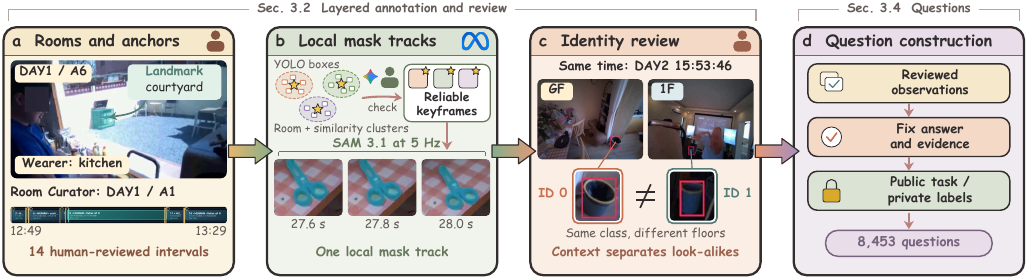}
\caption{\textbf{From reviewed observations to evidence-linked questions.}
\textbf{a.} Room timelines and landmark locations supply spatial context.
\textbf{b.} SAM~3.1 tracks Gemini- and human-checked keyframes.
\textbf{c.} Annotators resolve instance identity using reference views and
room context. \textbf{d.} Accepted observations fix question facts before
wording. Full review interfaces appear in \cref{fig:supp_tools}.}
\label{fig:pipeline}
\end{figure*}

We build \bench\ by selecting recordings, annotating rooms and objects
with human review, and converting the annotations into verified questions.

\FloatBarrier\Needspace{24\baselineskip}
\subsection{Source recordings and scope}
\begin{wraptable}{r}{0.54\textwidth}
\caption{\textbf{Source collection and annotation coverage.} L/M denote
landmark instances/movable labels. Room references retain their source-local scope.}
\label{tab:components}
\centering\scriptsize
\setlength{\tabcolsep}{3pt}
\renewcommand{\arraystretch}{1.0}
\begin{tabularx}{\linewidth}{@{}lCC@{}}
\toprule
\textbf{Attribute} & \textbf{Week-long} & \textbf{Short-horizon} \\
\midrule
\rowcolor{tableband}\multicolumn{3}{l}{\textit{Recordings and spatial coverage}} \\
Video source & EgoLife & Ego4D-derived \\
Source hours & $\sim$300 & 23.46 \\
Recording units & 42 wearer-days & 43 videos \\
Room references & 17 zones & 400 local entries \\
\rowcolor{tableband}\multicolumn{3}{l}{\textit{Reviewed annotations and questions}} \\
Label space & 34 L + 21 M & 123 raw names \\
Sampling, Hz & 1 L / 5 M & 1 / 2 \\
\rowcolor{resultours}Visible observations & \textbf{1,034,075 masks} & \textbf{14,270 boxes} \\
Local tracks & 92,669 & 170 \\
\midrule
Reviewed questions & \textbf{7,666} & \textbf{787; eval. only} \\
\bottomrule
\end{tabularx}
\end{wraptable}

\bench\ spans approximately 324 hours of long- and short-horizon recordings,
summarized in \cref{tab:components}.

\paragraph{Long-horizon recordings.}
EgoLife~\cite{egolife} provides approximately 300 hours over 42 wearer-days
from six wearers in 32,001 source MP4 files. We add
17,269 room intervals over 17 named rooms and zones, covering all recorded
wearer-hours, and 1,034,075 masks of 34 registered landmark instances and
21 movable-object labels in \cref{fig:statistics}b.

\paragraph{Short-horizon recordings.}
The 43 selected Ego4D-derived full videos span 23.46 hours. We align
EgoEnv room annotations with EgoMask and RefEgo object
annotations~\cite{egomask,refego,egoenv} by source-video ID and time
interval. The 400 source-local room references span
81 scopes and 18 room types in \cref{fig:statistics}a. EgoTracks supplies
tracking lineage and ego\_plan provides full-duration copies of 15
videos~\cite{egoplan_media}. Supp.~Sec.~\ref{sec:supp_sources} details
selection and alignment.

\par\WFclear
\subsection{Layered annotation and review}
\label{sec:annotation}
Room annotation comes first in \cref{fig:pipeline}a because shared room identities
constrain where instances appear. SAM~3.1~\cite{sam31} then turns reliable anchors
into local tracks, which annotators link into physical instances before
accepted observations fix question facts. Supp.~Sec.~\ref{sec:supp-data}
details each step.

\paragraph{Room timelines.}
The room layer records each wearer's room over time.
Annotators edit intervals in the Online Room Curator System. Reviewed
intervals cover 4,257 segments with gaps of at most 0.5\,s.
Reviewers corrected 3,383 intervals. Room IDs are shared across wearers and
days, and unknown and outdoor states stay explicit.

\paragraph{Registered landmarks.}
Landmarks are registered visual references, such as fixtures and designated
reference objects, each with a reference room, floor, and zone.
Because landmarks rarely move in the videos, 1\,Hz sampling suffices to
capture their spatial relation to the wearer.
SAM~3.1~\cite{sam31}, which builds on promptable and memory-based
segmentation~\cite{sam,sam2,sam3,xmem,cutie,deva}, propagates accepted
landmark detections at this rate. All 34 landmark labels link to reviewed instance
records observed by multiple wearers across days.

\paragraph{Movable objects.}
Movable objects are portable everyday items, including bottles, scissors,
and bowls. Because people may move these objects at any time, we sample
them at 5\,Hz for finer-grained tracks.
A YOLO detector proposes candidate boxes, and confidence, persistence,
duplicate, and room filters retain 19,921 anchors. These anchors are
clustered by appearance similarity within each room segment. Gemini~3.8 Flash~\cite{gemini38flash} scores whether
a cluster shows the same object from its appearance, function, and scene
context, and annotators check every selected keyframe. SAM~3.1 tracks these
keyframes forward and backward at 5\,Hz into local mask tracks in
\cref{fig:pipeline}b. A track stays within one processing unit, so
annotators link tracks into physical instances in \cref{fig:pipeline}c. Room
context separates look-alikes, such as two bins seen at once
on two floors.

\paragraph{Context and logical checks.}
Room states, conflicts with registered floors, duplicate detections, and
simultaneous placements of one instance in disconnected rooms trigger further
review. Annotators inspect the video to remove false boxes and correct identity
links or room boundaries. Accepted masks then fix the question facts in
\cref{fig:pipeline}d. Supp.~Sec.~\ref{sec:supp_context_rules} details the
checks and corrections.

\subsection{A unified representation}
Questions in \cref{sec:benchmark} query identity, place, time, and observer
across both sources, so every accepted annotation becomes one observation
record. A record links a target instance to its wearer, time, box, mask,
local track, category, attributes, and provenance.
Questions ask about both observers and targets, so the schema separates three
locations. \emph{Observer location} is the wearer's
room from the room intervals, \emph{registered landmark location} comes from
the landmark registry, and \emph{\mbox{object location}} records the
established position of a target. The grill in \cref{fig:pipeline}a, for
example, lies in the courtyard but is seen from the kitchen. Supp.~Sec.~\ref{sec:supp_sampling}
maps both sources to common source time and geometry.

\subsection{Question design and verification}
\label{sec:benchmark}
\paragraph{Task definition.}
Each question specifies an object or room, public references, and a wearer
and time scope. Answers use observations within that scope and, for
historical questions, before the query time. Models select an answer
option, and agents also cite supporting observations. For the detergent bottle in
\cref{fig:overview}A and B, a model must decide whether a new crop shows the
same bottle, which wearer saw it, or who saw it last. The seven abilities
in \cref{fig:statistics}c form two groups: object identity and memory, and room identity and visit
reasoning. In the object group, Q01 compares reference images to distinguish
the same instance, different instances of one category, and different
categories. Q02 asks which of two wearers saw the target in a shared 6-s
window, Q03 who saw it last, and Q04 which time windows contain it. Q02 to Q04 require
matching sightings across separate recordings. In 103 of 120 evaluated
Q04 questions, the object appears in a window ending over 24 hours
before the query time, and in 78 it is seen by multiple wearers. In the
room group, Q05 finds a shortest path between rooms identified from images,
and Q06 applies the Q01 distinctions to room references, using room type
as the category. Q07 orders, times, or counts visits and recalls the
preceding room.

\paragraph{Five room-visit subtypes.}
Q07A orders the visited rooms, including returns, and Q07D
identifies the room before the current stay. Q07B times the current stay
up to the query time, Q07C times a specified completed visit
among repeated visits, and Q07E counts visits overlapping the observation
interval.
Supp.~Sec.~\ref{sec:supp_question_taxonomy} lists the templates.

\paragraph{Type-specific generation and verification.}
Reviewed identities and categories determine Q01/Q06 reference pairs.
Scoped target sightings determine Q02/Q03/Q04 answers. Q05 uses shortest
paths on the supplied room graph, and Q07 uses reviewed room intervals.
Answers, options, and evidence are fixed before language-model rewriting.
Historical checks cover all queried wearers and candidate windows before
the query time.
After model-assisted review, experts verify every question,
answer, and supporting evidence against source videos and room timelines
in the QA Audit Workbench. Supp.~Sec.~\ref{sec:qa_refinement} details this review.

\subsection{Splits, access conditions, and metrics}
\label{sec:protocol}
\paragraph{Splits.}
Week-long questions are grouped by core task and split into 5,750 training,
383 validation, and 1,533 test questions in a 75/5/20\% ratio.
Core task groups are disjoint.
Cross-environment evaluation reserves all 787 short-horizon questions and
their environments for testing, with SFT and GRPO confined to the week-long data.

\paragraph{Access conditions.}
Beyond the query, direct VLMs inspect preselected frames, while agents select video intervals
and retrieve coarse text memory summaries. Q05 supplies a room graph for path computation after visual
endpoint identification. Private room timelines, instance sightings, and
answers serve only question construction and scoring.

\paragraph{Metrics.}
Micro accuracy weights questions equally. Week-long Macro averages the seven
abilities so that the five Q07 subtypes do not dominate the overall score,
\begin{equation}
 \mathrm{Acc}_{\mathrm{macro}}=
 \frac{1}{K}\sum_{k=1}^{K}\frac{1}{N_k}
 \sum_{i\in\mathcal Q_k}\mathbf{1}[\hat a_i=a_i],
 \label{eq:accuracy}
\end{equation}
where $\mathcal Q_k$ contains $N_k$ queries of ability $k$ and $K=7$.
Here $\hat a_i$ and $a_i$ are the predicted and reference answers.
Supp.~Sec.~\ref{sec:supp_training_diagnostics} reports subtype scores.
A shared parser
extracts choices using only model responses and public options. Outputs
without a resolved choice count as incorrect. Evidence scoring checks
instance identity, witness, and time. Joint scores multiply answer
correctness by evidence credit. Assigning unresolved evidence scores zero
or one gives lower and upper bounds. Bootstrap intervals group related
questions by core task or source video.

\section{\agentname: Spatial-memory agent}
\label{sec:agent}

\subsection{Retrieval, visual inspection, and reasoning}
\begin{wrapfigure}[25]{R}{0.45\textwidth}
\centering
\includegraphics[width=\linewidth]{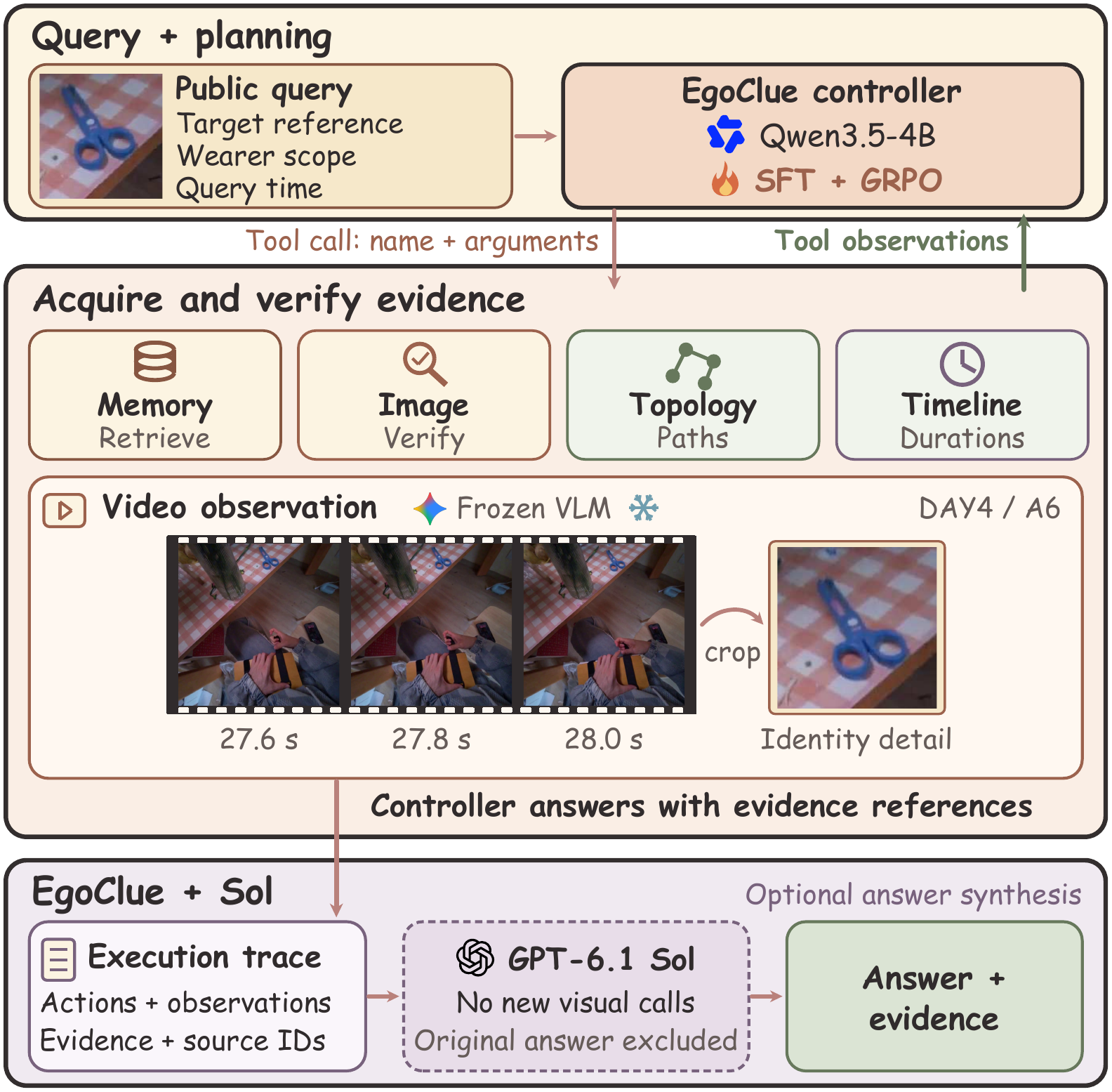}
\caption{\textbf{\agentname: finding, verifying, and linking evidence.}
The trained 4B controller selects tools and video intervals, then refines
its search from returned observations. The optional \agentname{} + Sol stage
synthesizes an answer from the trace without new visual calls.}
\label{fig:agent}
\end{wrapfigure}

Answering \bench\ questions involves finding candidate sightings,
verifying their identity, and relating them in space and time. Following
tool-augmented reasoning~\cite{react,toolformer,toolllm}, \agentname\ uses
Qwen3.5-4B~\cite{qwen35} as its controller to decide where to look and how
to combine observations from five tools in \cref{fig:agent}. \textbf{Memory} retrieves intervals from text summaries,
\textbf{Video} inspects intervals for targets, rooms, and events, and
\textbf{Image} compares references with individual frames.
\textbf{Topology} computes shortest paths between identified rooms, and
\textbf{Timeline} computes durations from observed boundaries.
Image and Video share a fixed visual backend. Memory queries a text model,
while Topology and Timeline compute relations deterministically.
Each returns text for the controller's next decision.

Memory uses a text model to select candidate intervals from prebuilt captions,
transcripts, and coarse annotation categories. Video inspection starts at 1 FPS, with local refinement guided by returned
observations. For visit duration, Video identifies entry and exit events and
Timeline computes their separation. For witness queries, the controller
verifies each candidate sighting and compares their times.

\agentname\ uses SFT+GRPO by default. In the last stage of \cref{fig:agent},
optional GPT-6.1 Sol synthesis reads the public question and trace preceding the controller's final answer, producing
a choice and evidence references without additional visual calls.
Supp.~Sec.~\ref{sec:supp_synthesis} specifies its inputs and cost.

\par\WFclear

\begin{table*}[!t]
\caption{\textbf{Spatial memory and reasoning across model and system families.}
Accuracy in \% on 1,320 week-long and 787 cross-environment questions.
Direct VLMs use public references and up to 32 frames. Agents share the budget
in \cref{tab:supp_access_budget}. Macro/Micro average abilities/questions.
Bold/underline mark group-best/runner-up aggregates.
$\Delta$ reports Micro differences in percentage points relative to direct-group
leaders, 4B base for controllers, or SFT+GRPO for synthesis.}
\label{tab:main}
\centering\small
\setlength{\tabcolsep}{3pt}
\renewcommand{\arraystretch}{1.0}
\begin{tabular}{@{}lrrrrrrrrrrr@{}}
\toprule
\multirow{2}{*}{Model or system} & \multicolumn{8}{c}{Week-long answer accuracy} &
\multicolumn{2}{c}{Week-long} & Cross-env. \\
\cmidrule(lr){2-9}\cmidrule(lr){10-11}\cmidrule(lr){12-12}
& Q01 & Q02 & Q03 & Q04 & Q05 & Q06 & Q07 & \textbf{Macro} & \textbf{Micro} & $\Delta$ & Micro \\
\midrule
Uniform random, expected & 33.33 & 25.00 & 25.00 & 25.00 & 25.00 & 33.33 & 25.00 & 27.38 & 26.52 & -- & 27.12 \\
Public-text ridge & 42.50 & 28.33 & 27.50 & 27.50 & 25.00 & 43.33 & 30.67 & 32.12 & 31.59 & -- & 27.06 \\
\midrule
\rowcolor{resultgroup}\multicolumn{12}{@{}l}{\textit{Direct open-weight VLMs}} \\
\mlogo{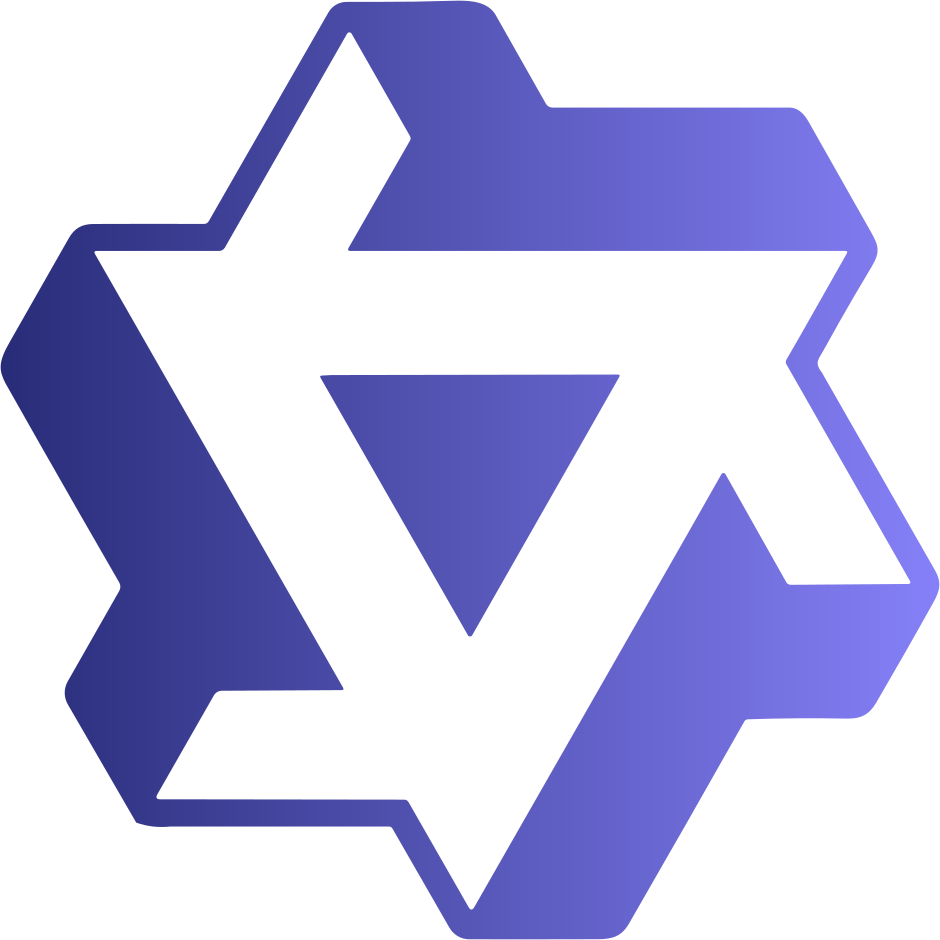}Qwen3.8-27B~\cite{qwen38} & 38.33 & 50.00 & 27.50 & 57.50 & 52.50 & 62.50 & 45.67 & \textbf{47.71} & \textbf{46.97} & 0.00 & \textbf{58.07} \\
\mlogo{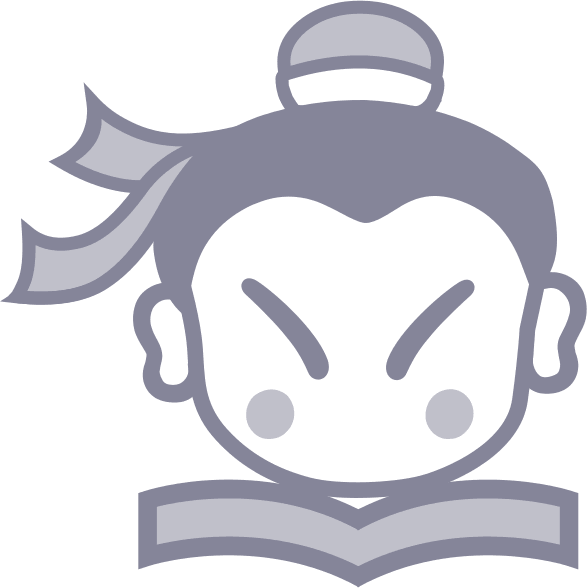}VideoChat-Flash-7B~\cite{videochatflash} & 62.50 & 26.67 & 25.00 & 32.50 & 63.33 & 30.83 & 43.83 & 40.67 & \underline{41.82} & \loss{-5.15} & 38.37 \\
\mlogo{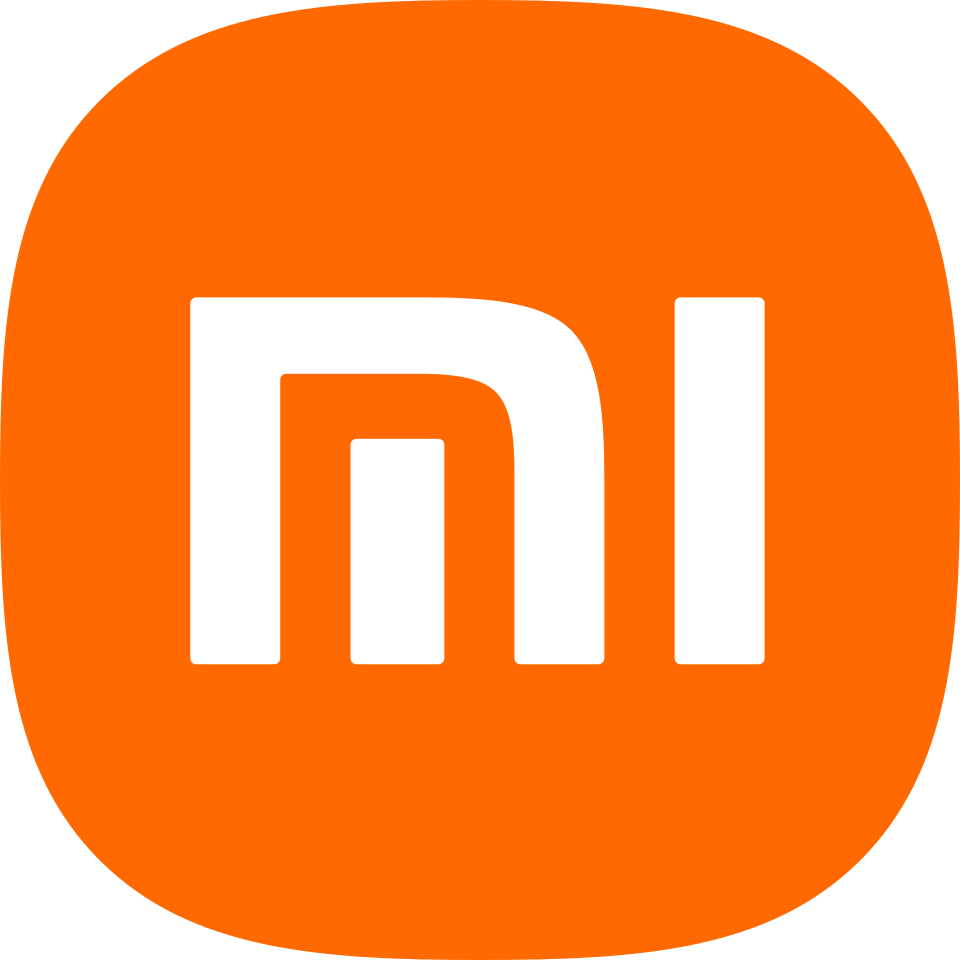}MiMo-VL-7B-RL~\cite{mimovl} & 35.00 & 43.33 & 17.50 & 37.50 & 47.50 & 36.67 & 46.00 & 37.64 & 40.68 & \loss{-6.29} & 46.25 \\
\mlogo{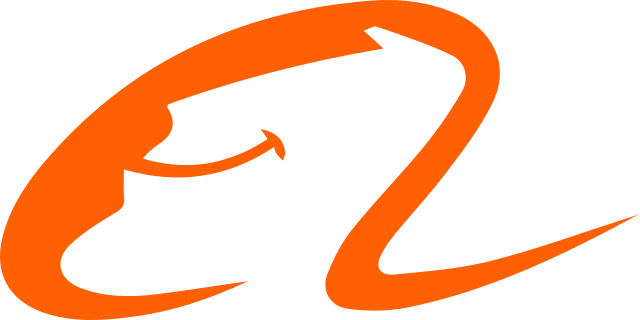}VideoLLaMA3-7B~\cite{videollama3} & 49.17 & 45.83 & 20.83 & 38.33 & 52.50 & 48.33 & 37.50 & \underline{41.79} & 40.23 & \loss{-6.74} & 47.78 \\
\mlogo{internlm-color}InternVideo3-8B~\cite{internvideo3} & 37.50 & 57.50 & 23.33 & 40.83 & 49.17 & 33.33 & 39.33 & 40.14 & 39.85 & \loss{-7.12} & 44.09 \\
\mlogo{internlm-color}InternVL3.5-8B~\cite{internvl35} & 43.33 & 44.17 & 22.50 & 43.33 & 59.17 & 35.00 & 37.83 & 40.76 & 39.70 & \loss{-7.27} & 47.90 \\
\mlogo{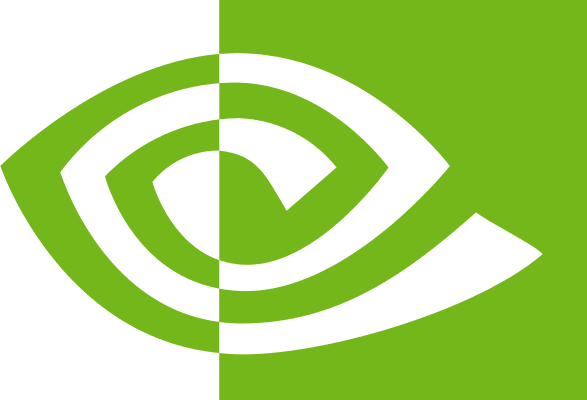}LongVILA-R1-7B~\cite{longvilar1} & 47.50 & 32.50 & 15.00 & 31.67 & 31.67 & 39.17 & 24.17 & 31.67 & 28.94 & \loss{-18.03} & 28.46 \\
\mlogo{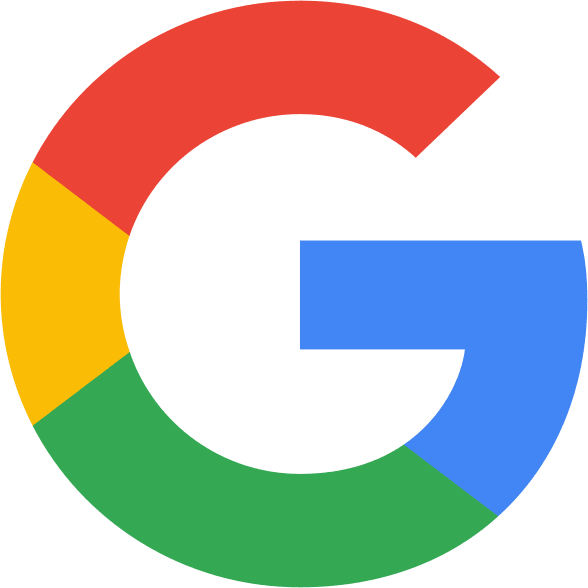}Gemma 4 12B~\cite{gemma4} & 25.83 & 50.00 & 5.83 & 29.17 & 14.17 & 36.67 & 30.83 & 27.50 & 28.71 & \loss{-18.26} & \underline{49.81} \\
\mlogo{qwen-color}Qwen3.5-4B~\cite{qwen35} & 29.17 & 35.83 & 25.83 & 30.83 & 10.00 & 34.17 & 28.67 & 27.79 & 28.11 & \loss{-18.86} & 41.93 \\
\midrule
\rowcolor{resultgroup}\multicolumn{12}{@{}l}{\textit{Direct proprietary VLMs}} \\
\mlogo{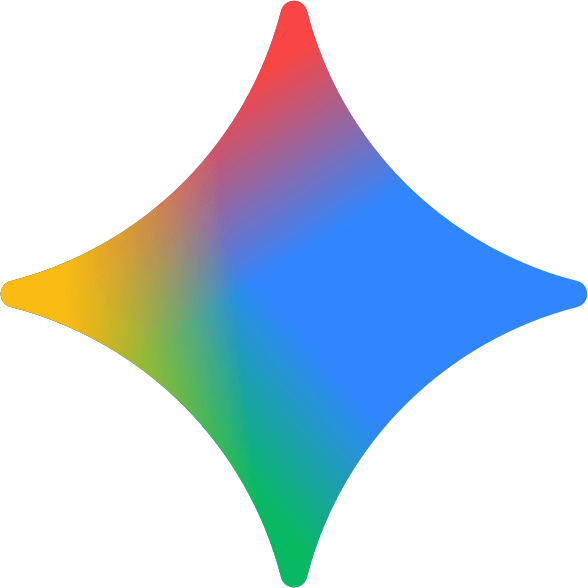}Gemini 4 Argon~\cite{gemini4argon} & 74.17 & 83.33 & 34.17 & 77.50 & 85.83 & 84.17 & 74.00 & \textbf{73.31} & \textbf{73.56} & 0.00 & \textbf{86.28} \\
\mlogo{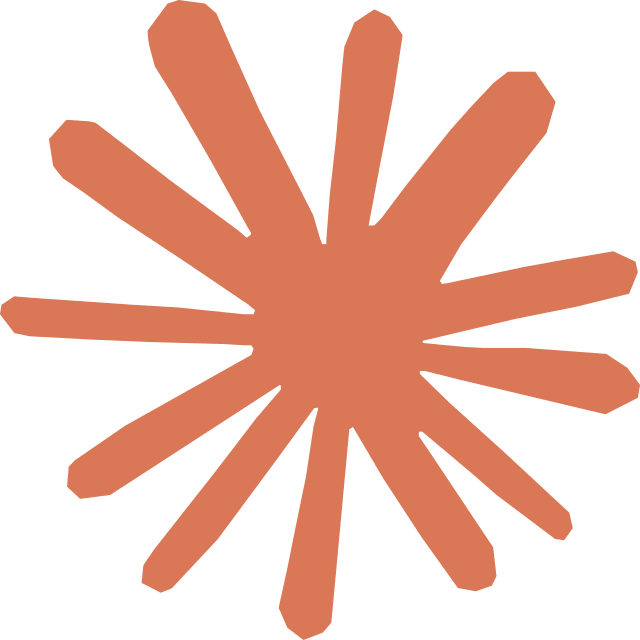}Claude Opus 5.5~\cite{claudeopus55} & 70.00 & 80.00 & 29.17 & 73.33 & 91.67 & 80.00 & 69.17 & \underline{70.48} & \underline{70.00} & \loss{-3.56} & \underline{80.81} \\
\mlogo{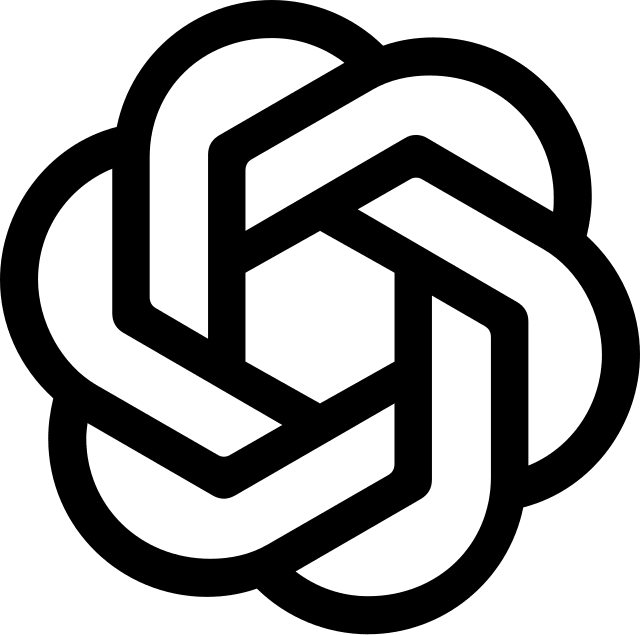}GPT-6.1 Sol~\cite{gpt61sol} & 62.50 & 77.50 & 28.33 & 61.67 & 74.17 & 65.00 & 55.50 & 60.67 & 58.79 & \loss{-14.77} & 67.60 \\
\midrule
\rowcolor{resultgroup}\multicolumn{12}{@{}l}{\textit{Prompted shared-tool controllers: $\Delta$ relative to Qwen3.5-4B base}} \\
\mlogo{qwen-color}Qwen3.5-4B~\cite{qwen35} (base) & 43.33 & 49.17 & 3.33 & 20.00 & 18.33 & 6.67 & 23.83 & 23.52 & 23.64 & 0.00 & 22.24 \\
\mlogo{openai}GPT-5.3 Chat~\cite{gpt53chat} & 69.17 & 66.67 & 27.50 & 52.50 & 62.50 & 80.83 & 72.83 & \textbf{61.71} & \textbf{65.76} & \gain{+42.12} & \underline{67.85} \\
\mlogo{gemini-color}Gemini 4 Argon~\cite{gemini4argon} & 63.33 & 60.00 & 7.50 & 35.00 & 54.17 & 52.50 & 47.00 & \underline{45.64} & \underline{46.14} & \gain{+22.50} & \textbf{71.54} \\
\mlogo{gemini-color}Gemini 3.5 Flash-Lite~\cite{gemini35flashlite} & 37.50 & 31.67 & 16.67 & 36.67 & 68.33 & 57.50 & 33.00 & 40.19 & 37.58 & \gain{+13.94} & 52.60 \\
\mlogo{claude-color}Claude Haiku 5.5~\cite{claudehaiku55} & 40.83 & 62.50 & 5.00 & 26.67 & 34.17 & 44.17 & 28.17 & 34.50 & 32.20 & \gain{+8.56} & 49.56 \\
\midrule
\rowcolor{resultgroup}\multicolumn{12}{@{}l}{\textit{Trained \agentname\ controllers: $\Delta$ relative to Qwen3.5-4B base}} \\
\mlogo{egowitness-glasses}\agentname-SFT & 64.17 & 61.67 & 15.83 & 32.50 & 55.00 & 80.00 & 67.00 & 53.74 & 58.56 & \gain{+34.92} & 68.11 \\
\rowcolor{resultours}\mlogo{egowitness-glasses}\agentname-SFT+GRPO & 71.67 & 68.33 & 38.33 & 53.33 & 73.33 & 87.50 & 81.33 & \textbf{67.69} & \textbf{72.65} & \gain{+49.02} & \textbf{78.40} \\
\midrule
\rowcolor{resultgroup}\multicolumn{12}{@{}l}{\textit{Evidence-synthesis system: no additional visual calls}} \\
\rowcolor{resultours}\mlogo{egowitness-glasses}\agentname{} + Sol~\cite{gpt61sol} & 72.50 & 67.50 & 44.17 & 55.83 & 77.50 & 87.50 & 83.67 & \textbf{69.81} & \textbf{74.85} & \gain{+2.20} & \textbf{80.56} \\
\bottomrule
\end{tabular}
\vspace{-3pt}
\end{table*}

\begin{table*}[!t]
\caption{\textbf{Individual tool removal from \agentname.} Left: all 1,320 test questions.
Right: questions on which Full invokes the removed tool, with Full and Ablated
counting correct answers. Signed changes are percentage points
relative to the Full evaluation results. Rows are ordered by overall
loss. Supp.~Sec.~\ref{sec:supp_tool_removal} details the paired comparison.}
\label{tab:ablation}
\centering\small
\setlength{\tabcolsep}{5pt}
\renewcommand{\arraystretch}{1.10}
\begin{tabularx}{\textwidth}{@{}l*{7}{C}@{}}
\toprule
\multirow{2}{*}{Configuration} & \multicolumn{3}{c}{All test questions} &
\multicolumn{4}{c}{Full-invocation subset} \\
\cmidrule(lr){2-4}\cmidrule(l){5-8}
& Correct & Accuracy, \% & $\Delta$ & Questions & Full & Ablated & $\Delta$ \\
\midrule
\rowcolor{resultours}Full & 959 & \textbf{72.65} & 0.00 & & & & \\
Without Video & 620 & 46.97 & \loss{-25.68} & 971 & 688 & 369 & \loss{-32.85} \\
Without Timeline & 845 & 64.02 & \loss{-8.64} & 178 & 151 & 103 & \loss{-26.97} \\
Without Topology & 881 & 66.74 & \loss{-5.91} & 107 & 83 & 56 & \loss{-25.23} \\
Without Memory & 900 & 68.18 & \loss{-4.47} & 111 & 44 & 31 & \loss{-11.71} \\
Without Image & 907 & 68.71 & \loss{-3.94} & 458 & 322 & 312 & \loss{-2.18} \\
\bottomrule
\end{tabularx}
\end{table*}

\begin{figure*}[!t]
\centering
\includegraphics[width=\textwidth]{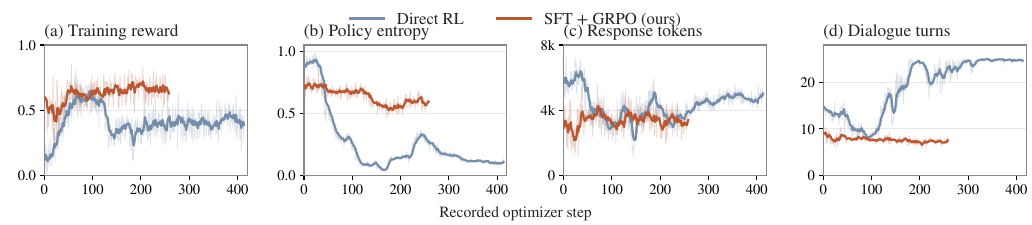}
\caption{\textbf{\agentname\ training dynamics.} Thin traces show raw scalars and
thick traces show EMA with coefficient 0.2.
Supp.~Sec.~\ref{sec:supp_optimization} details both training schedules.}
\label{fig:posttraining}
\end{figure*}

\subsection{Two-stage training}
To learn when to call which tool and when to answer, the controller is trained
in two stages. During SFT, the controller learns to select tools and answer
from teacher trajectories. GRPO then improves it with answer and evidence
rewards on its own rollouts.
Supp.~Sec.~\ref{sec:supp_optimization} details both stages.

GRPO samples four rollouts per question and masks tool observations from
the policy loss.
For a trajectory $\tau$, the reward combines answer
correctness $A$, task-specific evidence credit $E$, and format validity $F$:
\begin{equation}
 r(\tau)=
 \begin{cases}
 0.5A+0.5E+0.1F, & \text{evidence-scored tasks},\\
 A+0.1F, & \text{otherwise}.
 \end{cases}
 \label{eq:reward}
\end{equation}
Q02 and Q03 compute evidence credit from verified sighting anchors.
Submission validation checks format and references, while the training
evaluator supplies private answer and evidence scores.
Supp.~Sec.~\ref{sec:supp_grpo} defines the normalized group advantage
and clipped policy objective.

\section{Experiments}
\label{sec:experiments}
\subsection{Experimental setup}
\paragraph{Benchmark protocol and direct VLMs.}
All systems answer the same 1,320 questions, a type-balanced subset of the
1,533 test questions with 120 per detailed type. Metrics follow
\cref{sec:protocol}. We evaluate nine open-weight and three proprietary
VLMs, listed in \cref{tab:supp_model_registry}. These direct VLMs answer from public
references and up to 32 timestamped frames shared across models. Uniform
guessing and a public-text ridge classifier serve as baselines.

\paragraph{\agentname\ controllers.}
To isolate controller learning, Base, SFT, and SFT+GRPO share Gemini 3.6
Flash perception, GPT-5.3 Chat retrieval, limits of 12 decisions and
11 tool calls, and greedy decoding. SFT trains for 0.1 epoch on 1,100
teacher dialogues with a learning rate of $10^{-5}$. GRPO then trains on 4,650
other questions with four rollouts each and a learning rate of
$10^{-6}$ for 259 steps. GPT-5.3 Chat, Gemini 4 Argon, Gemini 3.5
Flash-Lite, and Claude Haiku 5.5 run as zero-shot controllers in the same
setting.

\subsection{Main results}
\paragraph{Instance-level memory remains challenging.}
\cref{tab:main} groups direct VLMs, prompted shared-tool controllers, trained
\agentname\ controllers, and evidence synthesis, and reports accuracy per
ability, the Macro and Micro aggregates, and cross-environment Micro.
Direct VLMs show a large gap between model families. Proprietary VLMs
reach 58.79\% to 73.56\% Micro accuracy, whereas open-weight VLMs reach
28.11\% to 46.97\%. The public-text ridge classifier reaches 31.59\%, only
5.07 points above uniform guessing, so most answers require visual evidence.
Accuracy also varies by ability. Room path Q05 is one of the two easiest
abilities for all three proprietary models, reaching up to 91.67\%.
Latest-witness Q03 is the hardest type for eleven of the twelve direct models,
and even Gemini reaches only 34.17\% from uniformly sampled frames. In \cref{tab:supp_error_patterns},
the proprietary models omit positive windows in 56\% of their Q04 errors
and undercount visits in 77\% of Q07E errors, while 89\% of their Q02
errors credit a wearer who did not see the target. These patterns suggest
that current VLMs struggle mainly to retrieve every relevant sighting from a
long multi-wearer history and verify its identity.

\paragraph{Training drives the gains.}
In \cref{tab:main}, the prompted Base
controller reaches 23.64\%.
With the same tools, the strongest prompted API controller, GPT-5.3 Chat,
reaches 65.76\%, 6.89 points below \agentname-SFT+GRPO. SFT
raises accuracy to 58.56\%, and GRPO reaches 72.65\%. GRPO
improves all seven abilities over SFT. The largest subtype gains are
25.00 points for completed-visit duration, 22.50 for last-observer
attribution, and 20.83 for historical-window verification. These subtypes
require Timeline computations or repeated Video checks of candidate
sightings. \cref{fig:posttraining} tracks reward, policy entropy, response
tokens, and dialogue turns. SFT+GRPO keeps a higher reward
with about seven turns, whereas RL without SFT reaches a lower reward with
about 25 turns. Its policy entropy also drops to about 0.1 within 100 steps,
while SFT+GRPO stays near 0.6, so SFT keeps exploration alive for GRPO.
On 1,021 applicable questions, GRPO raises joint answer-evidence
correctness, and its lower bound exceeds the SFT upper bound in
\cref{tab:supp_joint_audit}.

\paragraph{Evidence synthesis improves accuracy.}
Sol synthesizes answers from the observations the controller retains. This raises \agentname's
accuracy from 72.65\% to 74.85\%, the highest week-long Micro in \cref{tab:main},
with 95\% intervals in \cref{tab:supp_main_ci}.
The same Sol model scores 58.79\% under direct access, so the gain comes
from the evidence the controller gathers.
Cross-environment evaluation tests the week-long-trained system on 787
questions from other environments. The cross-environment setting in
Supp.~Sec.~\ref{sec:supp_synthesis} uses only Image, Video, and Timeline.
With the same budget as all controllers, \agentname\ reaches 78.40\%, above
every prompted controller, and synthesis raises it to 80.56\%.

\subsection{Ablation and cost analysis}
To identify which tools matter, we remove each of the five tools individually.
Questions, controller, backends, and budget remain fixed.
Removing Video lowers accuracy by 25.68 points, and the other removals cost
3.94 to 8.64 points in \cref{tab:ablation}.
The right half of \cref{tab:ablation} keeps questions where the full agent
invokes the removed tool, and there the Topology, Timeline, and Video drops
reach 25.23, 26.97, and 32.85 points.
Paired core-group bootstrap intervals exclude zero.
Without Video, Q07 visit reasoning falls from 81.33\% to 37.17\% and
latest-witness Q03 from 38.33\% to 21.67\%, while object identity Q01
changes by only 1.67 points because Image still compares the two references.
Removing Topology lowers room
path Q05 from 73.33\% to 54.17\%, and removing Timeline lowers the duration
questions Q07B and Q07C from 82.08\% to 58.33\%.
SFT initialization also matters. Without SFT, RL from the base controller
learns the output format, with 99.24\% validity, but reaches 40.68\%,
31.97 points below SFT+GRPO. GRPO also reduces mean tool calls
from SFT's 2.67 to 2.31 per question.
Against the prompted shared-tool controllers, \agentname\ needs the
fewest turns, 3.30 versus 4.34 to 8.07, and the fewest proprietary-model
tokens, 38.1k versus 44.3k to 97.4k per question, since its controller runs
locally.
At list prices, it is also the cheapest system above 65\% accuracy in
Supp.~Sec.~\ref{sec:supp_cost}.

\section{Conclusion}
\label{sec:conclusion}
\bench\ evaluates instance-level spatial memory through 8,453 questions that
link object and room identity to past sightings across days and wearers.
Over one million reviewed masks and room timelines ground each answer in
verifiable observations. Twelve direct VLMs struggle most on identifying who saw an
object last. As a strong baseline, \agentname\
searches for and verifies evidence with five tools. Removing Video lowers
accuracy most, by 25.68 points, highlighting the importance of revisiting
recorded intervals. Two-stage training
raises its 4B controller from 23.64\% to 72.65\%. Evidence synthesis
reaches 74.85\% on week-long questions and 80.56\% in unseen environments.

\FloatBarrier
\section*{LLM Usage Statement}
OpenAI Codex assisted with manuscript drafting and editing, source inspection,
code and LaTeX preparation, and figure layout. GPT-6.1 Sol assisted with
first-pass QA review and evidence-only answer synthesis. The research
pipeline also uses language and vision-language models for candidate
annotation review, question wording, teacher tool-use trajectories, and
memory construction. Qwen3.5-4B is the trained controller, with Gemini 3.6
Flash and GPT-5.3 Chat providing visual and retrieval services.
The method and supplementary material specify model roles and settings.
Trained experts verify the QA items after model-assisted review.
The authors retain responsibility for the scientific claims, annotations,
code, figures, and results.

\begingroup
\renewcommand{\bibfont}{\fontsize{8.5}{10.2}\selectfont}
\renewcommand{\bibsection}{}
\setlength{\columnsep}{18pt}
\begin{multicols}{2}[\section*{References}]
\bibliographystyle{ieeenat_fullname}\bibliography{literature}
\end{multicols}
\endgroup
\clearpage
\setcounter{page}{1}
\maketitlesupplementarycontents
\raggedbottom
\appendix
\renewcommand{\thepage}{S\arabic{page}}
\setcounter{table}{0}
\setcounter{figure}{0}
\renewcommand{\thetable}{S\arabic{table}}
\renewcommand{\thefigure}{S\arabic{figure}}
\makeatletter
\setlength{\@fptop}{0pt}
\setlength{\@fpsep}{12pt}
\setlength{\@fpbot}{0pt plus 1fil}
\setlength{\@dblfptop}{0pt}
\setlength{\@dblfpsep}{12pt}
\setlength{\@dblfpbot}{0pt plus 1fil}
\makeatother
\section{Data sources and annotation integration}
\label{sec:supp_sources}
\cref{tab:supp_source_roles} distinguishes original recordings from the
annotation resources applied to them. EgoLife provides the week-long
recordings and house assets. The short-horizon resources share Ego4D
source-video identifiers, which we use to align their annotations and
count each original video once.

\subsection{EgoLife recordings and new annotations}
\label{sec:supp_egolife_source}
We add room intervals, instance associations, and mask observations to
EgoLife~\cite{egolife}. Room IDs are shared across six wearers and seven days,
allowing an observation to be linked to both its target and its observing
wearer. \cref{sec:supp-data} specifies the annotation operations, acceptance
rules, and exported records. Question construction is described separately
in \cref{sec:supp_questions}.

\begin{table*}[!htbp]
\centering\small
\caption{\textbf{Source resources and their integration.}
Rows distinguish original media from inherited annotations.
Sampling rates describe the annotations retained in EgoWitness.
The shared Ego4D videos are counted once across resources.}
\label{tab:supp_source_roles}
\setlength{\tabcolsep}{4pt}
\renewcommand{\arraystretch}{1.08}
\begin{tabularx}{\textwidth}{@{}lCCY@{}}
\toprule
\textbf{Resource} & \multicolumn{2}{c}{\textbf{Reused input}} & \textbf{Integration in EgoWitness} \\
\cmidrule(lr){2-3}
 & Layer & Scope / sampling & \\
\midrule
\rowcolor{tableband}\multicolumn{4}{@{}l}{\textit{Week-long recordings}} \\
EgoLife~\cite{egolife} & Videos, house assets & Six wearers, seven days &
New room intervals, instance links, and masks \\
\midrule
\rowcolor{tableband}\multicolumn{4}{@{}l}{\textit{Short-horizon recordings and annotations}} \\
Ego4D~\cite{ego4d} & Videos & 43 original videos &
Shared video IDs and time axes \\
EgoMask / EgoTracks~\cite{egomask,egotracks} & Masks, track lineage & 1\,Hz retained &
Boxes and local tracks aligned to rooms \\
RefEgo~\cite{refego} & Boxes, expressions & 2\,Hz, width 640 &
Target deduplication and coordinate scaling \\
EgoEnv~\cite{egoenv} & Room intervals, references & 400 source-local entries &
Observer-room context for object sightings \\
\bottomrule
\end{tabularx}
\par\vspace{2pt}\raggedright\footnotesize
Media completion uses 15 full-duration 224p versions from the
\texttt{MickeyLLG/ego\_plan} repository~\cite{egoplan_media}.
These copies belong to the same 43-video inventory.
\end{table*}

\subsection{Ego4D-derived annotations and alignment}
\label{sec:supp_crossroom}

\paragraph{Matching annotation sources.}
We associate object annotations from EgoMask and RefEgo with observer-room
intervals from EgoEnv using source-video identifiers and aligned time
intervals~\cite{egomask,refego,egoenv}.
For each object observation, the join finds the compatible room interval
on the same source-video timeline:
\begin{equation}
\begin{aligned}
\mathcal D_{\mathrm{short}} ={}&
(\text{EgoMask}\Join\text{EgoEnv})\\
&{}\cup(\text{RefEgo}\Join\text{EgoEnv}).
\end{aligned}
\label{eq:supp_source_join}
\end{equation}
Here $\Join$ denotes this temporal association, and the union combines
the two aligned object sources while retaining their provenance.
The selected subset contains 43 original videos, 83 source
clips, 170 source-local tracks, and 172 cross-room observation candidates.
\cref{tab:supp_crossroom_sources} lists the per-source counts.
Four original videos occur in both object sources. We preserve each source's
track IDs and record cross-room observation support as a separate attribute.
EgoTracks~\cite{egotracks} provides EgoMask's upstream tracking annotations
over Ego4D media~\cite{ego4d}.

\begin{table}[!htbp]
\centering\small
\caption{\textbf{Selected short-horizon data.} Counts from EgoMask-Train
and RefEgo. Totals deduplicate videos. Tracks retain source-local scope.
``Matched'' denotes an unambiguous association with the camera wearer's
occupied room.}
\label{tab:supp_crossroom_sources}
\setlength{\tabcolsep}{3pt}
\begin{tabularx}{\linewidth}{@{}l*{3}{>{\raggedleft\arraybackslash}X}@{}}
\toprule
\textbf{Quantity} & \textbf{EgoMask} & \textbf{RefEgo} & \textbf{Total} \\
\midrule
\rowcolor{tableband}\multicolumn{4}{@{}l}{\textit{Source media}} \\
Original videos & 37 & 10 & 43 \\
Source clips & 70 & 13 & 83 \\
\rowcolor{tableband}\multicolumn{4}{@{}l}{\textit{Tracks and object observations}} \\
Source-local tracks & 157 & 13 & 170 \\
Observation candidates & 159 & 13 & 172 \\
Visible boxes & 13,988 & 282 & 14,270 \\
\rowcolor{resultours}Matched visible boxes & 13,402 & 253 & \textbf{13,655} \\
Retained annotation rate, Hz & 1 & 2 & mixed \\
\bottomrule
\end{tabularx}
\end{table}

\paragraph{Media duration.}
The 43 full videos span 23.46 hours, equivalent to 84,441.037 seconds,
and occupy 17.94 GB. The manifest records each video's resolution. EgoLife reports approximately
300 hours of recordings~\cite{egolife}, and this collection adds approximately
24 hours, giving the approximately 324-hour source scale.

Visible-box and room-matched-box counts
in \cref{tab:supp_crossroom_sources} specify the usable object observations.

\paragraph{Track selection.}
Within these 43 videos, we select 170 tracks with unambiguous EgoEnv
room matches. Cross-room support describes the observer's
room contexts, and a stationary target can be seen from different rooms.

\paragraph{Time and geometry alignment.}
EgoMask retains 1 Hz observations although its frame filenames use a
5 Hz index: the original-video time is the clip offset plus the filename
index divided by five. RefEgo retains its nominal 2 Hz convention, with
time $(i+j-1)/2$ seconds, where $i$ and $j$ are its image ID and
image-clip ID. RefEgo
boxes refer to annotation images of width 640, so coordinates are
normalized using the source annotation dimensions and rescaled to the
actual media dimensions. Duplicate expressions for the same RefEgo target
are deduplicated. The original aspect ratio is preserved.

\paragraph{Room semantics and identity scope.}
The room-reference catalog contains 400 source-local entries across 81 scopes
and 18 room types. Each room instance is scoped by source video, EgoEnv clip,
room label, and instance index. The temporal join assigns the observer's
occupied room, which can differ from the visible target's room.
Cross-room candidates require observations in at least two room contexts.
The join excludes overlapping intervals and a 0.5-second band around room
boundaries, and connectivity records the room transitions observed in the video.

\paragraph{Short-horizon questions.}
The short-horizon component contributes 787 reviewed questions from 40
videos to the 8,453-question benchmark. EgoWitness reserves all 787 for
evaluation and excludes their environments from SFT and GRPO. They cover
120 object-identity questions in Q01, 169 object-history questions in Q04,
80 room-identity questions in Q06, and 418 visit-reasoning questions
across Q07B to Q07E.
The 418 visit-reasoning questions carry 676 scoring-evidence references over
316 unique intervals of 2 to 4 seconds. Identity and object-history questions
are scored from public references and candidate windows. Each question retains the
source-video timeline and source-local room identity used to derive its
answer. Video, participant-group, and core-task identifiers determine
split membership. Deterministic checks verify answer consistency against the
selected room intervals.

\section{Room, object, and landmark annotation}
\label{sec:supp-data}
Ten annotators produced the room timelines and the movable-object and
landmark annotations.
\cref{fig:supp_tools} shows their two linked workspaces: room-interval
editing and instance association. We describe the editing operations and
export rules first, followed by identity checks, annotation fields, and
cross-wearer coverage.

\subsection{Room segment annotation}
\label{sec:supp_room_annotation}
\begin{figure*}[!htbp]
\centering
\includegraphics[width=0.88\textwidth,keepaspectratio]{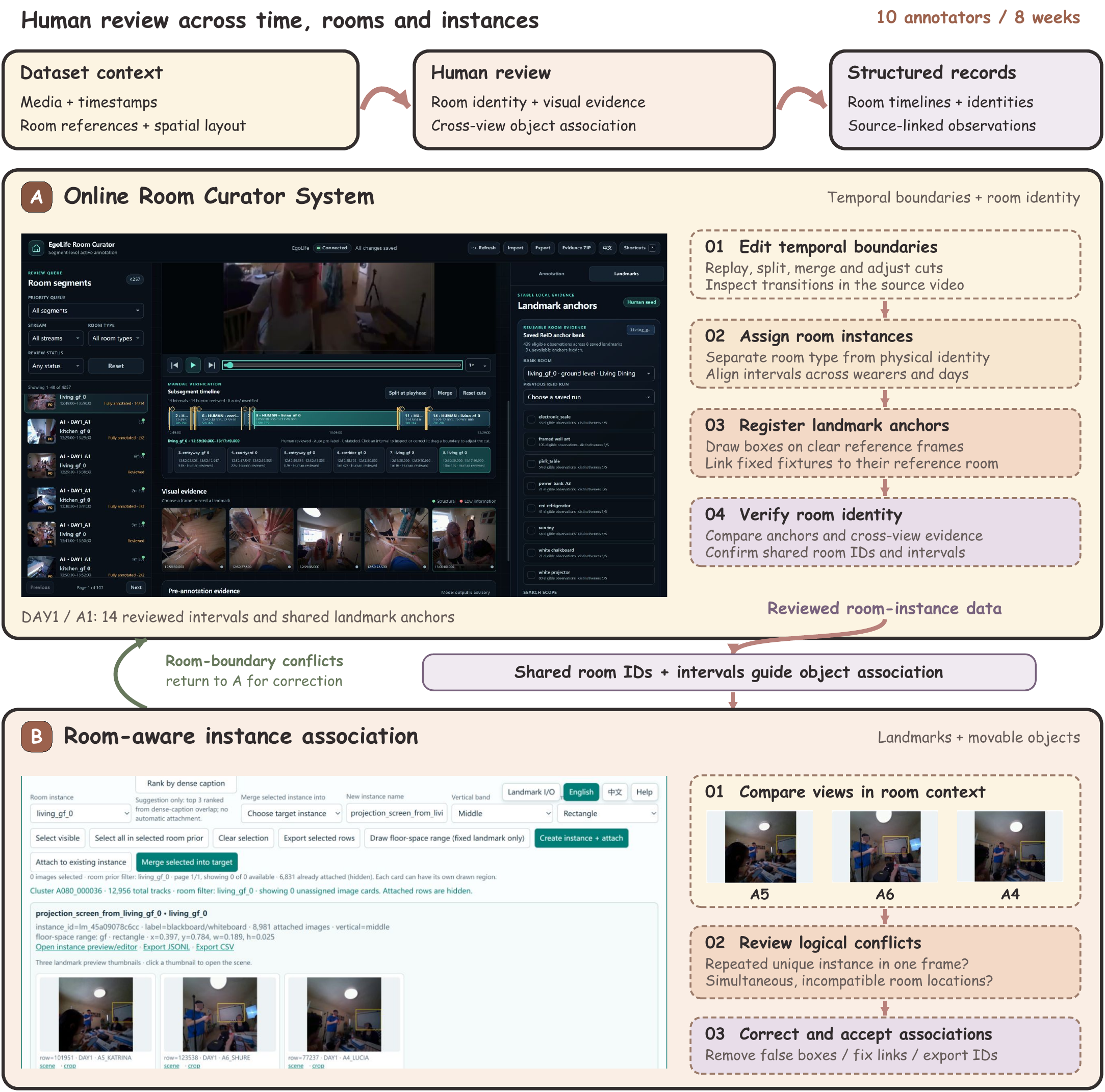}
\caption{\textbf{Annotation workspaces and review operations.}
\textbf{A.} The Online Room Curator System supports boundary editing,
room-instance assignment, landmark registration, and visual verification.
The DAY1/A1 example contains 14 reviewed intervals. Shared room IDs and
aligned intervals supply context to \textbf{B}, where annotators compare
landmark and movable-object instances, inspect logical conflicts, and correct
associations before export. A5, A6, and A4 show one shared instance.
Room-boundary conflicts return to A for interval correction.}
\label{fig:supp_tools}
\end{figure*}

In \cref{fig:supp_tools}A, annotators replay segments, adjust transition
boundaries, and assign room instances in the Online Room Curator System.
The interface organizes work by wearer, day, and review status.
Annotators inspect frames around a transition and split, merge, or adjust
its intervals. They assign room type and room instance
separately and record the decision.
The room label describes the camera wearer's occupied space, including
when the camera looks through a doorway into another room.

\renewcommand{\dblfloatpagefraction}{0.8}
\renewcommand{\dbltopfraction}{0.9}

\subsubsection{Spatial references and room identities}
\label{sec:supp_spatial_references}
\begin{figure*}[!htbp]
\centering
\includegraphics[width=0.92\textwidth,keepaspectratio]{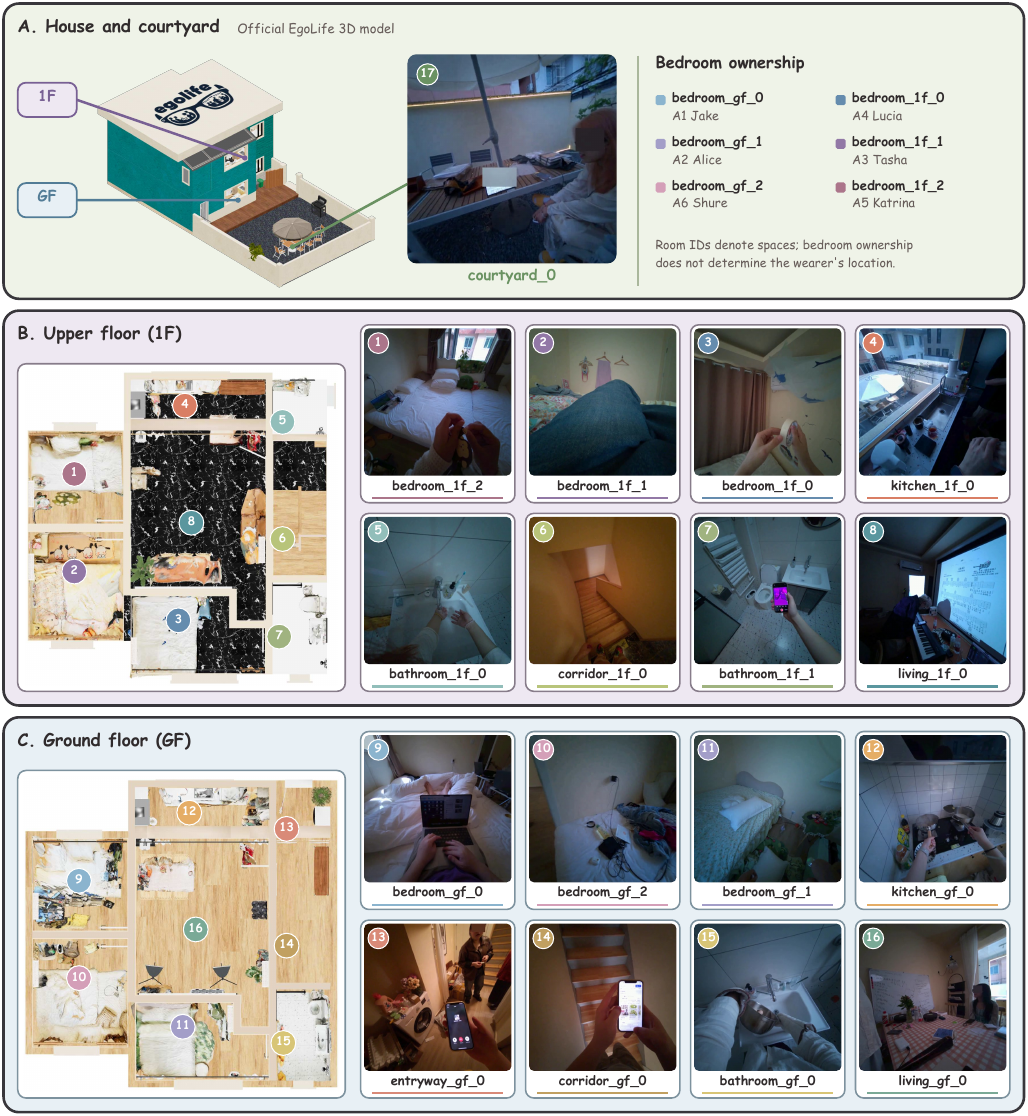}
\caption{\textbf{Spatial reference for the week-long environment.}
\textbf{A.} The official EgoLife house model, a courtyard observation, and
bedroom ownership. \textbf{B and C.} Upper- and ground-floor plans paired
with real video observations of all 16 indoor room instances; the courtyard
is the seventeenth named zone. Numbered pins match the photo cards, which
retain the room IDs and colors used in \cref{fig:supp_room_timeline}.
Geometry and textures come from the official EgoLife assets~\cite{egolife}.
The pins are schematic labels.}
\label{fig:supp_house_layout}
\end{figure*}

\cref{fig:supp_house_layout}A locates the courtyard and bedroom assignments
within the shared home. Panels B and C pair the upper- and ground-floor
plans with video observations of the corresponding rooms. Together, the
panels identify all 17 room and zone IDs used in annotation. The exterior
model and floor plans use the official EgoLife house assets~\cite{egolife}.
GF denotes the ground floor and 1F the
upper floor. The courtyard belongs to the home and remains a named zone.
Annotators use these references to distinguish adjacent rooms at doorways,
corridors, and stairs, then assign the occupied room from the video.

\subsubsection{Room timeline scope}
\label{sec:supp_room_scope}
\begin{figure*}[!htbp]
\centering
\includegraphics[width=\textwidth]{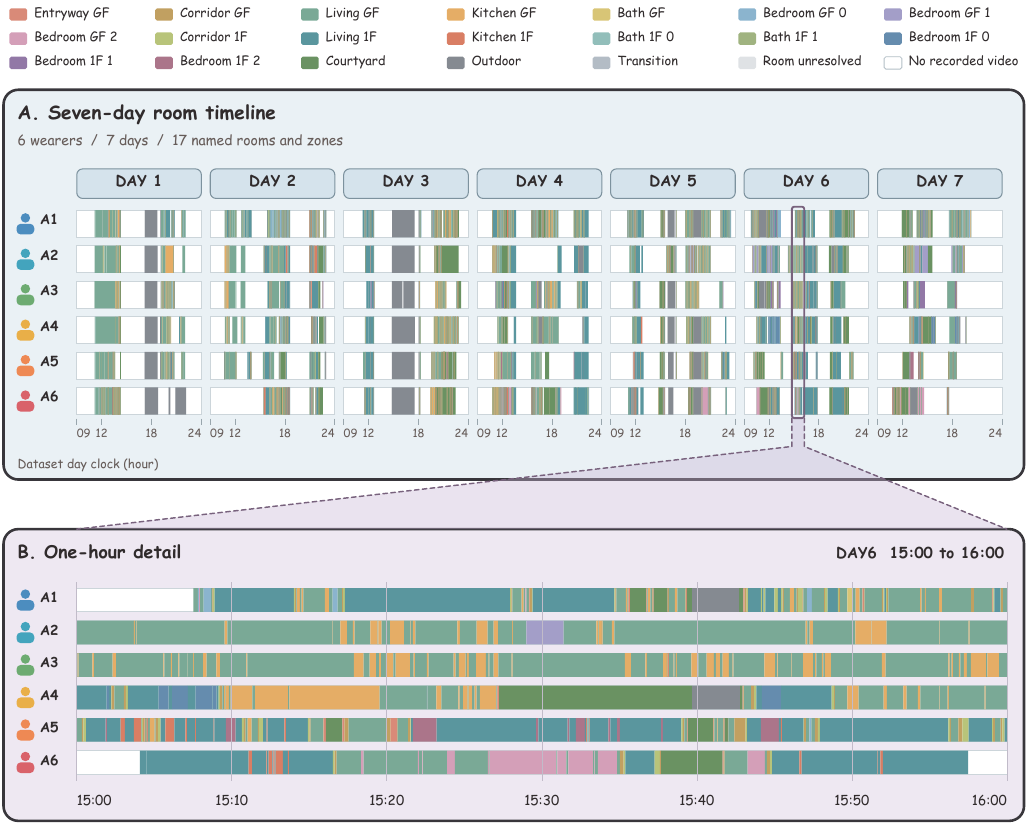}
\caption{\textbf{Room observations across wearers and days.}
\textbf{A.} Each row represents one wearer and each column one day, with
a common 09:00 to 24:00 day-clock axis. Colors identify 17 named rooms and
zones, with separate colors for outdoor, transition, and unresolved-room
states. White gaps indicate missing recordings.
\textbf{B.} The DAY6 window from 15:00 to 16:00, outlined in A, expands the room intervals
for all six wearers. Plotted intervals are clipped to the media manifest,
preserving their boundaries and recording gaps.}
\label{fig:supp_room_timeline}
\end{figure*}

\cref{fig:supp_room_timeline}A places the six wearers' room intervals on a
shared seven-day timeline. Panel B expands one hour on DAY6 to show their
different room observations at the same time. Shared colors identify rooms
and zones, while white gaps mark missing recordings.
The media manifest groups 32,001 source videos into these 42 wearer-day
streams. We compute recording coverage within each stream and sum durations
over wearers to obtain camera-hours.

\paragraph{Interval export and completion.}
The annotation store contains 4,408 source parent segments, 695 parent-review
records, and 17,076 saved subintervals. Saved subintervals replace their parent
annotation to avoid overlapping representations.
Otherwise, a completed parent review contributes one whole-parent interval.
This yields 4,257 retained parents and 17,269 intervals: 17,076 subintervals
plus 193 whole-parent reviews. Parent segments without recorded video are
excluded. The completion rule
requires saved, human-reviewed coverage, rejects overlapping child intervals,
and tolerates uncovered boundary gaps of at most 0.5 seconds.
The resulting intervals record 13,886 accepted and 3,383 corrected decisions.

\paragraph{Recording coverage.}
The retained intervals cover every recorded stream. The export
records the room-database snapshot separately from the object-layer versions.

\paragraph{Room IDs and occupancy states.}
The 17 room and zone IDs in \cref{fig:supp_house_layout} are shared across
recordings. Each ID identifies the same physical space. The courtyard is
an in-property zone with no assigned floor. Occupancy states are
stored separately: 16,964 intervals are inside a room, 295 are outdoors,
nine are transitions, and one is unknown. The occupancy state distinguishes
outdoor observations, transitions, and unknown states.

\par\noindent\begin{minipage}{\textwidth}

\subsection{Movable-object and landmark annotation}
\label{sec:supp_object_annotation}

\begin{minipage}[t]{0.47\textwidth}
\vspace{0pt}
In \cref{fig:supp_tools}B, accepted room intervals supply the context for
comparing object views with registered references. Annotators resolve
identity associations before exporting the reviewed records. The following
counts describe the reviewed prompts and their propagated mask observations,
and \cref{tab:assets} summarizes the accepted masks by layer.

\paragraph{Movable objects and identity.}
Filtering the 406,644 proposals from the 960-pixel detector retains 19,921
anchors in 2,403 class clips from 2,164 videos.
\end{minipage}\hfill
\begin{minipage}[t]{0.49\textwidth}
\vspace{0pt}
\captionsetup{type=table}
\caption{\textbf{EgoLife annotation inventory.} Primary counts are accepted
mask observations. Auxiliary layers are retained separately.}
\label{tab:assets}
\centering\scriptsize
\setlength{\tabcolsep}{4pt}
\renewcommand{\arraystretch}{1.08}
\begin{tabular*}{\linewidth}{@{\extracolsep{\fill}}lrr@{}}
\toprule & \textbf{Landmark} & \textbf{Movable} \\\midrule
Instance/target labels & 34 & 21 \\
Sampling rate, Hz & 1 & 5 \\
Mask observations & \textbf{795,319} & \textbf{238,756} \\
Annotated frames & 453,813 & 224,123 \\
Videos with accepted masks & 22,335 & 2,164 \\
Local SAM tracks & 90,261 & 2,408 \\\midrule
\multicolumn{2}{@{}l}{\textbf{Total accepted masks}} & \textbf{1,034,075} \\\midrule
Auxiliary detector boxes & 890,360 & -- \\
Original anchor boxes & -- & 19,921 \\\bottomrule
\end{tabular*}

\end{minipage}
\end{minipage}
\par
The filters use class-specific
confidence thresholds of 0.90 to 0.97 and persistence of three to five frames,
followed by duplicate and room checks in \cref{tab:supp_rules}.
Appearance clustering and Gemini~3.8 Flash~\cite{gemini38flash} organize
candidates for review. Annotators check several keyframes per track against
the reference views and surrounding video. SAM~3.1 propagates these checked
prompts forward and backward at 5\,Hz. Visual acceptance of the propagated
results yields 238,756 nonempty masks.
Each mask retains its reviewed target label, local track,
and quality flags.

\paragraph{Registered landmarks.}
A detector proposes landmark boxes, a verifier filters them, and SAM~3.1
propagates accepted prompts at 1\,Hz within local processing units, yielding
795,319 landmark masks. Each detector label indexes a registered instance whose reference
frames, room, floor, and zone support cross-segment comparison. Annotators
inspect conflicts with the registered floor in the source video.

\paragraph{Cross-view instance association.}
The A5, A6, and A4 views in \cref{fig:supp_tools}B show the same registered
instance from different wearers. The workspace groups candidate views by
room and time so annotators can compare their appearance and scene context.
Annotators compare each view with the reference frames and surrounding
video, then attach it to an existing instance, create an instance, or merge
entries. Exported associations retain the instance ID and source views.

All 34 released landmark names match source instance records with both
cross-wearer and cross-day observations, covering 70,286 source
observations. These counts describe the source association registry.
The propagated masks are counted separately in the released observation layer.

\par\WFclear
\Needspace{16\baselineskip}
\subsection{Context checks and visual acceptance}
\label{sec:supp_context_rules}
\begin{wraptable}{r}{0.49\textwidth}
\centering\scriptsize
\caption{\textbf{Context and quality rules by action.}
Counts describe boxes affected by each rule among the 406,644 movable
candidates. Held candidates proceed to visual review.}
\label{tab:supp_rules}
\setlength{\tabcolsep}{3pt}
\renewcommand{\arraystretch}{1.06}
\begin{tabularx}{\linewidth}{@{}Ylr@{}}
\toprule
\textbf{Rule} & \textbf{Action} & \textbf{Boxes} \\
\midrule
\rowcolor{tableband}\multicolumn{3}{@{}l}{\textbf{Movable candidates}\quad 406,644 boxes} \\
\multicolumn{3}{@{}l}{\footnotesize\textit{Filtering}} \\
Below class confidence & Reject & 357,116 \\
Below minimum persistence & Reject & 5,729 \\
Duplicate box & Suppress & 7,342 \\
Wearer outdoors & Exclude & 16,001 \\
\addlinespace[2pt]
\multicolumn{3}{@{}l}{\footnotesize\textit{Review holds}} \\
Unknown room or transition & Hold & 368 \\
Box quality & Hold & 148 \\
Uncertain detection & Hold & 19 \\
\bottomrule
\end{tabularx}
\end{wraptable}

The review loop in \cref{fig:supp_tools} links instance checks in panel B
to room-interval corrections in panel A. Annotators inspect a conflicting
association and correct either its identity or its room context.

\paragraph{Physical-instance consistency.}
We review identity assignments against two logical constraints.
First, each unique physical instance receives at most one accepted detection
per frame. Annotators remove erroneous or duplicate
boxes after inspecting the frame. Second, one physical instance cannot
occupy spatially disconnected rooms at the same time. We compare synchronized
observations and inspect conflicting locations in the source video.
Confirmed errors lead to box removal, identity reassignment, or room-interval
correction. Annotators then recheck the affected associations using the
corrected identity and room context.

These constraints apply to individually annotated movable instances and
registered landmarks.
Cross-room views through doorways or windows are checked against the object's
location, with each wearer's occupied room recorded separately.

\paragraph{Context rules and acceptance.}
\cref{tab:supp_rules} groups the movable-candidate rules into filtering and
review holds. A hold routes a candidate to visual review, while rejection or
suppression removes it. Simultaneous sightings assigned to
different floors trigger a joint review of object identity and room context.
Correcting room intervals resolves 29 one-second conflicts and updates the
room assignment of 338 boxes. Each accepted record stores the curator's
decision and its source observations. Model proposals retain separate
provenance fields.

\par\WFclear
\subsection{Unified observation records}
\label{sec:supp_representation}
Every accepted annotation from either source becomes one observation record,
linked to its source frame by a unique observation ID. Its other fields form
four groups: identifiers, locations, time and coordinates, and visibility and
review status.

\subsubsection{Category, label, instance, and track IDs}
Four identifiers describe the target at different levels:
\begin{itemize}
\item \textbf{Category ID:} one of 24 normalized semantic types, to which the
registry maps the 34 landmark and 21 movable labels. Color and material are
separate attributes.
\item \textbf{Release label:} one of the 55 original labels, kept through the
release class ID and a legacy identity-group ID.
\item \textbf{Physical-instance ID:} an optional key for one physical object,
linking its observations through reviewed labels and landmark associations.
The two bowl labels keep group-level identity.
\item \textbf{Track ID:} one SAM prompt trajectory, local to one run and
processing unit.
\end{itemize}
A shared category does not imply a shared object. Identity across videos,
wearers, and days comes from the reviewed associations in
\cref{sec:supp_object_annotation}.

\subsubsection{Observer, landmark, and object locations}
Three fields separate wearer and target locations:
\begin{itemize}
\item \textbf{Observer location:} the room or zone the camera wearer occupies,
from the reviewed room intervals.
\item \textbf{Registered landmark location:} the reference room, floor, and
zone from the landmark registry.
\item \textbf{Object location:} the target's location in this observation,
established independently of the observer location.
\end{itemize}
A wearer looking through a doorway or into the courtyard can occupy a
different space from the target, so a change in observer location alone does
not show that an object moved. Short-horizon room IDs stay local to their
source.

\subsubsection{Sampling, timestamps, and coordinates}
\label{sec:supp_sampling}
Landmark and movable masks keep nominal rates of 1\,Hz and 5\,Hz. Each record
stores its source-frame index, sampled-frame index, and video-relative
timestamp separately. A day-clock timestamp is meaningful only with its day
and wearer.

Temporal alignment maps the source-frame index and video-relative time
through the clip origin in the room manifest, and inherited day-clock values
are checked against this mapping before subsecond joins. The export keeps the
original timestamps and frame indices for boundary-level alignment.

EgoLife boxes give corner coordinates $(x_1,y_1,x_2,y_2)$ in a central
$976\times976$ crop of the $1408\times1408$ source frame; adding 216 to each
coordinate restores source coordinates. Masks keep their original compressed
COCO RLE in the same crop region. \cref{sec:supp_crossroom} gives the
EgoMask and RefEgo conversions.

\subsubsection{Visibility and review status}
An accepted mask shows that the target is visible at its timestamp. Frames
without an accepted mask are not treated as evidence of absence, and
confirming absence requires interval review. ``Last seen,'' visit-count, duration, and absence
questions therefore pair accepted observations with review of the whole
search interval, as \cref{sec:supp_questions} describes.

The primary object layers contain 795,319 landmark and 238,756 movable masks.
Human acceptance, the mask-generation model, and the filtering history are
stored as separate fields. Original anchors and historical detections are
auxiliary records linked to the 1,034,075 primary observations. Export checks
verify RLE decoding, foreground area, source-record identity, and record
consistency.

\subsection{Recurrence across days, wearers, and rooms}
\label{sec:supp_identity_recurrence}
\begin{figure*}[!htbp]
\centering
\includegraphics[width=\textwidth]{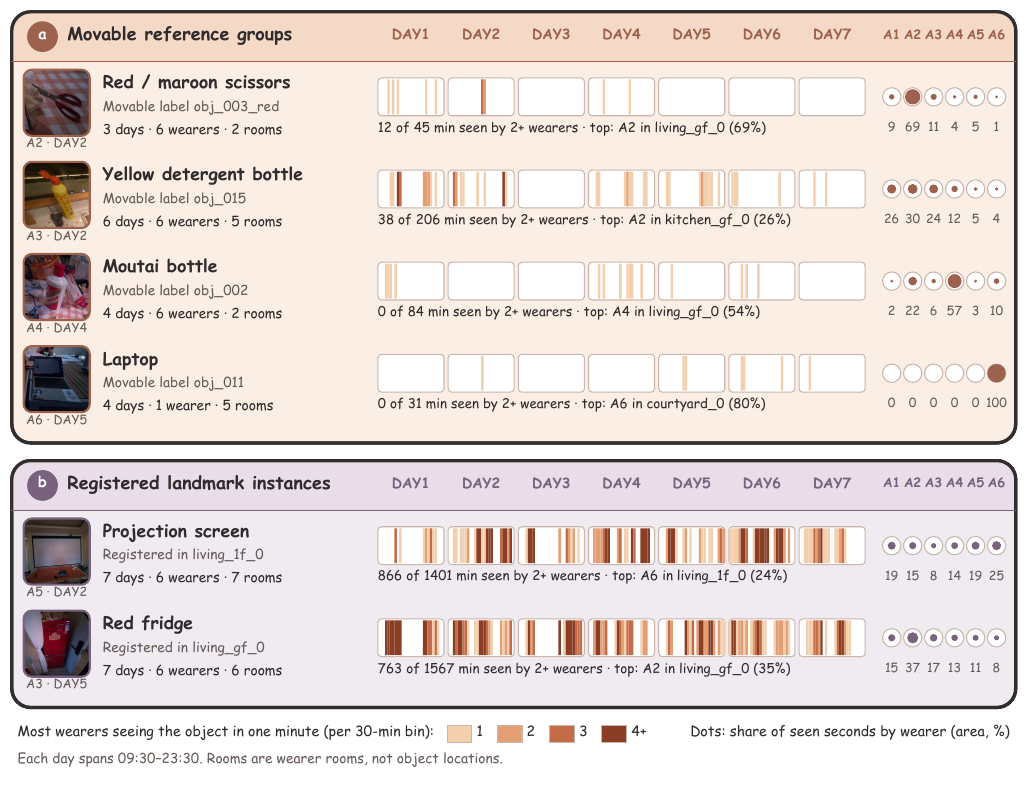}
\caption{\textbf{Recurrence of representative targets across the
shared week.} \textbf{a.} Four human-verified movable instances. \textbf{b.} Registered
landmark instances. Thumbnails are unedited source-frame crops around one
accepted mask of each target, labeled with its wearer and day. Each day
spans 09:30 to 23:30 in 30-minute bins. A bin's
shade gives the largest number of wearers who see the target label within one
wall-clock minute of that bin. Dots give each wearer's share of seen
seconds, and the note under each timeline gives the most frequent wearer and
wearer room. Room labels indicate wearer occupancy.}
\label{fig:supp_identity_group_coverage}
\end{figure*}

\cref{fig:supp_identity_group_coverage}a and b show representative movable
and landmark instances recurring across days and wearers. Their shared
time axes expose repeated sightings, while the wearer summaries show who
observes each target. \cref{tab:supp_identity_coverage} reports coverage for
all 21 movable labels and 34 registered landmarks. To compare layers
sampled at 5\,Hz and 1\,Hz, we count distinct wearer-seconds with an accepted mask and join
each second to the wearer's room in the Online Room Curator System.

All 21 movable targets recur on at least two days. Nineteen are seen by more
than one wearer, 14 by all six wearers, and 18 from more than one wearer room.
All 34 landmark instances are seen by all six wearers, and 31
on all seven days.

\definecolor{covheat}{RGB}{226,150,112}
\providecommand{\tabicon}[1]{\raisebox{-0.2em}{\includegraphics[page=#1,height=1em]{figures/supp_object_icons.pdf}}\hspace{0.4em}}
\begin{table*}[!htbp]
\centering\scriptsize
\caption{\textbf{Target recurrence across days, wearers, and rooms.} Rows marked \textsuperscript{g} are group-level labels. Landmark rows are registered instances, and registered rooms follow each name. Numbers separate distinct instances sharing a name. For landmarks, the last column names a room only when it differs from the registered room. Seen time counts distinct wearer-seconds with an accepted mask, which removes the 5\,Hz/1\,Hz sampling difference. Rooms are occupied wearer rooms from the Online Room Curator System. Same minute gives the wall-clock minutes in which at least two wearers see the target label, over all minutes with a sighting, and the largest number of wearers in one minute. Cell shading follows each wearer's share. Summary rows give totals, medians for days, wearers, and rooms, and pooled shares. Object icons are schematic.}
\label{tab:supp_identity_coverage}
\setlength{\tabcolsep}{2.5pt}\renewcommand{\arraystretch}{1.06}
\resizebox{\textwidth}{!}{%
\begin{tabular}{@{}lrrrrrrrrrrrrrl@{}}
\toprule
\multirow{2}{*}{\textbf{Target}} & \multicolumn{5}{c}{\textbf{Coverage}} & \multicolumn{2}{c}{\textbf{Same minute}} & \multicolumn{6}{c}{\textbf{Share by wearer (\%)}} & \multirow{2}{*}{\textbf{Top wearer, room (\%)}} \\
\cmidrule(lr){2-6}\cmidrule(lr){7-8}\cmidrule(lr){9-14}
& Masks & Tracks & Days & Wear. & Rooms & $\geq$2/seen & Max & A1 & A2 & A3 & A4 & A5 & A6 & \\
\midrule
\rowcolor{tableband}\multicolumn{15}{l}{\textit{Movable targets (21 labels)}} \\
\tabicon{1}Yellow detergent bottle & 32,410 & 342 & 6 & 6 & 5 & 38/206 & 4 & \cellcolor{covheat!18}26 & \cellcolor{covheat!21}30 & \cellcolor{covheat!17}24 & \cellcolor{covheat!9}12 & \cellcolor{covheat!3}5 & \cellcolor{covheat!3}4 & A2, kitchen\_gf\_0 (26) \\
\tabicon{3}Stainless-steel pot / bowl (no side handles)\textsuperscript{g} & 23,456 & 242 & 6 & 6 & 5 & 33/140 & 6 & \cellcolor{covheat!12}18 & \cellcolor{covheat!11}15 & \cellcolor{covheat!15}21 & \cellcolor{covheat!26}38 & \cellcolor{covheat!4}6 & \cellcolor{covheat!2}3 & A4, living\_gf\_0 (35) \\
\tabicon{6}Umbrella & 22,376 & 243 & 5 & 6 & 2 & 30/141 & 4 & \cellcolor{covheat!14}19 & \cellcolor{covheat!2}4 & \cellcolor{covheat!15}21 & \cellcolor{covheat!11}16 & \cellcolor{covheat!4}6 & \cellcolor{covheat!24}34 & A6, courtyard\_0 (34) \\
\tabicon{7}Seasoning / soap bottle & 16,786 & 172 & 5 & 6 & 2 & 11/121 & 2 & \cellcolor{covheat!7}11 & \cellcolor{covheat!11}15 & \cellcolor{covheat!18}26 & \cellcolor{covheat!11}15 & \cellcolor{covheat!22}32 & 1 & A5, kitchen\_gf\_0 (32) \\
\tabicon{8}Trash bin & 15,185 & 181 & 4 & 6 & 5 & 17/128 & 3 & \cellcolor{covheat!7}11 & \cellcolor{covheat!9}13 & \cellcolor{covheat!5}7 & \cellcolor{covheat!11}15 & \cellcolor{covheat!25}36 & \cellcolor{covheat!12}17 & A5, living\_1f\_0 (31) \\
\tabicon{9}Stainless-steel strainer & 15,012 & 131 & 7 & 6 & 3 & 19/77 & 3 & \cellcolor{covheat!9}13 & \cellcolor{covheat!12}18 & \cellcolor{covheat!8}11 & \cellcolor{covheat!19}27 & \cellcolor{covheat!20}29 & 2 & A5, living\_gf\_0 (29) \\
\tabicon{5}Small baking bowl / cup & 15,146 & 120 & 4 & 5 & 2 & 23/47 & 4 & \cellcolor{covheat!20}29 & \cellcolor{covheat!13}18 & \cellcolor{covheat!11}15 & \cellcolor{covheat!15}22 & \cellcolor{covheat!11}16 & 0 & A1, living\_gf\_0 (29) \\
\tabicon{10}Cooking-wine bottle & 14,049 & 141 & 4 & 6 & 2 & 4/108 & 2 & \cellcolor{covheat!20}28 & \cellcolor{covheat!11}15 & \cellcolor{covheat!28}40 & \cellcolor{covheat!2}2 & \cellcolor{covheat!8}11 & \cellcolor{covheat!2}3 & A3, kitchen\_gf\_0 (39) \\
\tabicon{11}Moutai bottle & 13,819 & 112 & 4 & 6 & 2 & 0/84 & 1 & 2 & \cellcolor{covheat!16}22 & \cellcolor{covheat!4}6 & \cellcolor{covheat!40}57 & \cellcolor{covheat!2}3 & \cellcolor{covheat!7}10 & A4, living\_gf\_0 (54) \\
\tabicon{12}Seasoning bottle & 11,082 & 100 & 4 & 4 & 2 & 8/56 & 2 & 0 & \cellcolor{covheat!13}18 & \cellcolor{covheat!31}45 & 1 & 0 & \cellcolor{covheat!25}36 & A3, kitchen\_gf\_0 (45) \\
\tabicon{13}Stainless-steel pot (two side handles) & 10,031 & 81 & 2 & 5 & 1 & 6/52 & 2 & \cellcolor{covheat!30}43 & \cellcolor{covheat!22}32 & \cellcolor{covheat!7}10 & \cellcolor{covheat!7}11 & \cellcolor{covheat!3}4 & 0 & A1, living\_gf\_0 (43) \\
\tabicon{15}Red / maroon scissors & 8,657 & 84 & 3 & 6 & 2 & 12/45 & 3 & \cellcolor{covheat!6}9 & \cellcolor{covheat!48}69 & \cellcolor{covheat!8}11 & \cellcolor{covheat!3}4 & \cellcolor{covheat!4}5 & 1 & A2, living\_gf\_0 (69) \\
\tabicon{16}Blue scissors & 8,278 & 64 & 3 & 6 & 2 & 1/40 & 2 & \cellcolor{covheat!2}3 & \cellcolor{covheat!58}84 & 1 & 1 & 2 & \cellcolor{covheat!7}10 & A2, living\_gf\_0 (84) \\
\tabicon{2}Detergent bottle & 7,415 & 81 & 3 & 5 & 2 & 5/57 & 2 & \cellcolor{covheat!13}18 & \cellcolor{covheat!3}5 & 0 & \cellcolor{covheat!4}5 & \cellcolor{covheat!18}25 & \cellcolor{covheat!33}47 & A6, kitchen\_1f\_0 (44) \\
\tabicon{17}Vase & 6,343 & 112 & 7 & 6 & 4 & 6/95 & 2 & \cellcolor{covheat!10}14 & \cellcolor{covheat!17}24 & \cellcolor{covheat!5}7 & \cellcolor{covheat!13}19 & \cellcolor{covheat!5}8 & \cellcolor{covheat!20}28 & A6, kitchen\_gf\_0 (26) \\
\tabicon{18}Apple Pencil & 5,346 & 41 & 3 & 2 & 1 & 0/30 & 1 & 0 & 2 & \cellcolor{covheat!69}98 & 0 & 0 & 0 & A3, living\_gf\_0 (98) \\
\tabicon{20}Laptop & 5,327 & 40 & 4 & 1 & 5 & 0/31 & 1 & 0 & 0 & 0 & 0 & 0 & \cellcolor{covheat!70}100 & A6, courtyard\_0 (80) \\
\tabicon{22}Flower / plant & 2,367 & 37 & 7 & 6 & 3 & 1/32 & 2 & \cellcolor{covheat!4}6 & \cellcolor{covheat!27}39 & \cellcolor{covheat!4}6 & \cellcolor{covheat!10}14 & \cellcolor{covheat!10}14 & \cellcolor{covheat!15}21 & A2, kitchen\_gf\_0 (34) \\
\tabicon{14}Pot with lid & 2,052 & 36 & 5 & 6 & 2 & 1/32 & 2 & \cellcolor{covheat!10}14 & \cellcolor{covheat!33}47 & \cellcolor{covheat!9}13 & \cellcolor{covheat!2}3 & \cellcolor{covheat!6}8 & \cellcolor{covheat!10}15 & A2, kitchen\_gf\_0 (41) \\
\tabicon{19}Whiteboard marker & 1,977 & 20 & 2 & 1 & 1 & 0/15 & 1 & 0 & 0 & 0 & 0 & 0 & \cellcolor{covheat!70}100 & A6, living\_gf\_0 (100) \\
\tabicon{4}Gray bowl\textsuperscript{g} & 1,642 & 28 & 6 & 6 & 2 & 1/25 & 4 & \cellcolor{covheat!13}18 & \cellcolor{covheat!18}26 & \cellcolor{covheat!14}20 & \cellcolor{covheat!14}20 & \cellcolor{covheat!9}13 & \cellcolor{covheat!2}3 & A3, kitchen\_gf\_0 (19) \\
\rowcolor{resultours}\textbf{All 21 targets} & \textbf{238,756} & \textbf{2,408} & \textbf{4} & \textbf{6} & \textbf{2} & \textbf{216/1,562} & \textbf{6} & \cellcolor{covheat!11}\textbf{16} & \cellcolor{covheat!15}\textbf{22} & \cellcolor{covheat!14}\textbf{19} & \cellcolor{covheat!12}\textbf{17} & \cellcolor{covheat!8}\textbf{12} & \cellcolor{covheat!10}\textbf{14} & A2, living\_gf\_0 (13) \\
\midrule
\rowcolor{tableband}\multicolumn{15}{l}{\textit{Registered landmark instances (34)}} \\
\tabicon{24}Dining table 1 {\color{black!50}(living\_gf\_0)} & 173,760 & 10,323 & 7 & 6 & 8 & 1,588/2,261 & 6 & \cellcolor{covheat!13}18 & \cellcolor{covheat!14}20 & \cellcolor{covheat!13}19 & \cellcolor{covheat!14}20 & \cellcolor{covheat!12}18 & \cellcolor{covheat!4}6 & A4 (20) \\
\tabicon{27}Projection screen 1 {\color{black!50}(living\_1f\_0)} & 74,669 & 6,194 & 7 & 6 & 7 & 866/1,401 & 6 & \cellcolor{covheat!13}19 & \cellcolor{covheat!11}15 & \cellcolor{covheat!6}8 & \cellcolor{covheat!10}14 & \cellcolor{covheat!13}19 & \cellcolor{covheat!17}25 & A6 (24) \\
\tabicon{24}Dining table 2 {\color{black!50}(living\_1f\_0)} & 60,558 & 3,881 & 7 & 6 & 7 & 600/1,207 & 6 & \cellcolor{covheat!10}14 & \cellcolor{covheat!10}15 & \cellcolor{covheat!10}15 & \cellcolor{covheat!11}16 & \cellcolor{covheat!11}16 & \cellcolor{covheat!17}24 & A6 (24) \\
\tabicon{27}Projection screen 2 {\color{black!50}(living\_gf\_0)} & 60,418 & 7,514 & 7 & 6 & 9 & 1,369/2,167 & 6 & \cellcolor{covheat!11}16 & \cellcolor{covheat!12}18 & \cellcolor{covheat!10}14 & \cellcolor{covheat!16}22 & \cellcolor{covheat!13}18 & \cellcolor{covheat!8}12 & A4 (22) \\
\tabicon{28}Blackboard / whiteboard 1 {\color{black!50}(living\_gf\_0)} & 44,399 & 4,392 & 5 & 6 & 7 & 776/1,181 & 6 & \cellcolor{covheat!13}19 & \cellcolor{covheat!8}11 & \cellcolor{covheat!11}16 & \cellcolor{covheat!14}20 & \cellcolor{covheat!13}19 & \cellcolor{covheat!11}15 & A4 (20) \\
\tabicon{29}Photographic softbox 1 {\color{black!50}(living\_1f\_0)} & 40,394 & 4,002 & 7 & 6 & 6 & 643/1,001 & 6 & \cellcolor{covheat!13}19 & \cellcolor{covheat!13}19 & \cellcolor{covheat!10}14 & \cellcolor{covheat!10}15 & \cellcolor{covheat!8}12 & \cellcolor{covheat!16}22 & A6 (22) \\
\tabicon{30}Cabinet shelf 1 {\color{black!50}(living\_gf\_0)} & 34,536 & 4,305 & 7 & 6 & 8 & 778/1,675 & 5 & \cellcolor{covheat!17}25 & \cellcolor{covheat!15}21 & \cellcolor{covheat!14}20 & \cellcolor{covheat!9}13 & \cellcolor{covheat!9}13 & \cellcolor{covheat!6}9 & A1 (24) \\
\tabicon{25}Long slat table {\color{black!50}(courtyard\_0)} & 34,381 & 1,644 & 7 & 6 & 2 & 258/512 & 6 & \cellcolor{covheat!10}15 & \cellcolor{covheat!9}13 & \cellcolor{covheat!6}8 & \cellcolor{covheat!11}15 & \cellcolor{covheat!16}22 & \cellcolor{covheat!18}26 & A6 (26) \\
\tabicon{30}Cabinet shelf 2 {\color{black!50}(living\_gf\_0)} & 30,266 & 4,484 & 7 & 6 & 9 & 857/1,786 & 6 & \cellcolor{covheat!11}15 & \cellcolor{covheat!12}17 & \cellcolor{covheat!9}12 & \cellcolor{covheat!17}25 & \cellcolor{covheat!12}18 & \cellcolor{covheat!10}14 & A4 (24) \\
\tabicon{31}Red fridge 1 {\color{black!50}(living\_gf\_0)} & 23,449 & 3,793 & 7 & 6 & 6 & 763/1,567 & 6 & \cellcolor{covheat!11}15 & \cellcolor{covheat!26}37 & \cellcolor{covheat!12}17 & \cellcolor{covheat!9}13 & \cellcolor{covheat!7}11 & \cellcolor{covheat!5}8 & A2 (35) \\
\tabicon{32}Sink counter cabinet 1 {\color{black!50}(kitchen\_gf\_0)} & 23,375 & 4,670 & 7 & 6 & 8 & 939/1,782 & 6 & \cellcolor{covheat!16}23 & \cellcolor{covheat!20}28 & \cellcolor{covheat!8}11 & \cellcolor{covheat!7}11 & \cellcolor{covheat!8}11 & \cellcolor{covheat!11}15 & A2, living\_gf\_0 (22) \\
\tabicon{23}Potted plant 1 {\color{black!50}(living\_gf\_0)} & 22,061 & 3,464 & 7 & 6 & 8 & 671/1,515 & 6 & \cellcolor{covheat!8}12 & \cellcolor{covheat!13}19 & \cellcolor{covheat!11}16 & \cellcolor{covheat!16}22 & \cellcolor{covheat!13}18 & \cellcolor{covheat!8}12 & A4 (22) \\
\tabicon{33}White microwave GF-0 {\color{black!50}(kitchen\_gf\_0)} & 20,587 & 3,474 & 7 & 6 & 7 & 658/1,553 & 6 & \cellcolor{covheat!16}22 & \cellcolor{covheat!14}20 & \cellcolor{covheat!18}25 & \cellcolor{covheat!8}11 & \cellcolor{covheat!11}16 & \cellcolor{covheat!4}6 & A1, living\_gf\_0 (18) \\
\tabicon{29}Photographic softbox 2 {\color{black!50}(living\_gf\_0)} & 19,164 & 2,513 & 7 & 6 & 6 & 485/976 & 5 & \cellcolor{covheat!14}19 & \cellcolor{covheat!9}13 & \cellcolor{covheat!10}14 & \cellcolor{covheat!15}21 & \cellcolor{covheat!14}20 & \cellcolor{covheat!8}12 & A4 (20) \\
\tabicon{34}Lamp {\color{black!50}(living\_gf\_0)} & 18,173 & 4,482 & 7 & 6 & 9 & 944/1,770 & 6 & \cellcolor{covheat!14}20 & \cellcolor{covheat!10}15 & \cellcolor{covheat!7}10 & \cellcolor{covheat!16}22 & \cellcolor{covheat!9}13 & \cellcolor{covheat!13}19 & A4 (20) \\
\tabicon{23}Potted plant 2 {\color{black!50}(living\_1f\_0)} & 14,956 & 1,910 & 7 & 6 & 6 & 341/781 & 5 & \cellcolor{covheat!12}17 & \cellcolor{covheat!12}17 & \cellcolor{covheat!3}4 & \cellcolor{covheat!16}22 & \cellcolor{covheat!13}19 & \cellcolor{covheat!14}20 & A4 (22) \\
\tabicon{28}Blackboard / whiteboard 2 {\color{black!50}(living\_gf\_0)} & 14,436 & 1,831 & 7 & 6 & 7 & 305/871 & 6 & \cellcolor{covheat!17}24 & \cellcolor{covheat!4}6 & \cellcolor{covheat!4}6 & \cellcolor{covheat!10}14 & \cellcolor{covheat!16}23 & \cellcolor{covheat!19}28 & A6, courtyard\_0 (25) \\
\tabicon{35}Black BBQ grill {\color{black!50}(courtyard\_0)} & 14,169 & 2,185 & 7 & 6 & 6 & 396/961 & 5 & \cellcolor{covheat!19}27 & \cellcolor{covheat!20}29 & \cellcolor{covheat!8}11 & \cellcolor{covheat!4}6 & \cellcolor{covheat!8}12 & \cellcolor{covheat!11}16 & A2 (25) \\
\tabicon{33}White microwave GF-1 {\color{black!50}(kitchen\_gf\_0)} & 12,928 & 2,287 & 7 & 6 & 7 & 396/1,139 & 6 & \cellcolor{covheat!16}23 & \cellcolor{covheat!15}21 & \cellcolor{covheat!16}23 & \cellcolor{covheat!8}12 & \cellcolor{covheat!9}13 & \cellcolor{covheat!5}8 & A1, living\_gf\_0 (21) \\
\tabicon{36}Laptop rack {\color{black!50}(bedroom\_gf\_0)} & 8,501 & 1,189 & 7 & 6 & 6 & 142/726 & 6 & \cellcolor{covheat!47}68 & \cellcolor{covheat!2}3 & \cellcolor{covheat!3}5 & \cellcolor{covheat!4}6 & \cellcolor{covheat!3}4 & \cellcolor{covheat!10}14 & A1 (56) \\
\tabicon{33}White microwave 1F-0 {\color{black!50}(kitchen\_1f\_0)} & 8,130 & 1,710 & 7 & 6 & 5 & 298/840 & 6 & \cellcolor{covheat!20}28 & \cellcolor{covheat!6}9 & \cellcolor{covheat!11}15 & \cellcolor{covheat!13}18 & \cellcolor{covheat!5}8 & \cellcolor{covheat!15}22 & A1, living\_1f\_0 (27) \\
\tabicon{26}Yellow bistro table {\color{black!50}(courtyard\_0)} & 7,575 & 1,319 & 7 & 6 & 8 & 213/686 & 5 & \cellcolor{covheat!13}18 & \cellcolor{covheat!12}17 & \cellcolor{covheat!10}14 & \cellcolor{covheat!5}7 & \cellcolor{covheat!11}16 & \cellcolor{covheat!19}27 & A5, kitchen\_gf\_0 (14) \\
\tabicon{20}Rack laptop 5 {\color{black!50}(bedroom\_gf\_0)} & 5,726 & 1,267 & 7 & 6 & 5 & 153/814 & 4 & \cellcolor{covheat!45}65 & \cellcolor{covheat!3}5 & \cellcolor{covheat!3}4 & \cellcolor{covheat!4}5 & \cellcolor{covheat!6}9 & \cellcolor{covheat!9}12 & A1 (45) \\
\tabicon{37}Wooden console bench {\color{black!50}(living\_gf\_0)} & 5,310 & 1,213 & 7 & 6 & 5 & 209/640 & 5 & \cellcolor{covheat!25}35 & \cellcolor{covheat!4}6 & \cellcolor{covheat!4}6 & \cellcolor{covheat!8}12 & \cellcolor{covheat!11}16 & \cellcolor{covheat!17}25 & A1 (35) \\
\tabicon{20}Fixed shelf laptop {\color{black!50}(living\_gf\_0)} & 3,956 & 879 & 7 & 6 & 6 & 84/597 & 4 & \cellcolor{covheat!49}70 & \cellcolor{covheat!4}6 & \cellcolor{covheat!3}4 & \cellcolor{covheat!4}6 & \cellcolor{covheat!4}6 & \cellcolor{covheat!6}8 & A1, bedroom\_gf\_0 (54) \\
\tabicon{38}Track light {\color{black!50}(kitchen\_gf\_0)} & 3,471 & 833 & 7 & 6 & 2 & 128/536 & 5 & \cellcolor{covheat!13}18 & \cellcolor{covheat!23}33 & \cellcolor{covheat!12}17 & \cellcolor{covheat!11}16 & \cellcolor{covheat!6}9 & \cellcolor{covheat!5}8 & A2, living\_gf\_0 (31) \\
\tabicon{20}Rack laptop 3 {\color{black!50}(bedroom\_gf\_0)} & 3,003 & 735 & 7 & 6 & 4 & 72/532 & 4 & \cellcolor{covheat!40}57 & \cellcolor{covheat!6}8 & \cellcolor{covheat!5}7 & \cellcolor{covheat!6}8 & \cellcolor{covheat!5}7 & \cellcolor{covheat!9}13 & A1 (37) \\
\tabicon{31}Red fridge 2 {\color{black!50}(living\_1f\_0)} & 2,791 & 679 & 7 & 6 & 3 & 101/447 & 4 & \cellcolor{covheat!16}22 & \cellcolor{covheat!6}9 & \cellcolor{covheat!5}8 & \cellcolor{covheat!9}14 & \cellcolor{covheat!13}19 & \cellcolor{covheat!20}28 & A6 (25) \\
\tabicon{23}Potted plant 3 {\color{black!50}(entryway\_gf\_0)} & 2,665 & 843 & 7 & 6 & 5 & 123/607 & 4 & \cellcolor{covheat!22}31 & \cellcolor{covheat!8}11 & \cellcolor{covheat!8}11 & \cellcolor{covheat!12}18 & \cellcolor{covheat!10}14 & \cellcolor{covheat!10}15 & A1, corridor\_gf\_0 (16) \\
\tabicon{21}Black console laptop {\color{black!50}(living\_gf\_0)} & 1,955 & 305 & 4 & 6 & 2 & 59/141 & 4 & \cellcolor{covheat!11}16 & \cellcolor{covheat!15}22 & \cellcolor{covheat!11}16 & \cellcolor{covheat!29}41 & 2 & \cellcolor{covheat!2}3 & A4 (41) \\
\tabicon{32}Sink counter cabinet 2 {\color{black!50}(kitchen\_1f\_0)} & 1,884 & 806 & 7 & 6 & 4 & 115/546 & 5 & \cellcolor{covheat!17}24 & \cellcolor{covheat!14}20 & \cellcolor{covheat!2}3 & \cellcolor{covheat!7}10 & \cellcolor{covheat!7}10 & \cellcolor{covheat!22}32 & A6, living\_1f\_0 (23) \\
\tabicon{20}Rack laptop 6 {\color{black!50}(bedroom\_gf\_0)} & 1,578 & 471 & 6 & 6 & 5 & 47/352 & 3 & \cellcolor{covheat!38}55 & \cellcolor{covheat!4}6 & \cellcolor{covheat!7}9 & \cellcolor{covheat!3}5 & \cellcolor{covheat!5}7 & \cellcolor{covheat!13}19 & A1 (28) \\
\tabicon{39}Washing machine {\color{black!50}(entryway\_gf\_0)} & 1,067 & 371 & 7 & 6 & 4 & 42/288 & 5 & \cellcolor{covheat!19}27 & \cellcolor{covheat!10}14 & \cellcolor{covheat!8}11 & \cellcolor{covheat!12}16 & \cellcolor{covheat!11}15 & \cellcolor{covheat!11}16 & A1 (16) \\
\tabicon{20}Rack laptop 2 {\color{black!50}(bedroom\_gf\_0)} & 1,028 & 293 & 7 & 6 & 4 & 6/245 & 2 & \cellcolor{covheat!53}76 & 1 & \cellcolor{covheat!6}8 & \cellcolor{covheat!6}9 & \cellcolor{covheat!2}3 & \cellcolor{covheat!2}3 & A1 (47) \\
\rowcolor{resultours}\textbf{All 34 instances} & \textbf{795,319} & \textbf{90,261} & \textbf{7} & \textbf{6} & \textbf{6} & \textbf{15,425/34,103} & \textbf{6} & \cellcolor{covheat!14}\textbf{20} & \cellcolor{covheat!12}\textbf{18} & \cellcolor{covheat!10}\textbf{14} & \cellcolor{covheat!12}\textbf{17} & \cellcolor{covheat!11}\textbf{16} & \cellcolor{covheat!10}\textbf{15} & A4, living\_gf\_0 (11) \\
\bottomrule
\end{tabular}}
\end{table*}

\Needspace{15\baselineskip}
\section{Question construction and verification}
\label{sec:supp_questions}
\cref{tab:supp_question_templates} shows the question forms, and
\cref{fig:supp_qa_workbench} shows how experts check their answers against
video evidence. This section details answer construction, visit-boundary
conventions, temporal coverage, and the review operations.

\Needspace{10\baselineskip}
\subsection{Question inventory and answer construction}
\label{sec:supp_question_taxonomy}
\begin{wraptable}{r}{0.48\textwidth}
\centering\scriptsize
\caption{\textbf{Question inventory.} Unique questions by partition.
All short-horizon questions are evaluation-only.}
\label{tab:supp_qa_inventory}
\begin{tabularx}{\linewidth}{@{}Yrrrr@{}}
\toprule Component & Train & Val. & Test & Total \\
\midrule
Week-long & 5,750 & 383 & 1,533 & 7,666 \\
Short-horizon & 0 & 0 & 787 & 787 \\
\midrule
All sources & 5,750 & 383 & 2,320 & 8,453 \\
\bottomrule
\end{tabularx}
\end{wraptable}

The collection contains 8,453 questions across seven abilities, Q01 to Q07.
Q07 comprises five visit-reasoning subtypes.
\cref{tab:supp_qa_inventory} lists the training and evaluation partitions.
Each question stores its source component, wearer scope, target, time
interval, public references, answer, and required supporting evidence.
Split assignment groups questions by their core-task identifier.
We derive answers from reviewed identities, target observations, room
intervals, and the supplied environment graph using four procedures.

\paragraph{Identity comparisons: Q01 and Q06.}
We pair object or room references using reviewed instance identities and
category labels. The paired identities determine whether the answer is
the same physical instance, different instances of the same category, or
different categories.
Room comparisons use room type as the category and room ID as the identity.

\paragraph{Historical observations: Q02, Q03, and Q04.}
We gather target sightings within each question's wearer scope and time
interval. Q02 determines which of the two queried wearers recorded the
target. Q03 compares sighting times across all permitted wearers before
the query time to select the latest witness. Q04 labels every candidate window
and selects the complete subset containing the target. An observation
supports a positive label only when its reviewed identity matches the
reference target. Missing recordings retain unknown visibility.

\par\WFclear
\paragraph{Spatial paths: Q05.}
Reference views identify the start and destination rooms. We compute
shortest paths between their room IDs on the supplied environment graph
to establish the answer. Thus the reference views determine the endpoints,
and the graph determines the path between them.

\paragraph{Room visits: Q07A to Q07E.}
We derive ordered stays from the reviewed room intervals. Q07A reads the
room sequence and retains returns. Q07B subtracts the current stay's entry
time from the query time. Q07C selects the requested visit ordinal and
subtracts entry from exit, requiring both boundaries within the observation
interval. Q07D retrieves the room immediately before the current stay.
Q07E counts separate visits overlapping the interval. The visit-boundary
conventions below handle stays crossing clip or query boundaries.

\paragraph{From annotations to question text.}
Each procedure fixes the answer, options, references, and evidence intervals
before language-model rewriting. Semantic checks compare the rewritten
question with these fixed fields. Experts then verify the wording, answer,
and evidence in the QA Audit Workbench, as detailed in
\cref{sec:qa_refinement}. Public question inputs and private answer records
are exported separately.

\paragraph{Evaluation composition.}
The evaluated week-long subset contains 120 questions for each of 11
detailed types. Q07 comprises five types and contributes 600 questions to
micro accuracy. The seven-ability macro average assigns Q07 the same weight
as each other ability. This evaluation subset is drawn from the test
partition in \cref{tab:supp_qa_inventory}.

\subsection{Question templates and answer forms}
\label{sec:supp_question_templates}
\cref{tab:supp_question_templates} gives the question and answer form for
each subtype. Colored brackets identify the reference, wearer, room, and
time parameters filled for each question. The response selects one answer
option and supplies the evidence required by that question. For Q04, one
option can encode several target-containing windows.

\paragraph{Visit boundaries and ordinals.}
A visit is uninterrupted occupancy in one physical room and continues
across adjacent clip boundaries. For Q07C and Q07E, the visit list includes
stays with positive-duration overlap with the observation interval.
A stay already ongoing at the interval start counts once. Entry exactly
at the interval end does not count. Q07C numbers this list before selecting
the requested ordinal and requires the selected visit's entry and exit
to lie within the interval. Duration options use the bounds specified
by each question.

\paragraph{Identifier mapping.}
Released identifiers use the internal codes Q1, Q2, Q3A, Q3B, and Q4 for
Q01, Q02, Q03, Q04, and Q06. Q05 and Q07A through Q07E retain their
identifiers.

\begingroup
\newcommand{\EWslot}[1]{\textcolor{cvprblue}{\textit{[#1]}}}
\begin{table*}[!htbp]
\centering\small
\caption{\textbf{Question templates for the seven spatial-memory abilities.}
Colored slots vary with the query's public references, wearers, and time scope.
Q07 expands into five visit-reasoning subtypes.}
\label{tab:supp_subtypes}
\label{tab:supp_question_templates}
\label{tab:supp_visit_templates}
\includegraphics[width=\textwidth]{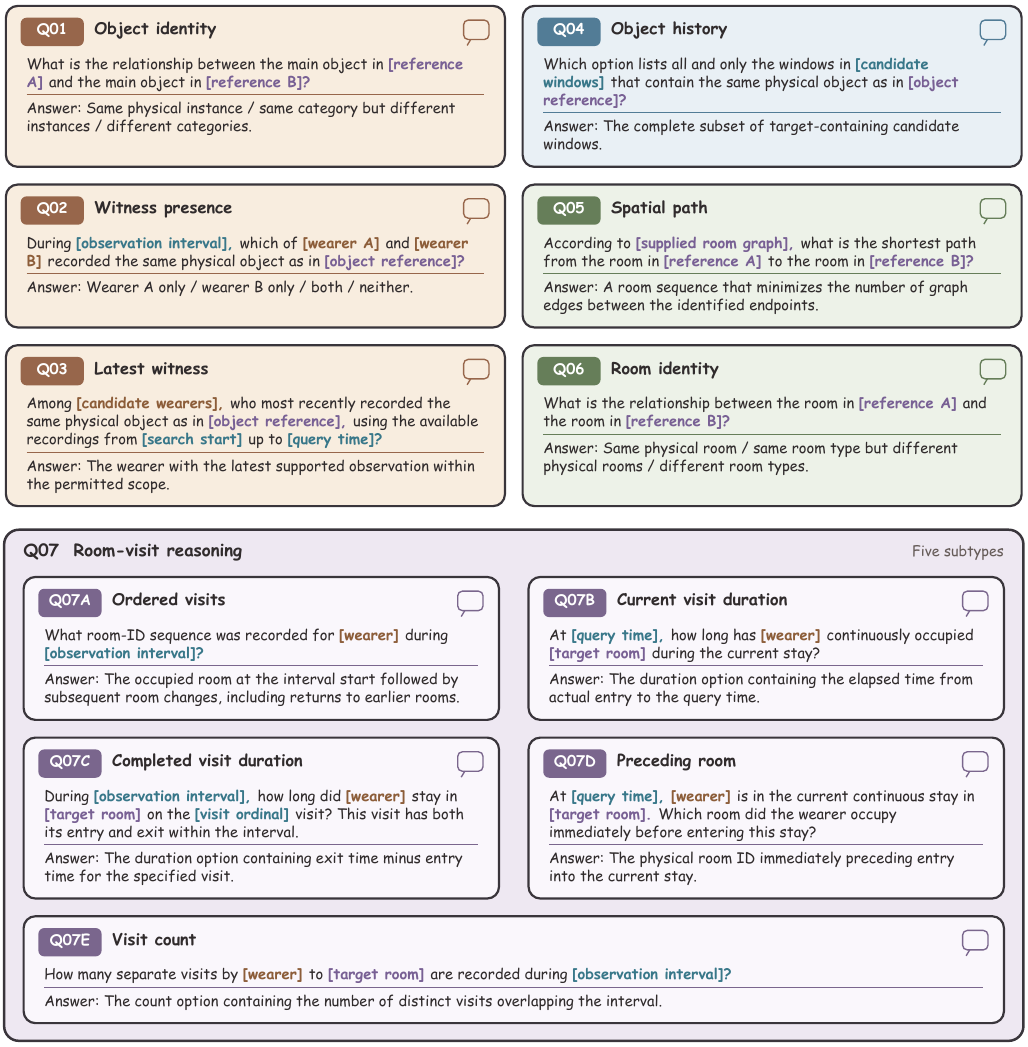}
\par\vspace{3pt}\raggedright\footnotesize
\textbf{Scope.} Candidate windows specify wearer and time.
Q02 checks both wearers, Q03 compares all permitted wearers before the
query time, and Q04 checks every candidate window.
Missing recordings retain unknown visibility.
Q05 computes paths on the supplied environment graph.
\end{table*}
\endgroup

\Needspace{24\baselineskip}
\subsection{Temporal and cross-wearer coverage}
\label{sec:supp_temporal_scope}
\begin{wraptable}{r}{0.50\textwidth}
\captionsetup{font=footnotesize}
\centering\footnotesize
\caption{\textbf{Historical support in the 120 evaluated Q04 questions.}
Positive windows are those selected by the frozen correct answer.
Cross-day and cross-wearer rows are separate counts.}
\label{tab:supp_q06_scope}
\setlength{\tabcolsep}{4pt}
\renewcommand{\arraystretch}{1.05}
\begin{tabularx}{\linewidth}{@{}Yr@{}}
\toprule Property & Questions \\
\midrule
\rowcolor{tableband}\multicolumn{2}{l}{\textit{Candidate scope}} \\
Candidates on multiple dates & 118 / 120 \\
Candidates from multiple wearers & 117 / 120 \\
\midrule
\rowcolor{tableband}\multicolumn{2}{l}{\textit{Target-positive support}} \\
\rowcolor{resultours}At least one positive window older than 24 h & \textbf{103 / 120} \\
Positive windows on multiple dates & 78 / 120 \\
Positive windows from multiple wearers & 78 / 120 \\
Positive-window span longer than 24 h & 70 / 120 \\
\bottomrule
\end{tabularx}
\end{wraptable}
\cref{tab:supp_q06_scope} measures how far the 120 evaluated Q04 questions
reach across dates and wearers. Each question provides an object reference
and four candidate windows. We select the target-containing windows using
the correct option and read their wearers, dates, and intervals from the
public window metadata.
This yields 300 positive windows among 480 candidates.
Window lengths range from 6 to 600 seconds, with a median of 120 seconds.

For query time $t_c$ and window end $e$, we measure window age as $t_c-e$.
Every sighting inside the window is therefore at least this old.
As shown in \cref{tab:supp_q06_scope}, 103 questions contain at least one
positive window lying entirely more than 24 hours before the query time.
Across these questions, 198 positive windows satisfy this condition.
The oldest positive-window age per question has a median of 78.9 hours and
a maximum of 150.7 hours. Positive windows span multiple dates in 78
questions and multiple wearers in 78 questions, counted separately.
The positive-window span runs from the earliest positive start to the
latest positive end.

\par\WFclear
\paragraph{A cross-day example.}
\cref{tab:supp_q06_example} lists all four candidates for query
\texttt{EWO10M\_Q3B\_e3b45c4031c16d89}, issued at DAY7 18:00:06.
The question asks which windows contain the referenced physical object.
The correct option B
selects W1, W2, and W3, linking A4 on DAY3, A3 on DAY5, and A6 on DAY1.
W4 on DAY7 is a distractor. All three positive windows precede the query by
more than 24 hours. Their temporal span exceeds four days.

\begin{table*}[!htbp]
\centering\small
\caption{\textbf{One Q04 query links three historical wearers and dates.}
The query time is DAY7 18:00:06.000. Each candidate spans ten minutes.
Positive labels come from correct option B, which selects W1 + W2 + W3.
Age is measured from each window's end to the query time.}
\label{tab:supp_q06_example}
\setlength{\tabcolsep}{5pt}
\renewcommand{\arraystretch}{1.08}
\begin{tabularx}{\textwidth}{@{}lCC>{\centering\arraybackslash}p{0.36\textwidth}rC@{}}
\toprule Window & Wearer & Day & Time interval & Age, h & Same target \\
\midrule
\rowcolor{resultours}W1 & A4 & DAY3 & 20:04:31.297 to 20:14:31.297 & 93.76 & \featyes \\
\rowcolor{resultours}W2 & A3 & DAY5 & 19:11:33.909 to 19:21:33.909 & 46.64 & \featyes \\
\rowcolor{resultours}W3 & A6 & DAY1 & 13:55:56.799 to 14:05:56.799 & 147.90 & \featyes \\
W4 & A6 & DAY7 & 14:23:37.783 to 14:33:37.783 & 3.44 & \featno \\
\bottomrule
\end{tabularx}
\end{table*}

\subsection{Question and answer vocabulary}
\label{sec:supp_qa_vocabulary}
\begin{figure*}[!htbp]
\centering
\includegraphics[width=\textwidth]{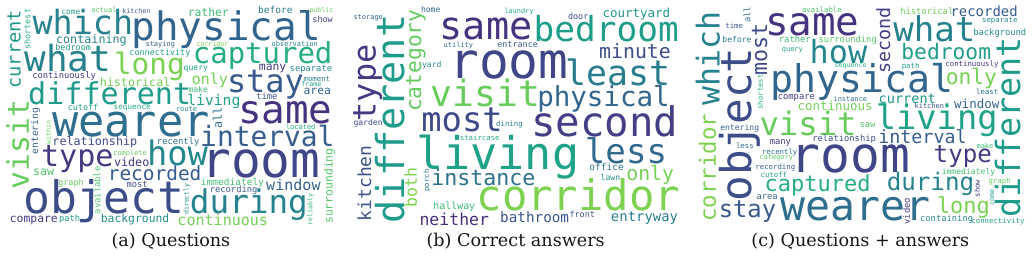}
\caption{\textbf{Question and answer vocabulary in EgoWitness.}
The panels show questions, correct answers, and their combined vocabulary
for 7,666 week-long and 787 short-horizon questions.
Font sizes reflect relative token frequencies within each panel.
The corpus uses question text and correct-answer text only.}
\label{fig:supp_qa_vocabulary}
\end{figure*}

\cref{fig:supp_qa_vocabulary} summarizes the English vocabulary of all
8,453 questions and their correct answers across both benchmark components.
Room, object, wearer, same, and physical dominate the questions, and room
names with same and different dominate the answers, matching the focus on
instance identity, observers, and places.
We select one question formulation per identifier and remove repeated
paraphrases, capture metadata, and response-format instructions.
Correct-answer text comes from the gold option, using fixed English
translations for Chinese options.

We lowercase words, normalize common inflections, and remove stopwords,
timestamps, and numeric identifiers. Room identifiers contribute their
category words without floor or instance suffixes.
This preprocessing yields lexical tokens from 6,884 answers. The remaining
1,569 answers contain only wearer or candidate-window codes and contribute
no tokens to the answer panel.
The combined panel sums the question and answer counts without source
reweighting.

\subsection{QA refinement and expert verification}
\label{sec:qa_refinement}

\begin{figure*}[!htbp]
\centering
\includegraphics[width=\textwidth]{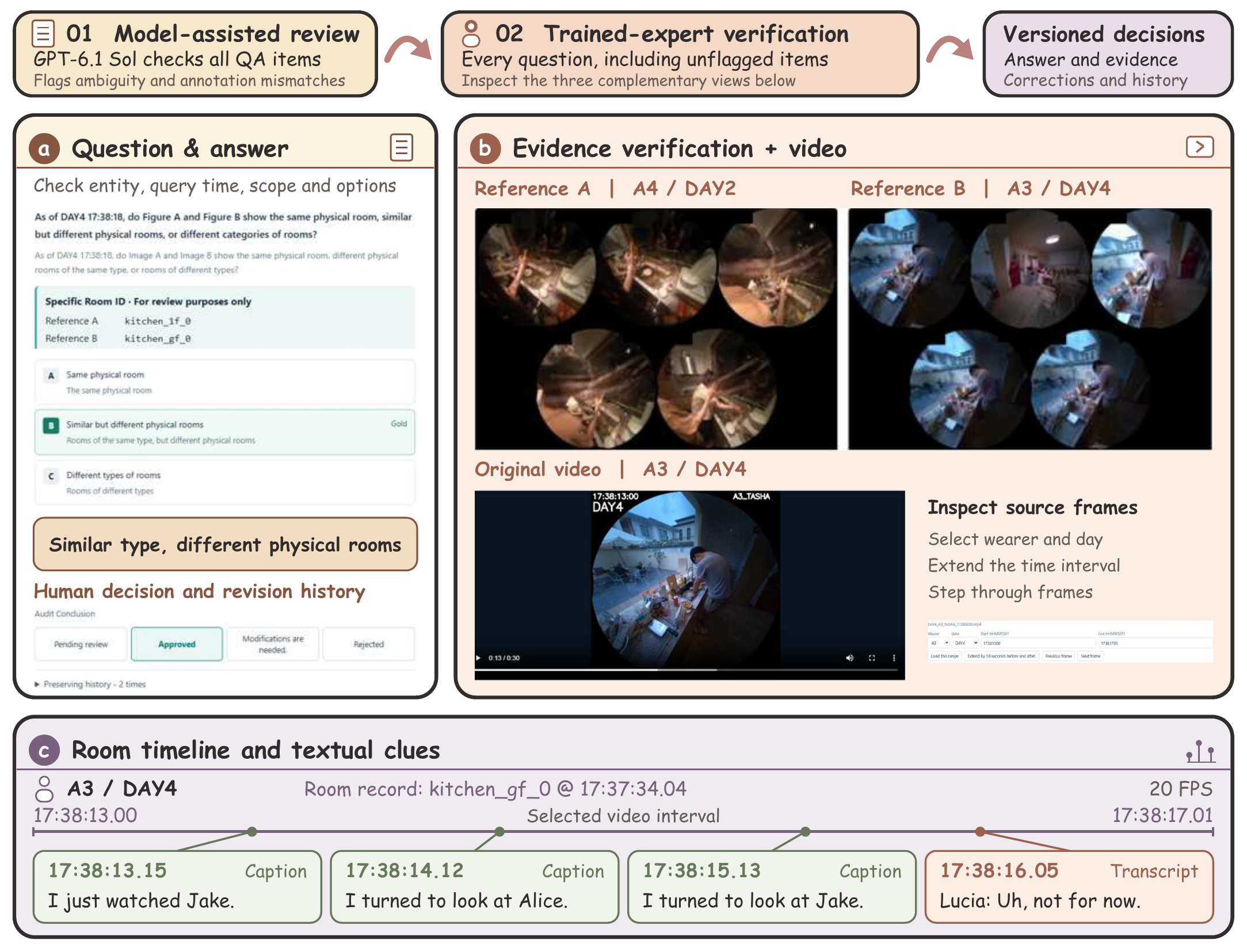}
\caption{\textbf{Two-pass QA review.}
The top row links model-assisted review, expert verification, and versioned
decisions. \textbf{a.} Cropped question and answer view.
\textbf{b.} Cropped multi-frame references and source video for identity verification.
\textbf{c.} Redrawn room context and textual cues aligned with the selected
interval. Timecodes use HH:MM:SS.FF at 20 FPS. The example compares kitchens
observed by A4 on DAY2 and A3 on DAY4.}
\label{fig:supp_qa_workbench}
\end{figure*}

\cref{fig:supp_qa_workbench} shows two review passes: a model-assisted
examination followed by expert verification in the QA Audit Workbench.
All questions across the seven task types and both benchmark components
have completed both passes. Experts establish the final wording, answer
labels, and supporting evidence.

The first step below describes the model pass. Steps 2 to 5 describe the
expert pass through the workbench's question, evidence, and context views.

\paragraph{Step 1: Model-assisted preliminary review.}
GPT-6.1 Sol compares each question, its answer choices, the proposed
answer, and the associated annotations. It flags unclear references,
ambiguous wording, and annotation mismatches for expert inspection.
These flags accompany each item into the expert pass.

\paragraph{Step 2: Question and answer refinement.}
In \cref{fig:supp_qa_workbench}a, experts check the target, query time,
permitted observation scope, and meaning of each answer choice.
The bilingual display lets them compare Chinese and English formulations.
They correct wording or labels when the question and intended answer
express different spatial or temporal relations.

\paragraph{Step 3: Visual identity verification.}
\cref{fig:supp_qa_workbench}b pairs reference images with source-video
playback. Experts compare physical identity across views, using enlarged
references and frame-level navigation to inspect occlusion and similar
instances. For object questions, they distinguish individual objects within
a category. For room questions, they distinguish physical rooms within a
room type. The illustrated example compares similar kitchens observed by
A4 on DAY2 and A3 on DAY4.

\paragraph{Step 4: Temporal and witness verification.}
The context view in \cref{fig:supp_qa_workbench}c aligns room intervals
with captions and transcripts to locate relevant video segments. Experts
then inspect those segments in panel b, switching wearers and extending
playback around event boundaries. For visit questions, they verify entry,
exit, and visit order. For witness questions, they check the permitted
wearers, dates, query time, and evidence intervals. Latest-witness verification
includes checking for later sightings before the query time. Unobserved
periods retain unknown visibility.

\paragraph{Step 5: Final labeling and revision tracking.}
Experts record approval, a revision request, or rejection in the question
view of \cref{fig:supp_qa_workbench}a. They save corrections to wording,
answer choices, the correct answer, explanations, and evidence windows.
The workbench stores these decisions and edits in a versioned history
linked to the original records.

\Needspace{20\baselineskip}
\section{\agentname\ training}
\label{sec:supp_training}

\subsection{Training setup}
\label{sec:supp_optimization}
\begin{wraptable}{r}{0.52\textwidth}
\centering\scriptsize
\caption{\textbf{Training configuration.} The visual and retrieval services remain fixed across the Base, SFT, and GRPO controller comparison.}
\label{tab:supp_settings}
\begin{tabularx}{\linewidth}{@{}Yr@{}}
\toprule Item & Value \\
\midrule
Controller & Qwen3.5-4B \\
RL algorithm / rollout group & GRPO / 4 \\
Prompt batch & 18 \\
RL learning rate & $10^{-6}$ \\
KL coefficient / clip & 0.001 / 0.2 \\
Training temperature / top-$p$ & 1.0 / 1.0 \\
\midrule
SFT dialogues / assistant-step samples & 1,100 / 3,835 \\
SFT learning rate / global batch & $10^{-5}$ / 12 \\
SFT checkpoint & step 32 / epoch 0.1 \\
RL checkpoint / total updates & step 259 / 259 \\
\bottomrule
\end{tabularx}
\end{wraptable}

\cref{tab:supp_settings} summarizes the SFT and GRPO configurations.
SFT uses 1,100 teacher dialogues expanded into 3,835 assistant-step samples.
A GPT-5.1 Chat controller generated the dialogues with the same tools and
backends, 20 decisions, and 19 tool calls. The 1,100 dialogues take 100 per
detailed type without using model outputs. They are selected by format
validity alone, without filtering by answer correctness.
We update the language parameters while freezing the vision tower and aligner.
Training uses ms-swift, BF16, and DeepSpeed ZeRO-3 on six GPUs, without
packing, with a cosine learning-rate schedule and 5\% warmup.
The step-32 checkpoint at 0.1 epoch of a 64-step SFT run initializes GRPO, which uses the
remaining 4,650 training questions and four rollouts per question.
The resulting 18,600 trajectories produce 259 updates.

\paragraph{Training schedules.}
The main GRPO run comprises 259 updates from the 0.1-epoch SFT checkpoint,
which was fixed before RL training began.

SFT/2 is the step-16 checkpoint of the same SFT run, at 0.05 epoch, and
SFT/4 is a separate 8-step run of 0.025 epoch. Their RL continuations both use the 4,650 GRPO questions.
The Direct-RL checkpoint is step 414.

\paragraph{Training curves.}
Main-paper Figure~5 plots the SFT+GRPO and Direct-RL updates. Each curve
overlays raw scalars with an exponential moving average of coefficient 0.2. The KL curve measures divergence from the fixed reference policy.

\par\WFclear
\subsection{GRPO objective}
\label{sec:supp_grpo}
The outcome reward in main-paper Eq.~2 combines answer correctness,
task-specific evidence credit, and episode-format validity. The private
evaluator scores answers and supporting observations. Submission validation
checks structure, citations, and the permitted evidence scope.

\paragraph{Group-relative advantage.}
We use Group Relative Policy Optimization~\cite{grpo} for the multi-turn
controller. Tool observations condition subsequent decisions, as in
tool-augmented egocentric reasoning~\cite{egor1}. For each question $q$, the
rollout policy $\pi_{\mathrm{old}}$ generates $G=4$ trajectories. Let $R_{qi}$
be the outcome reward of trajectory $i$ defined above. Its group mean
$\mu_q$, sample standard deviation $\sigma_q$, and advantage $\hat u_{qi}$
are
\begin{equation}
\begin{aligned}
\mu_q &= \frac{1}{G}\sum_{i=1}^{G}R_{qi}, \\
\sigma_q &= \sqrt{\frac{1}{G-1}\sum_{i=1}^{G}(R_{qi}-\mu_q)^2}, \\
\hat u_{qi} &= \frac{R_{qi}-\mu_q}{\sigma_q+10^{-6}}.
\end{aligned}
\label{eq:supp_grpo_advantage}
\end{equation}
The same advantage weights every generated token in a trajectory. A group
with identical rewards has zero advantage and remains in the batch.

\paragraph{Masked policy objective.}
Let $\mathcal B$ be an optimization batch of trajectories. We omit the
question index below. At position $t$ of trajectory $i$, $x_{it}$ is the
token and $h_{it}$ contains its prompt and preceding tokens, including tool
observations. The binary mask $m_{it}$ is one for sampled assistant tokens
and zero for tool observations, injected context, and padding. The token
count is $N=\sum_{i\in\mathcal B}\sum_t m_{it}$. For policy $p$, write
$\ell_{it}^{p}=\log\pi_p(x_{it}\mid h_{it})$. Here $\theta$ denotes the
trainable policy, $\mathrm{old}$ its rollout version, and $\mathrm{ref}$ the
fixed SFT checkpoint at epoch 0.1. With clipping width $\epsilon=0.2$, the
probability ratio $\rho_{it}$ and clipped surrogate $s_{it}$ are
\begin{equation}
\begin{aligned}
\rho_{it} &= \exp\!\bigl(\operatorname{clip}
    (\ell_{it}^{\theta}-\ell_{it}^{\mathrm{old}},-20,20)\bigr), \\
s_{it} &= \min\!\left\{\rho_{it}\hat u_i,
    \operatorname{clip}(\rho_{it},1-\epsilon,1+\epsilon)\hat u_i\right\}.
\end{aligned}
\label{eq:supp_grpo_clip}
\end{equation}
The implementation applies dual clipping with $c=3$ to negative advantages,
giving $\tilde s_{it}=\max\{s_{it},c\hat u_i\}$ when $\hat u_i<0$ and
$\tilde s_{it}=s_{it}$ otherwise. For the sampled KL penalty, define
$d_{it}=\operatorname{clip}(\ell_{it}^{\mathrm{ref}}-
\ell_{it}^{\theta},-20,20)$ and
$k_{it}=\operatorname{clip}(\exp(d_{it})-d_{it}-1,-10,10)$.
With coefficient $\beta=0.001$, we minimize
\begin{equation}
\mathcal L_{\mathrm{GRPO}}(\theta)
=\frac{1}{N}\sum_{i\in\mathcal B}\sum_t
m_{it}\bigl[-\tilde s_{it}+\beta k_{it}\bigr].
\label{eq:supp_grpo_objective}
\end{equation}
This is a token-level mean over all generated assistant tokens in the
batch. The KL penalty enters the loss, and the outcome reward excludes it. No entropy bonus is used.

\section{Evaluation setup}
\label{sec:supp_models}

\subsection{Models and visual inputs}
\cref{tab:supp_model_registry} identifies the twelve direct VLMs and the
API controllers.
The shared-tool comparison uses the public Qwen3.5-4B controller, its SFT
checkpoint, its SFT+GRPO checkpoint, and four zero-shot API controllers.
All seven conditions share the visual and retrieval services, public
references, and execution limits.

\begin{table*}[!htbp]
\centering\small
\caption{\textbf{Evaluated direct models and API controllers.} Each entry gives the model and its checkpoint or API identifier. The shared-tool Base, SFT, and GRPO conditions use Qwen3.5-4B.}
\label{tab:supp_model_registry}
\begingroup
\newlength{\EWtxtA}\newlength{\EWtxtB}\newlength{\EWtxtC}\newlength{\EWblank}
\settowidth{\EWtxtA}{\footnotesize\ttfamily InternVideo3-8B-Instruct}
\settowidth{\EWtxtB}{\footnotesize\ttfamily VideoChat-Flash-Qwen2-7B\_res448}
\settowidth{\EWtxtC}{\footnotesize\ttfamily Efficient-Large-Model/}
\setlength{\EWblank}{\dimexpr(\textwidth-4\tabcolsep-\EWtxtA-\EWtxtB-\EWtxtC)/3\relax}
\newsavebox{\EWentrybox}
\newcommand{\EWlogocell}[2]{\leavevmode
  \sbox{\EWentrybox}{\renewcommand{\arraystretch}{0.95}%
    \begin{tabular}[c]{@{}l@{}}#2\end{tabular}}%
  \hbox to\linewidth{\hbox to\dimexpr\linewidth-\EWblank\relax{\usebox{\EWentrybox}\hss}%
    \hbox to\EWblank{\hss$\vcenter{\hbox{\includegraphics[width=2.4em,height=1.9em,
      keepaspectratio]{figures/logos/#1}}}$\hss}}}
\newcommand{\EWmodelentry}[4]{\EWlogocell{#1}{\textbf{#2}\\
  {\footnotesize\ttfamily #3}\\{\footnotesize\ttfamily #4}}}
\newcommand{\EWapientry}[3]{\EWlogocell{#1}{\textbf{#2}\\{\footnotesize\ttfamily #3}}}
\setlength{\tabcolsep}{6pt}
\renewcommand{\arraystretch}{1.04}
\begin{tabular}{@{}p{\dimexpr\EWtxtA+\EWblank}p{\dimexpr\EWtxtB+\EWblank}p{\dimexpr\EWtxtC+\EWblank}@{}}
\toprule
\rowcolor{tableband}
\multicolumn{3}{@{}l}{\strut\textbf{Open-weight models}} \\
\addlinespace[3pt]
\EWmodelentry{qwen-color}{Qwen3.5-4B~\cite{qwen35}}{Qwen/}{Qwen3.5-4B} &
\EWmodelentry{google-color}{Gemma 4 12B~\cite{gemma4}}{google/}{gemma-4-12B-it} &
\EWmodelentry{qwen-color}{Qwen3.8-27B~\cite{qwen38}}{Qwen/}{Qwen3.8-27B} \\
\cmidrule(lr){1-1}\cmidrule(lr){2-2}\cmidrule(lr){3-3}
\EWmodelentry{internlm-color}{InternVL3.5-8B~\cite{internvl35}}{OpenGVLab/}{InternVL3\_5-8B} &
\EWmodelentry{xiaomi}{MiMo-VL-7B-RL~\cite{mimovl}}{XiaomiMiMo/}{MiMo-VL-7B-RL} &
\EWmodelentry{alibaba-color}{VideoLLaMA3-7B~\cite{videollama3}}{DAMO-NLP-SG/}{VideoLLaMA3-7B} \\
\cmidrule(lr){1-1}\cmidrule(lr){2-2}\cmidrule(lr){3-3}
\EWmodelentry{internlm-color}{InternVideo3-8B~\cite{internvideo3}}{yanziang/}{InternVideo3-8B-Instruct} &
\EWmodelentry{internlm-color}{VideoChat-Flash-7B~\cite{videochatflash}}{OpenGVLab/}{VideoChat-Flash-Qwen2-7B\_res448} &
\EWmodelentry{nvidia-color}{LongVILA-R1-7B~\cite{longvilar1}}{Efficient-Large-Model/}{LongVILA-R1-7B} \\
\midrule
\rowcolor{tableband}
\multicolumn{3}{@{}l}{\strut\textbf{Closed-source models}} \\
\addlinespace[3pt]
\EWapientry{openai}{GPT-6.1 Sol~\cite{gpt61sol}}{gpt-6.1-sol} &
\EWapientry{gemini-color}{Gemini 4 Argon~\cite{gemini4argon}}{gemini-4-argon} &
\EWapientry{claude-color}{Claude Opus 5.5~\cite{claudeopus55}}{claude-opus-5-5} \\
\midrule
\rowcolor{tableband}
\multicolumn{3}{@{}l}{\strut\textbf{Prompted API controllers}} \\
\addlinespace[3pt]
\EWapientry{openai}{GPT-5.3 Chat~\cite{gpt53chat}}{azure/gpt-5.3-chat} &
\EWapientry{gemini-color}{Gemini 3.5 Flash-Lite~\cite{gemini35flashlite}}{gemini-3.5-flash-lite} &
\EWapientry{claude-color}{Claude Haiku 5.5~\cite{claudehaiku55}}{claude-haiku-5-5} \\
\bottomrule
\end{tabular}
\endgroup
\end{table*}

Direct VLMs receive the question, public reference images, and 32
timestamped video frames selected before inference from public task
metadata. Week-long object questions, Q01 to Q04, are posed in Chinese,
and week-long room questions in English. Each question's public time windows are intersected with the
recordings of its listed wearers, the available media are concatenated in
time order, and one frame is decoded at the midpoint of each of 32 equal
segments. The windows are $\pm$15\,s around each reference capture for Q01,
the stated 6-s interval of both wearers for Q02, the listed candidate
windows for Q04, the stated observation interval for Q07A, Q07C, and Q07E,
the 600\,s before the query time for Q07B and Q07D, and the full permitted
scope up to the query time for Q03, Q05, and Q06. Wearers share frames in
proportion to their recorded footage. For Q03, uniform frames over a scope
with a median of 112\,h never come within five minutes of the latest
sighting, so answering requires search, which motivates \agentname.
The selected timestamps and references are shared across models. Model-specific
preprocessing records decoded timestamps, resolution, and encoded frame
count. Gemini uses static video processing in this setting.
Direct outputs contain answers, while agents choose additional observation
intervals through their tools. Audio and transcript access follow the same
condition across direct models.

\Needspace{26\baselineskip}
\subsection{Agent execution}
\begin{wraptable}{r}{0.52\textwidth}
\captionsetup{font=footnotesize}
\centering\footnotesize
\caption{\textbf{Agent access and execution limits.} Shared settings for Base, SFT, and SFT+GRPO on the week-long questions.}
\label{tab:supp_access_budget}
\setlength{\tabcolsep}{4pt}
\begin{tabularx}{\linewidth}{@{}lY@{}}
\toprule \textbf{Attribute} & \textbf{Shared setting} \\
\midrule
\rowcolor{tableband}\multicolumn{2}{l}{\textit{Public inputs and shared services}} \\
Wearers / query time & Question-defined \\
Visual references & Public instance and room catalogs \\
Private GT / demonstrations & Excluded / 0 \\
Controller observations & Textual tool outputs \\
Visual backend & Gemini 3.6 Flash \\
Retrieval backend & GPT-5.3 Chat \\
\rowcolor{tableband}\multicolumn{2}{l}{\textit{Per-episode execution limits}} \\
Decisions / tool actions & 12 / 11 \\
Deadline / final reserve & 400 s / 60 s \\
Context / response tokens & 16,384 / 12,288 \\
Tokens per turn & 4,096 \\
Temperature / top-$p$ & 0 / 1 \\
Request-attempt cap & 48 \\
\bottomrule
\end{tabularx}
\par\vspace{2pt}\raggedright\footnotesize
Response tokens include tool observations. Retries share the episode budget
and remain subject to its tool-action limit.
\end{wraptable}
\agentname's Memory, Video, Image, Topology, and Timeline interfaces map to
\tool{search\_memory}, \tool{video\_qa\_updated}, \tool{image\_qa},
\tool{query\_topology}, and \tool{reduce\_timeline}.
The runtime also exposes \tool{validate\_evidence} for submission checks.
The controller receives textual tool observations and no task-specific
demonstrations. Native function calling and JSON adapters use the same
argument schema and returned observations. Malformed calls receive an error,
and retries consume the episode budget.

Memory retrieves coarse category information from captions, transcripts,
and annotation categories. Its public interface omits instance identifiers,
precise sightings, discriminative attributes, private answers, and room
timelines. Image and Video compare the selected media with public references.
\cref{tab:supp_tools} specifies the
inputs, outputs, and implementation of each tool, and
\cref{tab:supp_access_budget} gives the fixed perception and retrieval
backends and per-episode limits.

\par\WFclear
\begin{table*}[!htbp]
\centering\footnotesize
\caption{\textbf{\agentname\ tool implementations.} Each tool returns text
to the controller. Shaded bands separate tools that call fixed external
models from tools that compute their results deterministically.}
\label{tab:supp_tools}
\setlength{\tabcolsep}{4pt}
\renewcommand{\arraystretch}{1.12}
\begin{tabularx}{\textwidth}{@{}>{\raggedright\arraybackslash}p{0.16\textwidth}>{\raggedright\arraybackslash}p{0.19\textwidth}>{\raggedright\arraybackslash}p{0.17\textwidth}>{\raggedright\arraybackslash}X@{}}
\toprule
\textbf{Tool} & \textbf{Inputs} & \textbf{Returns} & \textbf{Implementation and access limits} \\
\midrule
\rowcolor{tableband}\multicolumn{4}{@{}l}{\textit{Retrieval and visual inspection with fixed external models}} \\
\textbf{Memory}\newline\tool{search\_memory} &
Text query, level of day, hour, or 10-min segment, optional wearer and time scope &
Relevance note, selected records, and video locators for segment hits &
No vector index is built. The text model GPT-5.3 Chat selects record IDs from batches of at most 30 prebuilt records, newest first, and IDs outside the batch are rejected. Records come from dense captions, transcripts, and coarse annotation categories. Instance IDs, attributes, boxes, exact sighting times, and answers are removed. \\
\arrayrulecolor{black!25}\cmidrule(lr){1-4}\arrayrulecolor{black}
\textbf{Video}\newline\tool{video\_qa\_updated} &
Prompt and sources such as public references, wearer intervals, or subclips &
Text observation, evidence IDs, and sampled time points &
Gemini 3.6 Flash receives one silent MP4 per request at 1\,FPS with source timestamps. Subclips of at most 40\,s use 5\,FPS, and object checks of at most 2\,s use 20\,FPS. Room IDs appear as permuted labels, and the room graph is withheld. Media outside the question scope or after the query time are rejected. \\
\arrayrulecolor{black!25}\cmidrule(lr){1-4}\arrayrulecolor{black}
\textbf{Image}\newline\tool{image\_qa} &
Prompt, reference images, and frames at given times &
Text observation and evidence IDs &
The same Gemini model compares target references and frames in one request of at most 50 images. Access limits match Video. \\
\addlinespace[2pt]
\rowcolor{tableband}\multicolumn{4}{@{}l}{\textit{Deterministic spatial and temporal computation}} \\
\textbf{Topology}\newline\tool{query\_topology} &
Start and end room IDs &
All tied shortest paths, or none when unreachable &
Breadth-first search on the unweighted room adjacency graph. It receives no visual input, so Video or Image must first identify the endpoint rooms. \\
\arrayrulecolor{black!25}\cmidrule(lr){1-4}\arrayrulecolor{black}
\textbf{Timeline}\newline\tool{reduce\_timeline} &
Start and end timestamps &
Duration in seconds &
Subtraction at millisecond resolution. Both times must lie within the question scope and before the query time. Video or Image must first locate the visit boundaries. \\
\bottomrule
\end{tabularx}
\end{table*}

\subsection{Questions and score aggregation}
\label{sec:supp_eval_scope}
The week-long comparison uses 1,320 questions in 1,154 core groups from the
1,533-question test partition. Each condition contributes one output
per question. Q01 to Q06 each contain 120 questions, and Q07 contains 600.
Micro accuracy weights questions equally. Macro accuracy first pools Q07A
to Q07E, then averages the seven ability scores. Failed submissions and
exhausted budgets remain in the denominator.

All direct-VLM and controller comparisons use this subset, and only the
text baseline also reports the full 1,533-question partition.

The within-environment split holds out core tasks within the week-long
residences. The cross-environment setting reserves all 787
questions and their environments for evaluation and uses the same
execution limits, as \cref{sec:supp_external_evaluation} specifies.
\cref{sec:supp_split_diagnostics} reports source-identifier overlap and
results on non-overlapping subsets.

\paragraph{Answer extraction.}
A shared deterministic normalizer reads only the model response and public
options. It accepts an explicit choice letter, a leading choice label with
answer text, a JSON choice field, or an exact unique option-text match.
It removes closed reasoning blocks and accepts one complete answer block,
including a final answer following discussion of other options.
Responses without a unique final choice count as incorrect. The same rules
apply to all 24 model-setting groups and 25,284 outputs, and open-weight
models decode greedily.
\cref{tab:supp_parse_audit} reports extraction rates.

\paragraph{Evidence scores.}
\label{sec:supp_evidence_metrics}
Semantic $E$ measures agreement with the fixed task-specific evidence
matcher. Continuous joint correctness is $J=A E$, where $A$ is answer
correctness. Strict success $H$ requires a correct answer and the task's
strict semantic match. This criterion is defined per task. All conditions use the same exact-citation
normalization and semantic scorer. Unresolved evidence lies in $[0,1]$,
while unresolved $J$ and $H$ lie in $[0,A]$. The reported bounds retain
all applicable questions in the denominator.

The execution score $E_g$ additionally applies the runtime provenance gate
to the original submissions. Its resolved mean uses the available judgments,
and its bounds include every applicable question. Format validity checks
the complete episode and is reported separately from semantic evidence
scores.

\paragraph{Data access and privacy.}
The release contains annotations and their provenance. Source-video access
follows the original providers' terms. Wearer IDs identify camera streams,
and witness answers describe supported camera observations. Because linked
locations and wearers can reveal daily routines, public examples and media
permissions require privacy review separately from annotation acceptance.
Run records associate results with question IDs, checkpoints, service
versions, and execution settings.

\section{Additional results}
\label{sec:recorded}

\subsection{Uncertainty of the main comparison}
\label{sec:supp_main_ci}
\cref{tab:supp_main_ci} gives 95\% intervals for the week-long accuracies in
main-paper Table~3, from 5,000 resamples of the 1,154 core groups. Paired
over the same resamples, \agentname{} + Sol exceeds Claude Opus 5.5 by 4.85
points, with interval $[1.55,8.08]$, and differs from Gemini 4 Argon by
1.29 points, with interval $[-1.88,4.46]$.

\begin{table*}[!t]
\captionsetup{font=footnotesize}
\begin{minipage}[t]{0.53\textwidth}
\vspace{0pt}
\centering\fontsize{7.2}{8.6}\selectfont
\caption{\textbf{Week-long accuracy with 95\% intervals.} Micro and Macro in
percent with percentile intervals over 5,000 core-group resamples.}
\label{tab:supp_main_ci}
\setlength{\tabcolsep}{2pt}
\begin{tabularx}{\linewidth}{@{}Yrr@{}}
\toprule System & Micro & Macro \\
\midrule
Public-text ridge & 31.59 $[29.19,34.08]$ & 32.12 $[29.19,35.08]$ \\
\rowcolor{tableband}\multicolumn{3}{@{}l}{\textit{Direct open-weight VLMs}} \\
\mlogo{qwen-color}Qwen3.8-27B & 46.97 $[43.98,50.00]$ & 47.71 $[44.44,50.99]$ \\
\mlogo{internlm-color}VideoChat-Flash-7B & 41.82 $[38.95,44.76]$ & 40.67 $[37.72,43.70]$ \\
\mlogo{xiaomi}MiMo-VL-7B-RL & 40.68 $[37.72,43.68]$ & 37.64 $[34.60,40.77]$ \\
\mlogo{alibaba-color}VideoLLaMA3-7B & 40.23 $[37.16,43.28]$ & 41.79 $[38.42,45.21]$ \\
\mlogo{internlm-color}InternVideo3-8B & 39.85 $[37.00,42.84]$ & 40.14 $[36.91,43.55]$ \\
\mlogo{internlm-color}InternVL3.5-8B & 39.70 $[36.77,42.64]$ & 40.76 $[37.38,44.23]$ \\
\mlogo{nvidia-color}LongVILA-R1-7B & 28.94 $[26.42,31.70]$ & 31.67 $[28.77,34.92]$ \\
\mlogo{google-color}Gemma 4 12B & 28.71 $[25.90,31.57]$ & 27.50 $[24.74,30.22]$ \\
\mlogo{qwen-color}Qwen3.5-4B & 28.11 $[25.76,30.52]$ & 27.79 $[25.17,30.43]$ \\
\rowcolor{tableband}\multicolumn{3}{@{}l}{\textit{Direct proprietary VLMs}} \\
\mlogo{gemini-color}Gemini 4 Argon & 73.56 $[70.15,76.98]$ & 73.31 $[70.35,76.55]$ \\
\mlogo{claude-color}Claude Opus 5.5 & 70.00 $[66.40,73.49]$ & 70.48 $[67.41,73.71]$ \\
\mlogo{openai}GPT-6.1 Sol & 58.79 $[55.54,62.07]$ & 60.67 $[57.59,63.82]$ \\
\rowcolor{tableband}\multicolumn{3}{@{}l}{\textit{Prompted shared-tool controllers}} \\
\mlogo{qwen-color}Qwen3.5-4B, base & 23.64 $[20.99,26.34]$ & 23.52 $[20.97,26.12]$ \\
\mlogo{openai}GPT-5.3 Chat & 65.76 $[62.25,69.24]$ & 61.71 $[58.31,65.29]$ \\
\mlogo{gemini-color}Gemini 4 Argon & 46.14 $[42.77,49.50]$ & 45.64 $[42.73,48.48]$ \\
\mlogo{gemini-color}Gemini 3.5 Flash-Lite & 37.58 $[34.48,40.58]$ & 40.19 $[36.94,43.80]$ \\
\mlogo{claude-color}Claude Haiku 5.5 & 32.20 $[29.42,35.19]$ & 34.50 $[31.64,37.39]$ \\
\rowcolor{tableband}\multicolumn{3}{@{}l}{\textit{Trained controllers and synthesis}} \\
\mlogo{egowitness-glasses}\agentname-SFT & 58.56 $[54.89,62.29]$ & 53.74 $[50.87,56.83]$ \\
\rowcolor{resultours}\mlogo{egowitness-glasses}\agentname-SFT+GRPO & 72.65 $[69.38,75.79]$ & 67.69 $[64.48,70.92]$ \\
\rowcolor{resultours}\mlogo{egowitness-glasses}\agentname{} + Sol & 74.85 $[71.89,77.77]$ & 69.81 $[66.66,72.91]$ \\
\bottomrule
\end{tabularx}
\end{minipage}\hfill
\begin{minipage}[t]{0.44\textwidth}
\vspace{0pt}
\centering\footnotesize
\caption{\textbf{\agentname\ by ability and visit subtype.} Answer accuracy
on the same 1,320 questions, 120 per detailed type. Base is the public 4B
controller. GRPO starts from SFT and uses step 259. $\Delta$ is GRPO minus SFT,
in percentage points. Macro$_7$ pools Q07 subtypes before averaging.}
\label{tab:supp_training_results}
\setlength{\tabcolsep}{2pt}
\begin{tabular*}{\linewidth}{@{\extracolsep{\fill}}lrr>{\columncolor{resultours}}rr@{}}
\toprule
& \multicolumn{3}{c}{\textbf{Answer accuracy (\%) $\uparrow$}} & \textbf{Gain} \\
\cmidrule(lr){2-4}\cmidrule(l){5-5}
Type & Base & SFT & +GRPO & $\Delta$ pp \\
\midrule
Q01 & 43.33 & 64.17 & \textbf{71.67} & \gain{+7.50} \\
Q02 & 49.17 & 61.67 & \textbf{68.33} & \gain{+6.67} \\
Q03 & 3.33 & 15.83 & \textbf{38.33} & \gain{+22.50} \\
Q04 & 20.00 & 32.50 & \textbf{53.33} & \gain{+20.83} \\
Q05 & 18.33 & 55.00 & \textbf{73.33} & \gain{+18.33} \\
Q06 & 6.67 & 80.00 & \textbf{87.50} & \gain{+7.50} \\
\rowcolor{tableband}\multicolumn{5}{l}{\textit{Q07 visit-reasoning subtypes}} \\
Q07A & 38.33 & 78.33 & \textbf{85.83} & \gain{+7.50} \\
Q07B & 12.50 & 85.00 & \textbf{90.00} & \gain{+5.00} \\
Q07C & 21.67 & 49.17 & \textbf{74.17} & \gain{+25.00} \\
Q07D & 30.83 & 76.67 & \textbf{91.67} & \gain{+15.00} \\
Q07E & 15.83 & 45.83 & \textbf{65.00} & \gain{+19.17} \\
\midrule
Micro & 23.64 & 58.56 & \textbf{72.65} & \gain{+14.09} \\
Macro$_7$ & 23.52 & 53.74 & \textbf{67.69} & \gain{+13.95} \\
\bottomrule
\end{tabular*}
\par\vspace{2pt}\raggedright\footnotesize
Best accuracy in each row is bold. \cref{sec:supp_joint_results} reports
per-type semantic joint scores.
\end{minipage}
\end{table*}

\subsection{Post-training results}
\label{sec:supp_training_diagnostics}
\begin{figure*}[!htbp]
\centering
\includegraphics[width=\textwidth]{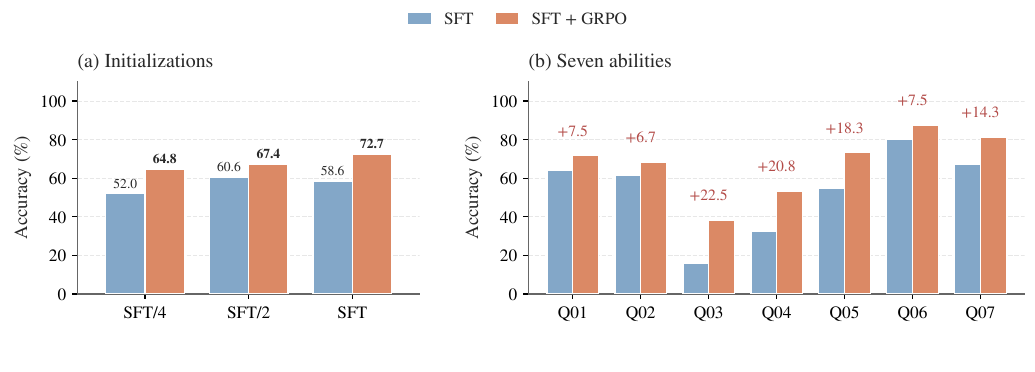}
\caption{\textbf{Post-training across initializations and abilities.}
Left compares micro accuracy before and after RL at three SFT
initializations under their recorded training schedules. Right gives
the SFT and step-259 SFT+RL comparison by ability.}
\label{fig:supp_initializations}
\end{figure*}

\begin{wrapfigure}{r}{0.43\textwidth}
\centering
\includegraphics[width=\linewidth]{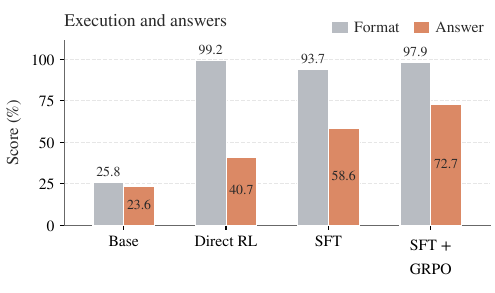}
\caption{\textbf{Execution quality and answers.} Format validity and answer accuracy on 1,320 questions.}
\label{fig:supp_quality}
\end{wrapfigure}

\cref{tab:supp_training_results} compares Base, SFT, and SFT+GRPO on the
same questions. GRPO improves answer accuracy over SFT on every detailed
type. The largest gains occur in completed-visit duration Q07C,
last-observer attribution Q03, and historical-window verification Q04.
\cref{tab:eight_recorded,fig:supp_initializations} compare additional
SFT initializations under the training schedules in \cref{sec:supp_optimization}.
RL raises every SFT initialization, by 12.80, 6.81, and 14.09 points for
SFT/4, SFT/2, and SFT, and all three end above the 40.68\% of Direct RL.
The 0.1-epoch initialization leads with 72.65\%.
\cref{fig:supp_quality} compares format validity with answer accuracy.
Direct RL reaches 99.24\% validity but 40.68\% accuracy, so valid output
format explains little of the SFT+GRPO gain.

\begin{table*}[!htbp]
\centering\small
\caption{\textbf{Post-training configurations on 1,320 questions.} Micro, Macro, and Format are percentages. $E_g$ is the resolved runtime-grounded evidence mean. SFT/2 and SFT/4 stop SFT at 0.05 and 0.025 epoch. Shading marks \agentname-SFT+GRPO (step 259), and bold marks the best value in each column. Metric definitions appear in \cref{sec:supp_evidence_metrics}.}
\label{tab:eight_recorded}
\setlength{\tabcolsep}{5pt}
\begin{tabularx}{\textwidth}{@{}l*{5}{>{\raggedleft\arraybackslash}X}@{}}
\toprule
 & \multicolumn{3}{c}{\textbf{Answers}} & \multicolumn{2}{c}{\textbf{Execution and evidence}} \\
\cmidrule(lr){2-4}\cmidrule(l){5-6}
\textbf{Controller} & Correct & Micro & Macro & Format & $E_g$ \\
\midrule
\rowcolor{tableband}\multicolumn{6}{@{}l}{\textit{Without SFT}} \\
Base Qwen3.5-4B~\cite{qwen35} & 312 & 23.64 & 23.52 & 25.76 & 0.099 \\
Direct RL, step 414 & 537 & 40.68 & 39.55 & 99.24 & 0.123 \\
\rowcolor{tableband}\multicolumn{6}{@{}l}{\textit{SFT initializations, before and after RL}} \\
SFT/4 & 686 & 51.97 & 52.14 & 81.52 & 0.414 \\
SFT/4 + RL & 855 & 64.77 & 62.74 & 99.09 & 0.456 \\
\addlinespace[2pt]
SFT/2 & 800 & 60.61 & 56.95 & 92.35 & 0.461 \\
SFT/2 + RL & 890 & 67.42 & 65.10 & \textbf{99.85} & 0.465 \\
\addlinespace[2pt]
SFT & 773 & 58.56 & 53.74 & 93.71 & 0.457 \\
\rowcolor{resultours}SFT + RL, step 259 & \textbf{959} & \textbf{72.65} & \textbf{67.69} & 97.88 & \textbf{0.482} \\
\bottomrule
\end{tabularx}
\end{table*}

\Needspace{22\baselineskip}
\subsection{Tool ablations}
\label{sec:supp_tool_removal}

\begin{figure*}[!htbp]
\centering
\includegraphics[width=\textwidth]{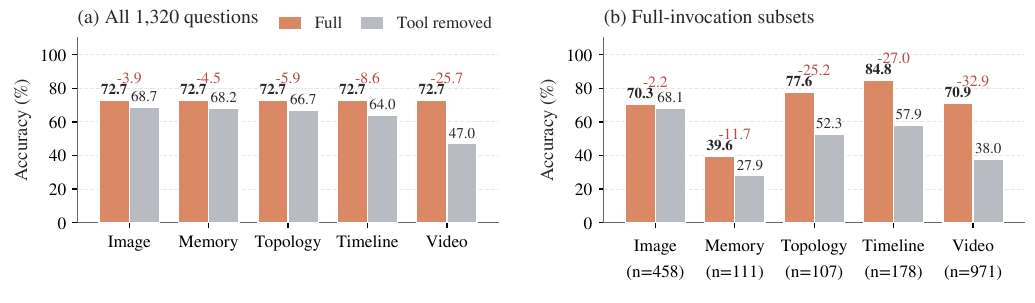}
\caption{\textbf{Tool removal on all questions and invocation-selected subsets.}
The left panel uses the same 1,320 questions for each removal. The right
panel uses a different Full-invocation subset for each tool. Subsets can
overlap and include unsuccessful calls.}
\label{fig:supp_ablation}
\end{figure*}

\cref{fig:supp_ablation,tab:supp_tool_transitions} compare individual
removals of Image, Memory, Topology, Timeline, and Video with the full agent.
The controller, visual backend, question IDs, and execution budget remain
fixed. Without Video, the controller cannot call the Video tool, while Image
keeps reference comparison and frame extraction at specified times. Without Image, independent image queries, standalone frame extraction,
and cached independent frames are disabled. Video retains public references
and can inspect them for reference-only tasks. Without the Topology query,
the supplied graph and transition checks remain available.

We report all questions, task-defined subsets, and subsets selected by
Full's tool calls. Task-defined subsets are Q05 for Topology and Q07B/Q07C
for Timeline. Invocation subsets include failed calls and can overlap.

\par\WFclear

\begin{table*}[!htbp]
\caption{\textbf{Paired tool-removal outcomes.} Full-invocation subsets are
defined from full-agent execution traces. Task-applicable subsets use question
types independently of observed calls. Full and Removed give correct-answer
counts. The final column is Removed minus Full accuracy in percentage points.}
\label{tab:supp_tool_transitions}
\centering\small
\setlength{\tabcolsep}{4pt}
\begin{tabularx}{\textwidth}{@{}lYrrrr@{}}
\toprule
Removed tool & Subset & Questions & Full & Removed & $\Delta$ pp \\
\midrule
Image & All questions & 1,320 & 959 & 907 & \loss{$-3.94$} \\
Memory & All questions & 1,320 & 959 & 900 & \loss{$-4.47$} \\
Topology & All questions & 1,320 & 959 & 881 & \loss{$-5.91$} \\
Timeline & All questions & 1,320 & 959 & 845 & \loss{$-8.64$} \\
Video & All questions & 1,320 & 959 & 620 & \loss{$-25.68$} \\
\midrule
\rowcolor{resultgroup}Topology & Q05 & 120 & 88 & 65 & \loss{$-19.17$} \\
\rowcolor{resultgroup}Timeline & Q07B / Q07C & 240 & 197 & 140 & \loss{$-23.75$} \\
\midrule
Image & Full-invocation & 458 & 322 & 312 & \loss{$-2.18$} \\
Memory & Full-invocation & 111 & 44 & 31 & \loss{$-11.71$} \\
Topology & Full-invocation & 107 & 83 & 56 & \loss{$-25.23$} \\
Timeline & Full-invocation & 178 & 151 & 103 & \loss{$-26.97$} \\
Video & Full-invocation & 971 & 688 & 369 & \loss{$-32.85$} \\
\bottomrule
\end{tabularx}
\end{table*}

On the task-defined subsets, removing Topology changes accuracy from
73.33\% to 54.17\%, and removing Timeline changes it from 82.08\% to 58.33\%.
Across all questions, 5,000 resamples of the 1,154 core groups give 95\%
percentile intervals of $[-6.33,-1.49]$, $[-7.13,-1.83]$, $[-8.49,-3.29]$,
$[-11.09,-6.08]$, and $[-28.86,-22.34]$ percentage points for Image, Memory,
Topology, Timeline, and Video removal, respectively. All five intervals
exclude zero.

\Needspace{17\baselineskip}
\subsection{Inference cost}
\label{sec:supp_cost}

\cref{tab:supp_resource_results} shows that GRPO lowers tool calls,
external-model calls, policy tokens, and episode latency relative to SFT.
Mean tool calls fall from 2.67 to 2.31, and mean latency from 175.71 to
151.59 seconds.

\cref{tab:supp_controller_cost} compares \agentname\ with the prompted API
controllers of main-paper Table~3. \agentname\ needs the fewest turns and
proprietary-model tokens. Its tool backends use about as many tokens as those
of GPT-5.3 Chat, so the saving comes from running the controller locally.
\cref{fig:supp_cost_bar} compares API fees at comparable accuracy.
\agentname{} + Sol reaches the highest accuracy, 74.85\%, for \$90, below
the \$110 and \$115 of direct Gemini 4 Argon and Claude Opus 5.5, and the
4B controller alone costs \$67 at 72.65\%.

\par\WFclear

\begin{table*}[!htbp]
\centering\footnotesize
\caption{\textbf{Per-question cost of shared-tool controllers.} All controllers
answer the same 1,320 week-long questions with the same tools, backends, and
budget. Turns count controller generations. Token counts include Gemini
thinking. $^{\dagger}$Local 4B controller, excluded from proprietary totals.
Bold marks the best accuracy, turns, controller tokens, and totals.}
\label{tab:supp_controller_cost}
\setlength{\tabcolsep}{4pt}
\begin{tabularx}{\textwidth}{@{}Yrrrrrrr@{}}
\toprule
Controller & Micro, \% & Turns & \makecell[r]{Tool\\calls} & \makecell[r]{Controller\\tokens} & \makecell[r]{Tool-backend\\tokens} & \makecell[r]{Proprietary\\total} & \makecell[r]{Proprietary\\per correct} \\
\midrule
\rowcolor{resultours}\mlogo{egowitness-glasses}\agentname-SFT+GRPO, 4B & \textbf{72.65} & \textbf{3.30} & 2.31 & \textbf{18.3k}$^{\dagger}$ & 38.1k & \textbf{38.1k} & \textbf{52.4k} \\
\mlogo{openai}GPT-5.3 Chat~\cite{gpt53chat} & 65.76 & 4.34 & 2.55 & 25.0k & 38.3k & 63.3k & 96.3k \\
\mlogo{gemini-color}Gemini 4 Argon~\cite{gemini4argon} & 46.14 & 4.83 & 2.78 & 32.1k & 28.9k & 61.0k & 132.2k \\
\mlogo{gemini-color}Gemini 3.5 Flash-Lite~\cite{gemini35flashlite} & 37.58 & 7.06 & 1.15 & 32.6k & 11.7k & 44.3k & 117.8k \\
\mlogo{claude-color}Claude Haiku 5.5~\cite{claudehaiku55} & 32.20 & 8.07 & 2.22 & 69.4k & 27.9k & 97.4k & 302.4k \\
\bottomrule
\end{tabularx}
\end{table*}

Per-question inference totals exclude one-time memory construction and
optional synthesis.

\begin{figure*}[!t]
\begin{minipage}[t]{0.44\textwidth}
\vspace{0pt}
\captionsetup{font=footnotesize}
\centering
\includegraphics[width=\linewidth]{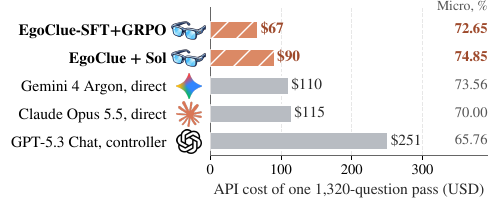}
\caption{\textbf{API cost at comparable accuracy.} Fees for one pass over the
1,320 questions for all systems above 65\% in main-paper Table~3, from recorded
tokens at November 2026 list prices. Gemini thinking is billed as output, and
GPT-5.3 Chat uses the \texttt{gpt-5.3} price. \agentname\ fees exclude
local 4B inference.}
\label{fig:supp_cost_bar}
\end{minipage}\hfill
\begin{minipage}[t]{0.53\textwidth}
\vspace{0pt}\captionsetup{type=table,font=footnotesize}
\centering\footnotesize
\caption{\textbf{Recorded inference resources.} SFT and GRPO use the same
1,320 questions. Changes are relative to SFT. Latency includes service
queueing, and controller tokens exclude visual-backend inputs.}
\label{tab:supp_resource_results}
\setlength{\tabcolsep}{3pt}
\begin{tabularx}{\linewidth}{@{}Y r >{\columncolor{resultours}}r r@{}}
\toprule
Metric & SFT & +GRPO & $\Delta$ \\
\midrule
\rowcolor{tableband}\multicolumn{4}{l}{\textit{Calls and tokens per QA: relative change, \%}} \\
Tool calls & 2.67 & \textbf{2.31} & \gain{$-13.37$} \\
External model calls & 1.90 & \textbf{1.59} & \gain{$-16.19$} \\
Policy tokens & 635.5 & \textbf{496.2} & \gain{$-21.92$} \\
Observation tokens & 3,524.5 & \textbf{3,459.5} & \gain{$-1.84$} \\
\rowcolor{tableband}\multicolumn{4}{l}{\textit{Episode latency, seconds: relative change, \%}} \\
Mean & 175.71 & \textbf{151.59} & \gain{$-13.73$} \\
Median & 140.44 & \textbf{118.78} & \gain{$-15.43$} \\
95th percentile & 360.36 & \textbf{352.63} & \gain{$-2.15$} \\
\bottomrule
\end{tabularx}
\par\vspace{2pt}\raggedright\footnotesize
Changes use unrounded measurements.
\end{minipage}
\end{figure*}

\subsection{Nonvisual baselines and training overlap}
\label{sec:supp_reviewer_audits}
Uniform guessing has expected correctness $1/k_i$ for a question with
$k_i$ options. The week-long set contains 240 three-option and 1,080
four-option questions, yielding 26.52\% Micro and 27.38\% Macro accuracy.
The cross-environment set contains 200 three-option and 587 four-option
questions, yielding 27.12\% expected Micro accuracy.

\paragraph{Answer extraction.}
\cref{tab:supp_parse_audit} separates response formatting from answer
accuracy using the normalizer in \cref{sec:supp_eval_scope}.
Parsed records successful choice extraction, and all questions contribute
to accuracy.

\begin{table}[!htbp]
\centering\footnotesize
\caption{\textbf{Direct-output normalization.} Parsed is the percentage
assigned a choice by the common normalizer. $A$ is normalized accuracy on all
1,320 week-long and 787 cross-environment questions, as reported in the main model comparison and \cref{tab:transfer}.}
\label{tab:supp_parse_audit}
\setlength{\tabcolsep}{3pt}
\begin{tabular*}{\linewidth}{@{\extracolsep{\fill}}lrrrr@{}}
\toprule
& \multicolumn{2}{c}{Week-long} & \multicolumn{2}{c}{Cross-env.} \\
\cmidrule(lr){2-3}\cmidrule(l){4-5}
Model & Parsed & $A$ & Parsed & $A$ \\
\midrule
\mlogo{qwen-color}Qwen3.8-27B~\cite{qwen38} & 92.80 & 46.97 & 99.75 & 58.07 \\
\mlogo{internlm-color}VideoChat-Flash-7B~\cite{videochatflash} & 100.00 & 41.82 & 100.00 & 38.37 \\
\mlogo{xiaomi}MiMo-VL-7B-RL~\cite{mimovl} & 99.39 & 40.68 & 99.75 & 46.25 \\
\mlogo{alibaba-color}VideoLLaMA3-7B~\cite{videollama3} & 100.00 & 40.23 & 100.00 & 47.78 \\
\mlogo{internlm-color}InternVideo3-8B~\cite{internvideo3} & 100.00 & 39.85 & 100.00 & 44.09 \\
\mlogo{internlm-color}InternVL3.5-8B~\cite{internvl35} & 100.00 & 39.70 & 100.00 & 47.90 \\
\mlogo{nvidia-color}LongVILA-R1-7B~\cite{longvilar1} & 90.30 & 28.94 & 87.04 & 28.46 \\
\mlogo{google-color}Gemma 4 12B~\cite{gemma4} & 69.47 & 28.71 & 88.56 & 49.81 \\
\mlogo{qwen-color}Qwen3.5-4B~\cite{qwen35} & 82.05 & 28.11 & 96.06 & 41.93 \\
\midrule
\mlogo{gemini-color}Gemini 4 Argon~\cite{gemini4argon} & 99.92 & 73.56 & 100.00 & 86.28 \\
\mlogo{claude-color}Claude Opus 5.5~\cite{claudeopus55} & 99.92 & 70.00 & 99.75 & 80.81 \\
\mlogo{openai}GPT-6.1 Sol~\cite{gpt61sol} & 100.00 & 58.79 & 100.00 & 67.60 \\
\bottomrule
\end{tabular*}
\end{table}

\Needspace{24\baselineskip}
\paragraph{Public-text prediction.}
\begin{wraptable}{r}{0.49\textwidth}
\captionsetup{font=footnotesize}
\centering\footnotesize
\caption{\textbf{Nonvisual baselines on the week-long setting.}
Majority selects the most frequent training answer per detailed type.
Ridge uses public text and validation-selected regularization. Q07 pools
five subtypes. These scores use the same question sets as the main model
comparison.}
\label{tab:supp_text_only}
\setlength{\tabcolsep}{4pt}
\begin{tabularx}{\linewidth}{@{}Yrrr@{}}
\toprule Ability & Questions & Majority & Ridge \\
\midrule
Q01 & 120 & 30.83 & 42.50 \\
Q02 & 120 & 18.33 & 28.33 \\
Q03 & 120 & 22.50 & 27.50 \\
Q04 & 120 & 25.00 & 27.50 \\
Q05 & 120 & 25.00 & 25.00 \\
Q06 & 120 & 35.00 & 43.33 \\
Q07 & 600 & 25.00 & 30.67 \\
\midrule
\rowcolor{resultours}Main set, Micro & 1,320 & 25.61 & \textbf{31.59} \\
Main set, Macro & 1,320 & 25.95 & 32.12 \\
Full test, Micro & 1,533 & \blank & 32.35 \\
Full test, Macro & 1,533 & \blank & 32.82 \\
\bottomrule
\end{tabularx}
\end{wraptable}
We fit one TF-IDF ridge classifier per detailed task type using the public
English question, option text, and type. Word unigrams and bigrams use
training-only vocabulary and IDF statistics, a minimum document frequency
of two, and at most 30,000 features per type. Features use sublinear term
frequency, smoothed IDF, and row-wise $\ell_2$ normalization. One-versus-rest
ridge regression uses targets in $\{-1,+1\}$ and a regularized intercept.

We fit on 5,750 training questions and select a regularization coefficient
from $\{0.1,1,10\}$ by seven-ability macro accuracy on 383 validation
questions. Ties favor validation micro accuracy, then the larger coefficient.
The selected value is 1, giving validation micro and macro accuracy of
30.03\% and 29.03\%. The training-only fit is retained for testing.
\cref{tab:supp_text_only} compares it with the most frequent training
answer for each detailed type.

The classifier answers 417/1,320 questions correctly, compared with
338/1,320 for majority prediction. On the 787 cross-environment questions,
the same classifier scores 27.06\%, close to the 27.12\% expected from
uniform guessing.

Permuting the options of each test question 20 times lowers the week-long
Micro accuracy of the fitted classifier to $27.87\pm1.74\%$, close to uniform
guessing. Q02, whose windows all last 6\,s and whose options appear in all
24 orders, stays at $27.54\pm3.98\%$. Q01 drops from 42.50\% to
$34.75\pm4.22\%$ because some option orders co-occur with the answer
position, while Gemini 4 Argon and Claude Opus 5.5 differ by at most
2.3 points between questions with and without these orders.

\par\WFclear
\Needspace{22\baselineskip}
\paragraph{Training-overlap sensitivity.}
\label{sec:supp_split_diagnostics}
\begin{wraptable}{r}{0.49\textwidth}
\captionsetup{font=footnotesize}
\centering\footnotesize
\caption{\textbf{Train/test identifier overlap.} Recomputed from
assignment atoms. Shared atoms are distinct identifiers. Question counts
refer to the evaluated 1,320 and may overlap across rows.}
\label{tab:supp_split_atoms}
\setlength{\tabcolsep}{4pt}
\begin{tabular*}{\linewidth}{@{\extracolsep{\fill}}lrr@{}}
\toprule
Identifier type & \makecell[r]{Shared\\atoms} & \makecell[r]{Affected\\questions} \\
\midrule
Core group & 0 & 0 \\
Exact public input & 0 & 0 \\
Object episode & 503 & 364 \\
Object event & 209 & 445 \\
Reference & 566 & 562 \\
Room anchor & 229 & 227 \\
Room boundary & 2 & 2 \\
Room dwell & 2,335 & 576 \\
Room transition & 1,487 & 499 \\
\bottomrule
\end{tabular*}
\end{wraptable}
\cref{tab:supp_split_atoms} counts shared source identifiers between
training and test data. Core groups and exact public inputs are disjoint,
while references and source events can recur. These identifiers describe
source-record reuse. An answer fact also depends on the target, queried
relation, wearer scope, and time interval.

\cref{tab:supp_overlap_sensitivity} retains questions without training-shared
identifiers and selects these subsets from identifier manifests alone. Reference filtering applies to questions with reference
identifiers, object-event filtering to object tasks, and room-event
filtering to room tasks. The checkpoints and question inputs stay fixed.

\par\WFclear
\Needspace{17\baselineskip}
\begin{wraptable}{r}{0.52\textwidth}
\captionsetup{font=footnotesize}
\centering\footnotesize
\caption{\textbf{Training-overlap sensitivity.} Each row retains questions
with no training-shared atom of the indicated kind, within its applicable
domain. Checkpoints and inputs are unchanged. Values are micro accuracy
and the GRPO gain in percentage points. Rows can overlap.}
\label{tab:supp_overlap_sensitivity}
\setlength{\tabcolsep}{3pt}
\begin{tabularx}{\linewidth}{@{}Yrrrr@{}}
\toprule Unshared identifiers & $N$ & SFT & GRPO & $\Delta$ \\
\midrule
Reference & 114 & 35.09 & 56.14 & \gain{+21.05} \\
Object event or episode & 32 & 31.25 & 65.63 & \gain{+34.38} \\
Room event & 81 & 76.54 & 88.89 & \gain{+12.35} \\
\rowcolor{resultours}All recorded atoms & 113 & 63.72 & 82.30 & \gain{+18.58} \\
\bottomrule
\end{tabularx}
\end{wraptable}

The final row checks all recorded identifier types except the already
disjoint exact-public-input hash. It contains 113 questions in 93 core
groups. SFT answers 72 correctly and GRPO answers 93. Two thousand paired
core-group resamples give a 95\% gain interval of $[6.14,30.83]$ percentage
points. Public-text ridge, direct Gemini, and GRPO with synthesis score
36.28\%, 74.34\%, and 87.61\% on the same subset.

Q01 to Q06 contribute 4, 4, 18, 6, 4, and 9 questions. Q07A to Q07E
contribute 14, 21, 1, 31, and 1.

\par\WFclear
\subsection{Answer and evidence agreement}
\label{sec:supp_joint_results}
\cref{tab:supp_joint_audit} evaluates SFT, GRPO, and synthesis on the same
1,021 evidence-applicable questions using the semantic scores defined in
\cref{sec:supp_evidence_metrics}. The aggregate $E$ bounds are $[46.73,49.18]\%$, $[51.56,54.89]\%$,
and $[51.61,55.63]\%$. GRPO's aggregate $J$ lower bound exceeds the SFT
upper bound.

\begin{table*}[!htbp]
\centering\small
\caption{\textbf{Semantic joint correctness by week-long type.} Values are
percentages on the fixed applicable subset. Ranges bound unresolved scores.
Q01 and Q04 are answer-only under the frozen object contract. Q02 excludes
25 evidence-inapplicable cases and Q07E excludes 34, identically across runs.
Scores include Q03 exact-citation normalization, applied by
the same rule to SFT, GRPO, and synthesis with the frozen scorer.}
\label{tab:supp_joint_audit}
\setlength{\tabcolsep}{4pt}
\begin{tabular}{@{}lrcccccc@{}}
\toprule
& & \multicolumn{3}{c}{Continuous joint $J$} & \multicolumn{3}{c}{Strict semantic success $H$} \\
\cmidrule(lr){3-5}\cmidrule(l){6-8}
Type & Applicable & SFT & GRPO & + Sol & SFT & GRPO & + Sol \\
\midrule
Q02 & 95 & 34.74--49.47 & 33.68--47.37 & 43.16--61.05 & 34.74--49.47 & 33.68--47.37 & 43.16--61.05 \\
Q03 & 120 & 0.00--8.33 & 0.00--17.50 & 0.00--20.00 & 0.00--8.33 & 0.00--17.50 & 0.00--20.00 \\
Q05 & 120 & 40.00 & 52.08 & 54.17 & 29.17 & 37.50 & 39.17 \\
Q06 & 120 & 56.67 & 62.92 & 60.00 & 45.00 & 52.50 & 51.67 \\
Q07A & 120 & 61.66 & 69.79 & 68.95 & 55.83 & 61.67 & 62.50 \\
Q07B & 120 & 44.84 & 47.86 & 47.82 & 56.67 & 60.00 & 60.00 \\
Q07C & 120 & 33.45 & 47.05 & 47.49 & 33.33 & 46.67 & 47.50 \\
Q07D & 120 & 59.89 & 69.67 & 68.74 & 65.83 & 75.00 & 72.50 \\
Q07E & 86 & 15.36 & 30.95 & 35.44 & 10.47 & 24.42 & 27.91 \\
\midrule
\rowcolor{resultours}All & 1,021 & 39.37--41.72 & 46.80--50.13 & 47.80--51.82 & 37.71--40.06 & 44.37--47.70 & 45.54--49.56 \\
\bottomrule
\end{tabular}
\end{table*}

\paragraph{Latest-witness evidence.}
Q03 requires the observing wearer and the latest supporting sighting.
The common citation normalizer restores a submission only when its role,
wearer, date, and exact time window uniquely match evidence returned by a
successful tool call in that trajectory. It preserves the predicted answer,
window, and observation. Ambiguous matches remain unresolved.

Visual evidence must establish both target identity and temporal recency.
A visible earlier sighting establishes presence but does not establish the
latest sighting. Unresolved visibility or recency judgments retain the
bounds in \cref{tab:supp_joint_audit}. Missing recordings remain unknown,
and coverage checks range over the permitted wearers up to the query time.

\section{Answer synthesis}
\label{sec:supp_synthesis}

\subsection{Synthesis inputs}
\agentname{} + Sol passes the controller's tool observations and preceding
decisions to GPT-6.1 Sol after removing the terminal answer turn.
Sol produces the final choice and evidence references without new visual
calls or additional tools. It receives no private answers or scores.
Decoding uses a 4,096-token output limit.
The instruction requests chronological event consolidation, consistent visit
counting, and evidence-role alignment, with unknown intervals distinguished
from absence.

For episodes without a submitted answer, synthesis reads the recorded
trajectory in the same way.

\subsection{Cross-environment setup}
\label{sec:supp_external_evaluation}
All 787 cross-environment questions and their environments are excluded
from SFT and GRPO. All tool systems use the week-long execution budget in
\cref{tab:supp_access_budget}. Gemini 3.6 Flash remains the visual backend. Image, Video, Timeline, and submission validation are
available. There is no prebuilt Memory or supplied Topology graph.

\cref{tab:transfer} compares the twelve direct VLMs and the
prompted shared-tool controllers with this week-long-trained system before
and after synthesis. Without training in these environments,
\agentname-SFT+GRPO exceeds every prompted controller and every open-weight
direct VLM, 78.40\% against at most 71.54\% and 58.07\%.
GRPO raises \agentname\ from 68.11\% after SFT to 78.40\%, with Q04 rising
from 43.20\% to 85.21\%.

\begin{table*}[!t]
\centering\small
\caption{\textbf{Cross-environment model and system results.} All 787 questions
are evaluation-only. Q01, Q04, Q06, and Q07 contain 120, 169, 80, and 418
questions. Macro weights these four abilities equally. Micro weights all
questions equally. Direct models answer from their frame inputs, and all tool
systems use the week-long execution budget. All direct rows use the same
normalizer. Bold marks the best Micro within each group.}
\label{tab:transfer}
\setlength{\tabcolsep}{5pt}
\begin{tabularx}{\textwidth}{@{}Yrrrrrr@{}}
\toprule Model or system & Q01 & Q04 & Q06 & Q07 & Macro & Micro \\
\midrule
\rowcolor{resultgroup}\multicolumn{7}{@{}l}{\textit{Direct visual reasoning, ordered by Micro}} \\
\mlogo{gemini-color}Gemini 4 Argon~\cite{gemini4argon} & 90.00 & 95.27 & 83.75 & 82.06 & 87.77 & \textbf{86.28} \\
\mlogo{claude-color}Claude Opus 5.5~\cite{claudeopus55} & 77.50 & 94.08 & 80.00 & 76.56 & 82.03 & 80.81 \\
\mlogo{openai}GPT-6.1 Sol~\cite{gpt61sol} & 70.83 & 73.96 & 56.25 & 66.27 & 66.83 & 67.60 \\
\mlogo{qwen-color}Qwen3.8-27B~\cite{qwen38} & 53.33 & 76.33 & 62.50 & 51.20 & 60.84 & 58.07 \\
\mlogo{google-color}Gemma 4 12B~\cite{gemma4} & 49.17 & 55.62 & 53.75 & 46.89 & 51.36 & 49.81 \\
\mlogo{internlm-color}InternVL3.5-8B~\cite{internvl35} & 57.50 & 71.01 & 52.50 & 34.93 & 53.98 & 47.90 \\
\mlogo{alibaba-color}VideoLLaMA3-7B~\cite{videollama3} & 65.00 & 43.79 & 48.75 & 44.26 & 50.45 & 47.78 \\
\mlogo{xiaomi}MiMo-VL-7B-RL~\cite{mimovl} & 30.00 & 86.39 & 42.50 & 35.41 & 48.57 & 46.25 \\
\mlogo{internlm-color}InternVideo3-8B~\cite{internvideo3} & 49.17 & 65.68 & 37.50 & 35.17 & 46.88 & 44.09 \\
\mlogo{qwen-color}Qwen3.5-4B~\cite{qwen35} & 25.83 & 57.40 & 53.75 & 38.04 & 43.75 & 41.93 \\
\mlogo{internlm-color}VideoChat-Flash-7B~\cite{videochatflash} & 50.00 & 50.89 & 21.25 & 33.25 & 38.85 & 38.37 \\
\mlogo{nvidia-color}LongVILA-R1-7B~\cite{longvilar1} & 25.83 & 38.46 & 42.50 & 22.49 & 32.32 & 28.46 \\
\midrule
\rowcolor{resultgroup}\multicolumn{7}{@{}l}{\textit{Prompted shared-tool controllers, ordered by Micro}} \\
\mlogo{gemini-color}Gemini 4 Argon~\cite{gemini4argon} & 69.17 & 72.19 & 86.25 & 69.14 & 74.19 & \textbf{71.54} \\
\mlogo{openai}GPT-5.3 Chat~\cite{gpt53chat} & 70.00 & 76.92 & 65.00 & 64.11 & 69.01 & 67.85 \\
\mlogo{gemini-color}Gemini 3.5 Flash-Lite~\cite{gemini35flashlite} & 60.00 & 59.17 & 61.25 & 46.17 & 56.65 & 52.60 \\
\mlogo{claude-color}Claude Haiku 5.5~\cite{claudehaiku55} & 57.50 & 52.07 & 48.75 & 46.41 & 51.18 & 49.56 \\
\mlogo{qwen-color}Qwen3.5-4B~\cite{qwen35} (base) & 22.50 & 15.38 & 21.25 & 25.12 & 21.06 & 22.24 \\
\midrule
\rowcolor{resultgroup}\multicolumn{7}{@{}l}{\textit{Week-long-trained systems, target environments excluded from post-training}} \\
\mlogo{egowitness-glasses}\agentname-SFT & 86.67 & 43.20 & 81.25 & 70.33 & 70.36 & 68.11 \\
\mlogo{egowitness-glasses}\agentname-SFT+GRPO & 87.50 & 85.21 & 83.75 & 72.01 & 82.12 & 78.40 \\
\rowcolor{resultours}\mlogo{egowitness-glasses}\agentname{} + Sol~\cite{gpt61sol} & 84.17 & 88.17 & 81.25 & 76.32 & 82.47 & \textbf{80.56} \\
\bottomrule
\end{tabularx}
\end{table*}

\subsection{Answer and evidence gains}
\begin{figure*}[!htbp]
\centering
\includegraphics[width=0.94\textwidth]{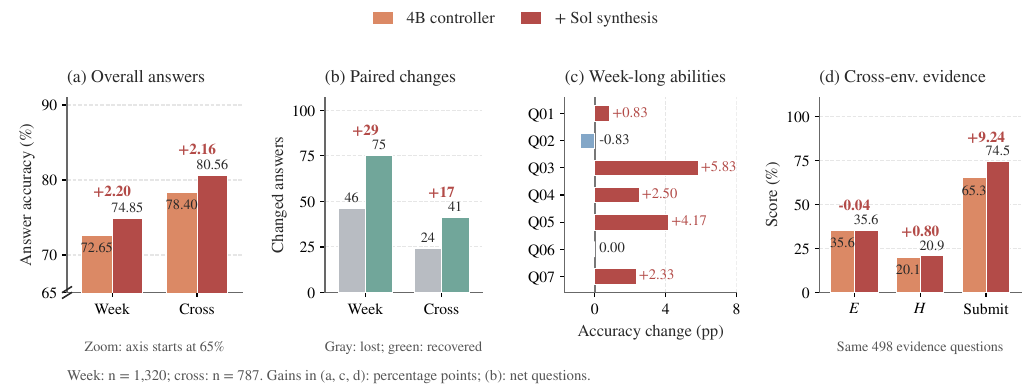}
\caption{\textbf{Evidence-only synthesis.}
\textbf{(a)} Answer accuracy.
\textbf{(b)} Incorrect-to-correct and correct-to-incorrect answer changes.
\textbf{(c)} Accuracy change by week-long ability.
\textbf{(d)} Semantic evidence $E$, strict success $H$, and evidence-submission
validity on the same 498 cross-environment questions.
Accuracy comparisons use 1,320 week-long and 787 cross-environment questions
under their respective controller settings. Orange denotes the controller
and red denotes synthesis. Gains in (a), (c), and (d) are percentage points.}
\label{fig:supp_synthesis}
\end{figure*}

\cref{tab:supp_synthesis,fig:supp_synthesis} compare answer accuracy before
and after synthesis. On the week-long set, 5,000 paired resamples of
1,154 core groups give a 95\% gain interval of $[0.59,3.82]$ percentage
points.

\begin{table*}[!htbp]
\centering\small
\caption{\textbf{Answer synthesis.} Corrected and regressed count paired answer changes.}
\label{tab:supp_synthesis}
\setlength{\tabcolsep}{6pt}
\begin{tabular}{@{}lrrrrrr@{}}
\toprule
Evaluation scope & Questions & 4B, \% & + Sol, \% & $\Delta$, pp & Corrected & Regressed \\
\midrule
\rowcolor{resultours}Week-long & 1,320 & 72.65 & \textbf{74.85} & \gain{+2.20} & 75 & 46 \\
\rowcolor{resultours}Cross-environment & 787 & 78.40 & \textbf{80.56} & \gain{+2.16} & 41 & 24 \\
\bottomrule
\end{tabular}
\end{table*}

For the 1,021 week-long evidence-applicable questions, semantic $E$ bounds
are $[51.56,54.89]\%$ before and $[51.61,55.63]\%$ after synthesis.
\cref{tab:supp_joint_audit} gives the per-type joint results.

\Needspace{18\baselineskip}
\begin{wraptable}{r}{0.50\textwidth}
\captionsetup{font=footnotesize}
\centering\footnotesize
\caption{\textbf{Video-cluster uncertainty for synthesis.} Changes are
percentage points relative to the 4B controller. CI denotes the 95\%
paired video-bootstrap interval. Evidence metrics use their applicable
subset, with each metric's denominator listed below.}
\label{tab:supp_cross_paired}
\setlength{\tabcolsep}{4pt}
\begin{tabularx}{\linewidth}{@{}Yrrc@{}}
\toprule Metric & Questions & $\Delta$ & 95\% CI \\
\midrule
Answer $A$ & 787 & \gain{+2.16} & $[0.11,4.65]$ \\
Semantic $E$ & 498 & $-0.04$ & $[-0.93,0.86]$ \\
Joint $J$ & 498 & \gain{+1.09} & $[0.42,1.81]$ \\
Strict $H$ & 498 & +0.80 & $[0.00,1.80]$ \\
\bottomrule
\end{tabularx}
\end{wraptable}
On the same 498 cross-environment evidence-applicable questions, semantic
$E$ stays at 35.6\%, while joint $J$ rises from 31.65\% to 32.74\% and
strict success $H$ from 20.08\% to 20.88\%.
Evidence-submission validity increases from 325/498 to 371/498,
or 65.26\% to 74.50\%. This validity check concerns submission structure
and required fields, while the semantic scores assess the evidence match.

We pair question IDs and bootstrap source videos 3,000 times, retaining
all questions from each sampled video. Answer evaluation spans 40 videos,
and evidence evaluation spans 37. \cref{tab:supp_cross_paired} reports
95\% percentile intervals. The answer and joint evidence gains exclude zero.

\par\WFclear

\subsection{Synthesis cost}
The 1,320 week-long synthesis requests use 10,239,769 prompt tokens and
252,425 completion tokens. The 787 cross-environment requests use
8,413,064 and 90,231, respectively. Cross-environment synthesis adds a
mean 4.95 seconds per question, with median 4.30 and 95th percentile
7.54 seconds. These measurements cover the additional synthesis call.

\begingroup
\small
\section{Case studies and error analysis}
\label{sec:supp_case_studies}
This section examines how models fail on the seven abilities. The eleven
detailed types form two families: object-instance memory (Q01--Q04)
and room and visit memory (Q05, Q06, and Q07A--Q07E). Every example reports
the same models: the three proprietary frontier VLMs of the main results
table under direct access, and \agentname-SFT+GRPO with its 4B controller.
Object questions were posed in Chinese, and the figures show English
renderings.

\subsection{Representative failure cases}
\label{sec:supp_case_failures}
\Cref{fig:supp_case_failures} gives one example for each of six abilities,
grouped by the three error types of the question design: identity recognition,
historical retrieval, and spatial-temporal reasoning.

\paragraph{Type I: Identity recognition.}
These errors confuse physical identity with appearance.
\begin{itemize}\setlength{\itemsep}{1pt}
\item Q01: \emph{Are the main objects in Image A and Image B the same physical
instance?} All three frontier VLMs answer ``same physical instance''; the
ground truth is ``same category, different instances''. Both images show a
plain white projection screen, but one hangs in the ground-floor living room
and the other in the first-floor living room.
\item Q06: \emph{Do Image A and Image B show the same physical room?} All three
answer ``same type, different rooms''; both views show \texttt{living\_1f\_0}.
The views share no visible surface, so the match must come from the room's
fixed layout.
\end{itemize}

\paragraph{Type II: Historical retrieval.}
These errors stop at an earlier sighting or skip a short one.
\begin{itemize}\setlength{\itemsep}{1pt}
\item Q03: \emph{As of DAY7 11:16:01, who most recently recorded the projection
screen?} Gemini 4 Argon and GPT-6.1 Sol answer A4, who last recorded it on
DAY6 evening; Claude Opus 5.5 and the ground truth give A6, who recorded it
again at DAY7 11:03:33.
\item Q04: \emph{Which of four windows contain the lamp?} All three omit W3.
That window is mostly kitchen and living-room activity, and the wearer enters
the lamp's bedroom only once, for 89\,s.
\end{itemize}

\paragraph{Type III: Spatial-temporal reasoning.}
These errors lose room boundaries or graph structure.
\begin{itemize}\setlength{\itemsep}{1pt}
\item Q05: \emph{What is the shortest path between the rooms at two references?}
All three route through \texttt{living\_gf\_0}, which has no edge to either
first-floor endpoint; the endpoints are adjacent.
\item Q07E: \emph{How many separate visits did A2 make to the bedroom?} All
three answer one visit; the reviewed timeline contains three, separated by
trips of 11.5\,s and 26.4\,s.
\end{itemize}

\begin{figure*}[!htbp]
\centering
\includegraphics[width=\textwidth]{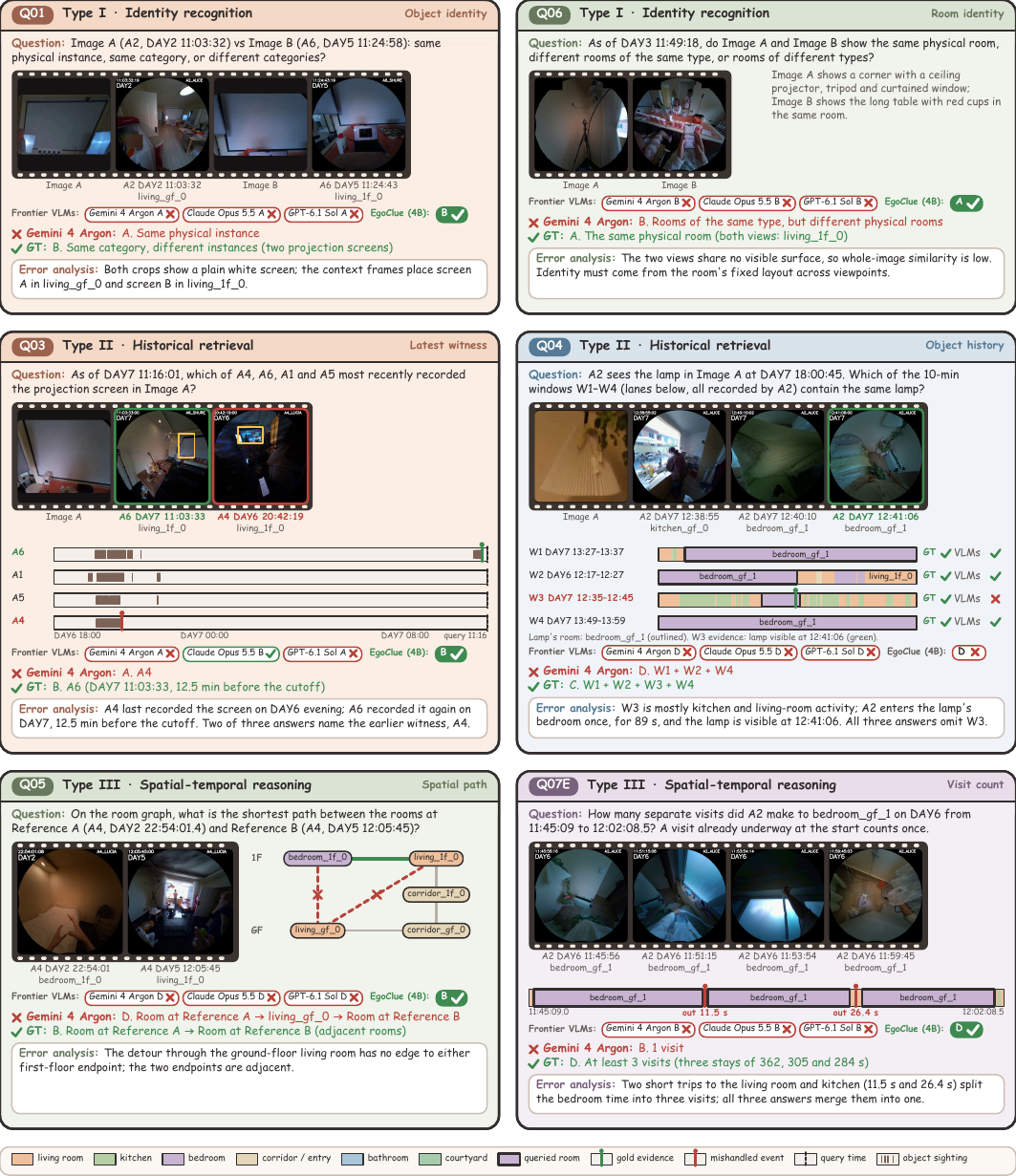}
\caption{\textbf{Representative failure cases.} Six abilities grouped by error
type. Chips report the three frontier VLMs and \agentname{} (green when
correct), and the red line repeats Gemini 4 Argon's option. Green and red
frame outlines mark gold evidence and the frame behind a wrong answer; yellow
boxes are released masks at the same timestamp. All timelines share the legend
below: room bands come from reviewed room intervals, and Q03 ticks are released
screen masks in 1-min bins. Q06 references show one of five views.}
\label{fig:supp_case_failures}
\end{figure*}

\Needspace{22\baselineskip}
\subsection{Insights and implications}
\label{sec:supp_case_insights}
\begin{wraptable}{r}{0.52\textwidth}
\captionsetup{font=footnotesize}
\centering\footnotesize
\setlength{\tabcolsep}{3pt}
\caption{\textbf{Frontier-VLM error patterns by question type.} Error pools the
stored direct predictions of Gemini 4 Argon, Claude Opus 5.5, and GPT-6.1 Sol
(358--360 answered per type). Shares are fractions of errors.}
\label{tab:supp_error_patterns}
\begin{tabularx}{\linewidth}{@{}lrY@{}}
\toprule
Type & Err.\,(\%) & Dominant error direction \\
\midrule
\rowcolor{tableband}\multicolumn{3}{@{}l}{\textit{Object-instance memory}} \\
Q01 & 31.1 & different categories 44\%; merge 36\% \\
Q02 & 19.7 & over-report 89\%; under-report 11\% \\
Q03 & 69.4 & -- \\
Q04 & 29.3 & omit a positive window 56\% \\
\rowcolor{tableband}\multicolumn{3}{@{}l}{\textit{Room and visit memory}} \\
Q05 & 16.2 & shorter 48\%; longer 45\% \\
Q06 & 23.6 & type confusion 49\%; split 28\% \\
Q07A & 16.1 & extra rooms 52\%; missing 36\% \\
Q07B & 17.2 & overestimate 68\% \\
Q07C & 41.5 & underestimate 55\% \\
Q07D & 38.6 & -- \\
Q07E & 55.3 & undercount 77\% \\
\bottomrule
\end{tabularx}
\end{wraptable}
\Cref{tab:supp_error_patterns} counts error directions of the three
frontier VLMs on all 120 questions per type. Three gaps follow.

\paragraph{Identity without place.}
Q01 errors split among category confusions (44\%), merging two instances
(36\%), and splitting one instance (21\%). These errors are consistent with
appearance matching that ignores instance location.

\paragraph{Retrieval stops early.}
Q03 has the highest frontier error, 69.4\%. In Q04, 56\% of errors omit a
positive window.

\paragraph{Short exits split visits.}
In Q07E, 77\% of errors undercount visits, and in Q07B, 68\% overestimate the
current stay. The exits separating Q07E visits last a median of 92.5\,s, and
31\% last at most 30\,s.

\paragraph{Directions.}
The gaps suggest three requirements: verifying instance identity together with
location, retrieving over every candidate window and the interval before the
query time, and computing durations and counts from explicit room boundaries.
\agentname{} addresses these with its Image, Memory and Video, and Timeline
tools. Its accuracy is highest on boundary types, 90.0\% on Q07B and 91.7\%
on Q07D. On the hardest type, Q03, it reaches 38.3\%, compared with 34.2\% for the
best frontier VLM.

\par\WFclear
\endgroup
\clearpage
\subsection{Qualitative analysis}
\label{sec:supp_case_qualitative}
\begin{figure}[H]
\centering
\includegraphics[width=0.80\textwidth]{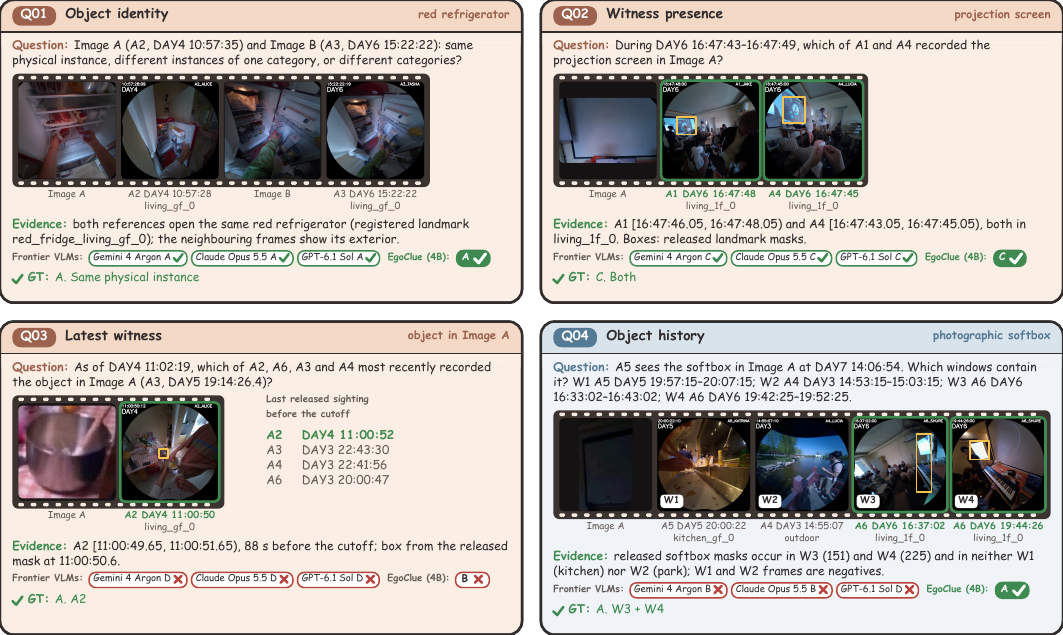}
\caption{\textbf{Evidence-annotated examples: object-instance memory.} Evidence
windows come from the gold records. Yellow boxes are released masks at the
shown timestamp. Frame labels give wearer, day, time, and the wearer's
reviewed room. In Q04, W1 and W2 frames illustrate the negative windows. Chips
report the three frontier VLMs and \agentname{} (green when correct).}
\label{fig:supp_case_objects}
\par\vspace{6pt}

\includegraphics[width=0.80\textwidth]{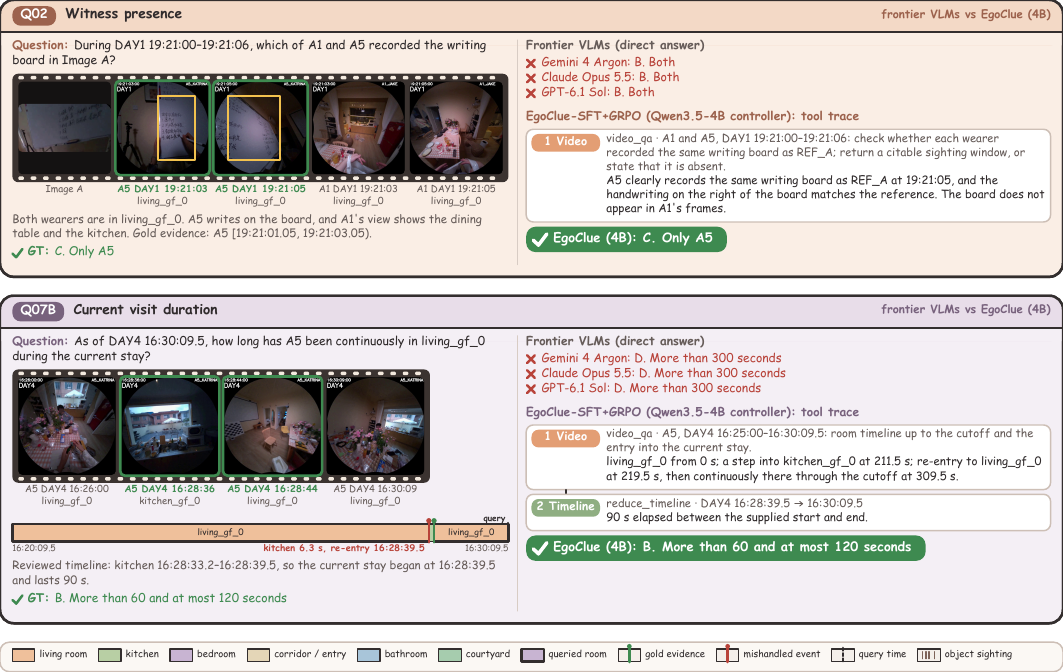}
\caption{\textbf{Frontier VLMs versus \agentname.} Both questions are wrong for
all three frontier VLMs and correct for \agentname-SFT+GRPO. Over all questions, the three
frontier VLMs are all wrong on 229 answered questions, and \agentname{}
answers 116 of them. Tool calls and results are condensed, and Chinese
controller text is translated.}
\label{fig:supp_case_agent}
\end{figure}

\begin{figure*}[!htbp]
\centering
\includegraphics[width=0.95\textwidth]{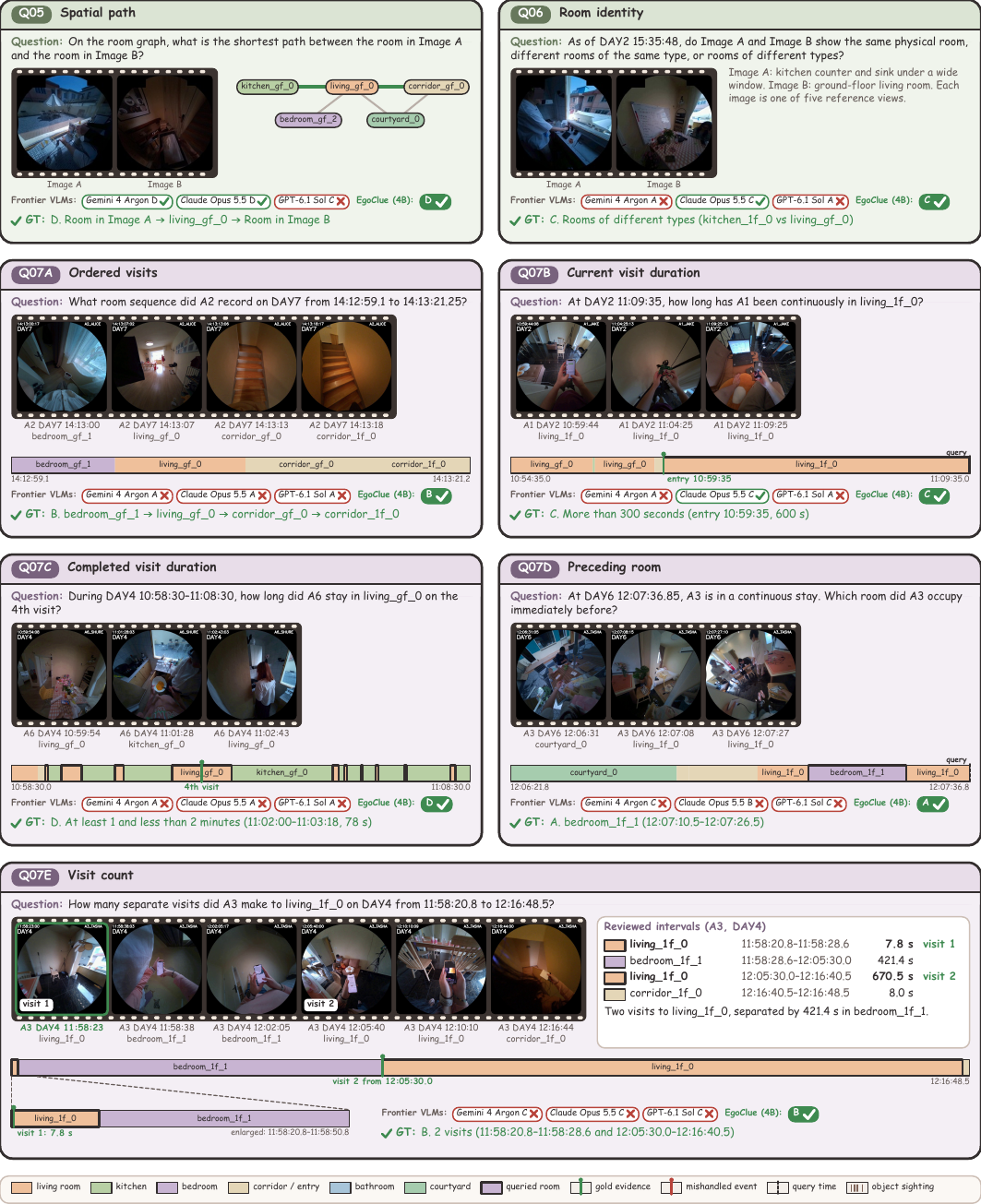}
\caption{\textbf{Evidence-annotated examples: room and visit memory.} Timelines
show reviewed room intervals over each question's interval with the shared
legend; the outlined segments are the queried room. Q05 and Q06 references
show one of five views, and the Q05 graph highlights the shortest path on the
supplied room graph. Q07E enlarges the first 30\,s of its interval, where the
first visit lasts 7.8\,s. Chips report the three frontier VLMs and \agentname{}.}
\label{fig:supp_case_rooms}
\end{figure*}

\Cref{fig:supp_case_objects,fig:supp_case_rooms} annotate one question per
type with references, evidence frames, answers, and model chips.

\Cref{fig:supp_case_agent} contrasts the frontier VLMs with \agentname{} on two
questions. In Q02, one Video call matches the board's handwriting to the
reference and separates the wearer who writes on the board from the one facing
the dining table. In Q07B, Video finds a 6-s kitchen visit and Timeline
measures the stay from re-entry.

\stopcontents[supp]

\end{document}